\documentclass[twocolumn]{article}
\usepackage[pagebackref,breaklinks,colorlinks]{hyperref}
\usepackage[a4paper, left=1.5cm, right=1.5cm, top=2cm, bottom=2.5cm]{geometry}
\usepackage[nolist,nohyperlinks]{acronym}
\usepackage{graphicx} 
\usepackage{svg} 
\usepackage{longtable}
\usepackage{array}
\usepackage{booktabs}
\usepackage{amsmath}
\usepackage{amssymb}
\usepackage{listings}
\usepackage{makecell}
\usepackage{xparse}
\usepackage{algorithm}
\usepackage{algpseudocode}
\usepackage{titling}   
\usepackage{lmodern}   
\usepackage{setspace}  
\usepackage{geometry}

\usepackage{listings}
\usepackage{xcolor}
\usepackage{colortbl}
\definecolor{iccvblue}{rgb}{0.21,0.49,0.74}
\definecolor{whitesmoke}{rgb}{0.85, 0.85, 0.85}

\colorlet{shadecolor}{gray!15}
\colorlet{shadecolor2}{gray!0}

\newcommand{\fixedimg}[1]{\includegraphics[width=2cm,height=2cm,keepaspectratio]{#1}}

\newcommand{\eg}{e.g.\ }

\title{MARS: a Multimodal Alignment and Ranking System for Few-Shot Segmentation}

\author{
  \begin{tabular}{c}
    Nico Catalano\textsuperscript{*} \\
    \texttt{nico.catalano@polimi.it}
  \end{tabular}
  \and
  \begin{tabular}{c}
    Stefano Samele\textsuperscript{*} \\
    \texttt{stefano.samele@polimi.it}
  \end{tabular}
  \and
  \begin{tabular}{c}
    Paolo Pertino\textsuperscript{*} \\
    \texttt{paolo.pertino@polimi.it}
  \end{tabular}
  \and
  \begin{tabular}{c}
    Matteo Matteucci \\
    \texttt{matteo.matteucci@polimi.it}
  \end{tabular}
}

\date{
  Politecnico di Milano \\
  \vspace{1ex}
  \textsuperscript{*}Equal contribution
}

\begin{document}

\maketitle

\begin{abstract}
\acl{FSS} aims to segment novel object classes given only a handful of labeled examples, enabling rapid adaptation with minimal supervision.
Current literature crucially 
lacks a selection method that goes beyond visual similarity between the query and example images, leading to suboptimal predictions. We present MARS, a plug-and-play ranking system that leverages multimodal cues to filter and merge mask proposals robustly. Starting from a set of mask predictions for a single query image, we score, filter, and merge them to improve results. Proposals are evaluated using multimodal scores computed at local and global levels. 
Extensive experiments on COCO-20\textsuperscript{i}, Pascal-5\textsuperscript{i}, LVIS-92\textsuperscript{i}, and FSS-1000 demonstrate that integrating all four scoring components is crucial for robust ranking, validating our contribution.
As MARS can be effortlessly integrated with various mask proposal systems, we deploy it across a wide range of top-performer methods and achieve new state-of-the-art results on multiple existing benchmarks. Code will be available upon acceptance.
\end{abstract}
    
\vspace{-7.5px}
\section{Introduction}
\begin{figure}[!t] 
    \centering
    \includesvg[width=0.850\columnwidth,inkscapelatex=false]{teaser.svg} 
    \vspace{-2mm}
    \caption{Model Overview: The query image and support set undergo visual and textual analysis. Combining different modalities, we select candidates from the mask proposals generator. The selected ones are merged to form the final prediction.}
    \label{fig:teaser} 
\end{figure}
\label{sec:intro}
Extending the capabilities of traditional segmentation models to new classes often requires retraining with large annotated datasets. This data-collecting process becomes especially prohibitive in domains such as medical imaging, industrial applications, or satellite imagery. 
\acf{FSS} \cite{BMVC2017_167,rakelly2018few} offers a promising alternative by reusing knowledge acquired from diverse classes and tasks, enabling models to adapt to new classes with only a handful of annotated examples. Despite the recent advancements in the field, top-performing methods often fail because of visual dissimilarity between the provided example and the query image. 
We contribute to this research field by presenting a Multimodal Alignment and Ranking System (MARS). The method is capable of scoring, filtering, and merging mask proposals from any existing \ac{FSS} method, effectively boosting the performance of several algorithms across different benchmark datasets. A high-level depiction of the model is presented in Figure \ref{fig:teaser}.
We achieve this goal leveraging four distinct scores: i) \textit{Local Conceptual Score} ($LC$), ii) \textit{Global Conceptual Score} ($GC$), iii) \textit{Local Visual Score} ($LV$), and iv) \textit{Global Visual Score} ($GV$). 
\textit{Visual} scores are derived from direct correspondences between the support and query images, following established approaches \cite{Matcher, GFSAM}. However, we conjecture that visual cues alone may prove insufficient in challenging scenarios characterized by intraclass variation, occlusions, challenging camera angles, and cut-offs. To overcome these limitations, we introduce two \textit{conceptual} scores based on the inferred class name and definition of the subject of interest, able to establish a higher-order semantic similarity.
Summarizing, our work makes the following contributions:

\begin{itemize}
    \item \textbf{Multimodal Approach:} We introduce a novel multimodal framework for \ac{FSS} that leverages both visual and conceptual cues.
    \item \textbf{Scoring System:} We propose a plug-and-play ranking with four scores: three novel and one EMD based.


    \item \textbf{State-of-the-Art Performance:} Extensive experiments on benchmark datasets demonstrate that our approach establishes new state-of-the-art results.
\end{itemize}

\section{Related Works}
Traditional \ac{FSS}  \cite{zhang2020sg,dong2018few,wang2021variational,liu2020part,hsnet21,vat22} approaches typically rely on pre-trained backbones networks, such as ResNet \cite{heZRS15, zagoruyko2016wide}, VGG \cite{simonyan2014very} or DINO-trained ViT \cite{dosovitskiy2020image, caron2021emerging, oquab2023dinov2}, to extract rich feature representations from both support and query images. These features are then processed by trainable modules that compare and match them to predict the query mask. 
Owing to their tailored design and dedicated training for few-shot scenarios, such methods are often referred to as specialist models.


In contrast, recent developments leverage foundational models like the \ac{SAM} \cite{kirillov2023segany}. \ac{SAM} is capable of segmenting any object when provided with various types of prompts (\eg, points, coarse masks, bounding boxes), yet it does not inherently offer semantic labeling. Consequently, adapting \ac{SAM} for \ac{FSS} shifts the emphasis from direct feature matching to an effective prompting strategy. This often requires identifying candidate points by comparing features from the support examples to those in the query images, using these points to generate prompts that guide \ac{SAM} in producing candidate mask proposals \cite{PERSAM, VRPSAM, Matcher, SegGPT, GFSAM}. 

Several innovative models have emerged along these lines. For instance, PerSAM \cite{PERSAM} introduces a training-free personalization approach for SAM. It uses one-shot data, a reference image and a rough mask of the desired concept, to generate a location confidence map that identifies the target object. Based on the confidence scores, two points are selected as positive and negative priors, which are then encoded as prompt tokens for SAM’s decoder to produce the segmentation.  

Similarly, VRP-SAM \cite{VRPSAM} incorporates a Visual Reference Prompt (VRP) encoder. Differently from SAM original prompt encoder, this one leverages both a support image with annotations and a query image to create prompt embeddings. It is composed of a Feature Augmenter, which embeds both images into a shared latent space and extracts prototype features from the support image, and a Prompt Generator that refines these features through attention mechanisms interacting with the query representation. The resulting semantic reference prompt embeddings replace traditional prompts, directly guiding SAM’s mask decoder.

Among the newer methods, Matcher \cite{Matcher} and GF-SAM \cite{GFSAM} stand out. Matcher is designed for \ac{FSS} by embedding both support and query images. Exploiting the natural correspondence between spatial locations in the feature volume and image patches, Matcher computes a similarity matrix to identify highly similar regions. High-similarity points are selected as prompts for SAM, and the subsequent mask predictions are aggregated to generate the final segmentation output.

GF-SAM takes a graph-based approach to prompt selection and ambiguity resolution. It constructs a directed graph where nodes represent point prompts and edges capture the relationships based on mask coverage. By partitioning this graph into weakly connected components, GF-SAM clusters candidate points and corresponding masks. 

Recent advancements also involve SegGPT \cite{SegGPT}, a trained-from-scratch segmentation model that unifies multiple tasks under an in-context learning framework. It processes all segmentation data in a standardized image format and employs an in-context coloring strategy, where random color mappings guide learning. This approach enables SegGPT to generalize across various segmentation tasks, including object, part, contour, and text segmentation in both images and videos.


\section{Method}
\label{sec:method}

The \ac{FSS} setting is defined by a semantic class $l$, a support set consists of $k$ samples $S(l) = \{(I^i, M^i_l)\}_{i=1}^k$, where $I^i$ is an RGB image, and a binary mask $M^i_l$   that highlights the pixels corresponding to class $l$ in $I^i$. The goal for a \ac{FSS} model $f_{\theta}$ is to predict the segmentation mask $\hat{M}_q$ for the query image $I_q$ :
\begin{equation}
    \hat{M}_q = f_{\theta}(I_q, S(l)).
\end{equation}

Our pipeline works with a set of mask proposals $M_P(I_q, S(l))$ generated by \ac{FSS} models. We refine these predictions by incorporating both visual and textual alignment. The process consists of five key steps, all leveraging pre-trained models:

\begin{enumerate}
    \item \textbf{Textual Information Retrieval Module}: We first use ViP-LLaVA \cite{cai2024vipllava} and WordNet \cite{miller1995wordnet} to extract a class name $c$ and a textual description $t$ from the support set images.
    \item \textbf{Visual-Text Alignment Module}: The query image $I_q$ is encoded using the CLIP \cite{radford2021learningtransferablevisualmodels} image encoder, while the class name $c$ is processed through the CLIP text encoder. We then compute the \textit{Refined Text Alignment} quantity, which contributes to the computation of the \textit{LC} score.
    \item \textbf{AlphaClip Module}: The dot product between AlphaCLIP \cite{sun2024alpha} embeddings of $\{c, t\}$, and $I_q$ is used as core quantity to compute \textit{GC} score.
    \item \textbf{Visual-Visual Alignment Module} Embedding features from the query image $I_q$ and support set $S(l)$ are extracted using a ViT Model \cite{oquab2023dinov2}. These features produce two key outputs: a \textit{Refined Visual Alignment} and a \textit{Cost Matrix}, which contribute to computing the \textit{LV} and \textit{GV} scores, respectively.
    \item \textbf{Filtering-Merging Module}: All the scores are always computed with respect to a mask proposal $m_i \in M_P(I_q, S(l))$. The final score is computed as a simple average of the four. A final threshold-based filtering and merging strategy outputs the best mask.
\end{enumerate}

\begin{figure}[H] 
    \centering
    \includesvg[width=\columnwidth,inkscapelatex=false]{MARS.svg} 
    \caption{Overview of the proposed MARS model. The diagram illustrates the key components and interactions within the framework. We highlight the Textual Information Retrieval, Visual-Text Alignment, Visual-Visual Alignment, and the Score-Filter-Merge Modules in different colors.}
    \label{fig:mars_overview} 
\end{figure}
\vspace{-2px}

A complete overview of our pipeline is shown in Fig.~\ref{fig:mars_overview}, with complete details provided in the following sections and Fig.~\ref{fig:VTP_VVP}.

\subsection{Textual Information Retrieval Module}

In the \ac{FSS} problem the model can leverage solely the information available in the support set to guide the segmentation of a novel class. To adhere to this principle and ensure continuity with the \ac{FSS} literature, we infer all necessary textual information directly from the support image rather than using externally provided class names. 

We employ pretrained ViP-LLaVA to extract the class name from the support image by overlaying a red contour on the target object and applying a 50\% zoom (as detailed in Section ~\ref{vipllava_supp_visual_prompts} of Supplemental Material), balancing emphasis on object detail with contextual information. In $N$-shot settings, where the support set consists of $N$ image-mask pairs, ViP-LLaVA is queried for each image individually, and a majority voting scheme is applied to select the most frequent class name for subsequent processing. This solution enables us to quickly extend all the modules based on textual information to the N-shot scenario.

In addition to the class name, we derive a broader textual description of the entity by leveraging WordNet. The extracted class name is used to query WordNet, which returns a set of candidate synsets. If a single synset is returned, its definition is directly adopted as the object's description. In cases where multiple synsets are retrieved, indicating potential ambiguity, we prompt ViP-LLaVA (using the same configuration as for class name extraction) to generate a description of the entity. 

\begin{figure}[!t]
    \centering
    \includesvg[width=\columnwidth,inkscapelatex=false]{MARSVTPExtractor_3.svg}\par
    \vspace{5mm} 
    \includesvg[width=\columnwidth,inkscapelatex=false]{MARSVVPExtractor_3.svg}
    \caption{Detailed view of the Visual-Text and Visual-Visual Alignment modules.
The \textbf{Visual-Text Alignment} module (top) extracts features from the query image using CLIP and aligns them with foreground and background textual features to generate the Softmax-GradCAM map, referred to as the unrefined text prior. This prior is subsequently refined through the PIR module, adapted from \cite{wang2024rethinking}.
The \textbf{Visual-Visual Alignment} module (bottom) computes similarity between query and support features using cosine similarity. Foreground and background priors are obtained by aggregating the respective similarity matrices, and the resulting visual prior is refined using the PIR module for downstream processing.}
    \label{fig:VTP_VVP}
\end{figure}

The VLM-generated description is then compared with the candidate definitions from WordNet, and the definition with the maximum words overlap is selected as the final class description. If no overlap is found between the VLM-generated description and any of the WordNet definitions, an empty description is used. Note that the description produced by ViP-LLaVA is used solely for matching purposes, as it may incorporate visual context specific to the support image.

The extracted textual information is used in two ways in the remainder of the pipeline. The Visual-Text Alignment module employs the class name to determine local conceptual information that enables the computation of the \textit{Local Conceptual Score} (LC). The AlphaClip Module uses the derived description together with the class name to produce the \textit{Global Conceptual Score} (GC).

\subsection{Visual-Text Alignment Module}
\label{vta_sec}
This module generates a Refined Text Alignment, a saliency map that highlights regions in the query image where the subject described by the class name is likely located. Leveraging the visual-text alignment capabilities of a pre-trained CLIP model \cite{radford2021learningtransferablevisualmodels}, the module first produces a coarse Text Alignment. Then it refines it through the PI-CLIP \cite{wang2024rethinking} Prior Information Refinement (PIR) process followed by min-max normalization. The final Refined Text Alignment is later used to compute the Local Conceptual Score by being multiplied element-wise with each binary mask proposal, with the resulting non-zero values averaged to yield a single score for ranking. A depiction of the model is provided in the upper part of Fig.~\ref{fig:VTP_VVP}.

The process inspired by \cite{wang2024rethinking, lin2023clip} begins by encoding the query image using the CLIP vision encoder. Discarding the classification token, we obtain the visual query features $F_q^v \in \mathbb{R}^{d \times (hw)}$, where $h,w$ are the embedding spatial dimensions and $d$ the number of features. A global query token is then computed by applying average pooling over the spatial dimensions:
\begin{equation}
    v_q = \frac{1}{hw}\sum_{i=1}^{hw}F_q^v(i), \quad v_q \in \mathbb{R}^{d \times 1}.
\end{equation}

In parallel, and using the class name from the Textual Information Retrieval Module, two simple text prompts, ``a \{predicted-class-name\}'', and ``a photo without \{predicted-class-name\}'', are embedded by the CLIP text encoder to obtain text features $F_{FG}^t$ and $F_{BG}^t$, which focus on the subject (foreground) and the background, respectively. Unlike PI-CLIP, which incorporates the names of all objects present in the query image using ground truth annotations, MARS employs only two prompts derived through the Textual Information Retrieval Module. The similarity between the query token and these text features is then computed using a softmax function with a temperature parameter $\tau$ (set to 0.01, as in PI-CLIP and \cite{lin2023clip}):

\begin{equation}
\resizebox{0.9\columnwidth}{!}{$
SM_{FG}, SM_{BG} = \operatorname{softmax}\left(\frac{v_q^T F_{FG}^t}{\|v_q\| \cdot \|F_{FG}^t\| \tau}, \frac{v_q^T F_{BG}^t}{\|v_q\| \cdot \|F_{BG}^t\| \tau}\right).
$}
\end{equation}

Using $SM_{FG}$, gradients with respect to each feature map are computed to obtain weights:
\begin{equation}
    w_m = \frac{1}{hw} \sum_{i,j} \frac{\partial SM_{FG}}{\partial F_q^m(i,j)},
\end{equation}
where $(i,j)$ denote spatial positions in the $m$-th feature map. The initial Text Alignement ($TA$) is then calculated as:
\begin{equation}
    TA = \operatorname{ReLU}\left(\sum_m w_m F_q^m\right).
\end{equation}

To enhance spatial coherence and semantic precision of the text prior, we employ the Prior Information Refinement (PIR) module \cite{wang2024rethinking}, which aggregates and normalizes the self-attention map $A$ from the CLIP vision encoder. A binary box mask $B$ is then derived from $TA$ via thresholding (with a threshold of 0.4, following \cite{wang2024rethinking,lin2023clipefficientsegmentertextdriven}). The Refined Text Alignment ($RTA$) is computed as:
\begin{equation}
    RTA = B \odot \text{PIR}(A) \cdot TA,
\end{equation}
where  $\odot$ represents the Hadamard product.

The resulting $RTA$ is further processed using min-max normalization. For each mask proposal $m_i \in M_P(I_q, S(l))$, where $M_P(I_q, S(l))$ is the set of all mask proposals for query image $I_q$, we compute the \textit{Local Conceptual Score} by taking the element-wise multiplication between $RTA$ and $m_i$, after the support mask is max‐pooled to the patch-feature spatial dimension of CLIP features. We then average the resulting values over all non-zero spatial positions. Finally, we also add a coverage term, which accounts for the ratio between the area of the mask proposal $m_i$ and the area of the mask obtained by merging all the masks in $M_P(I_q, S(l))$. \begin{align*} 
    \text{Cov}(m_i) &= \frac{\text{Area}(m_i)}{\text{Area}(\bigvee_{i=1}^N m_i)} \\
\end{align*} where $\text{Area}(\cdot)$ computes the area of the binary mask passed as an argument by counting its nonzero elements, and $\bigvee_{i=1}^N m_i$ represents the logical OR. The final formulation of the \textit{Local Conceptual} scores is as follows:
\begin{equation}
\begin{split}
    LC(m_i)  = \frac{\alpha}{\|m_i\|_1} \sum_{m_i(x,y) \neq 0} &  RTA(x,y) m_i(x,y)  + \\
    & + (1-\alpha)\text{Cov}(m_i)
\end{split}
\end{equation}

where $\|m_i\|_1$ denotes the sum of all elements in $m_i$, and $\alpha = 0.85$ a fixed weight. This score quantifies the alignment between the salient regions indicated by the textual prior and the mask proposal.

\subsection{AlphaCLIP Module}
In scenarios where several visual challenges lead to significant differences between the support and query images, purely visual approaches may fail to align the object of interest accurately. To address this, we generate a broad class description using WordNet from the class name predicted by ViP-LLaVA. This comprehensive textual description is then used to assess how well each mask proposal aligns with the expected semantics through the AlphaCLIP module \cite{sun2024alpha}, an enhanced version of the CLIP model that enables precise focus on specific regions within an image highlighted by a mask. 

Given a query image $I_q$, a set of $N$ mask proposals $M_P(I_q, S(l))$, and a single class name $c(m_i)$ and textual description $t(m_i)$, obtained by the Textual Information Retrieval module, a pre-trained AlphaCLIP computes an alignment score for each mask proposal relative to $(c,t)$. First, the text encoder $f_{txt}$ produces an embedding:
\begin{equation}
    \mathbf{e}_{txt} = f_{txt}(c, t).
\end{equation}
In this case, the text processed by AlphaCLIP is constructed by simply concatenating the class name \$c\$ with its corresponding definition \$t\$, resulting in the text: ``a \{c\}, \{t\}''.  For each mask proposal $m_i$, the image encoder $f_{img}$ processes the query image $I_q$ together with $m_i$ to generate an embedding:
\begin{equation}
    \mathbf{e}_{img}(m_i) = f_{img}(I_q, m_i), \quad i = 1, \dots, N.
\end{equation}
Both embeddings are normalized to have unit norm:
\begin{equation}
    \hat{\mathbf{e}}_{txt}(m_i) = \frac{\mathbf{e}_{txt}}{\|\mathbf{e}_{txt}\|_2}, \quad \hat{\mathbf{e}}_{img}(m_i) = \frac{\mathbf{e}_{img}^i}{\|\mathbf{e}_{img}^i\|_2}.
\end{equation}
The alignment score for each mask proposal is computed as the cosine similarity between the corresponding normalized image and text embeddings and then rescaled to the range $[0,1]$:
\begin{equation}
    GC(m_i) = \frac{\hat{\mathbf{e}}_{img}(m_i) \cdot \hat{\mathbf{e}}_{txt}(m_i) + 1}{2}.
\end{equation}

This yields the final \textit{Global Conceptual Score} for each mask proposal. As the alignment is computed between the image embedding restricted by the mask proposal and the text embedding, this score has a pure global nature with respect to the region of interest highlighted by the support set.

\subsection{Visual-Visual Alignment Module}
This module leverages solely visual information to compute both the \textit{Global Visual Score}  and the \textit{Local Visual Score}  for each mask proposal. It aims to capture holistic and localized visual similarities between the support and query images. %

Initially, patch features are extracted from the support and query images using a pre-trained DinoV2 ViT-Large using four registers \cite{darcet2024vision}, yielding feature maps $F_S$ and $F_Q$, respectively. A similarity matrix $S$ is then computed by comparing these features using a dot product. In the case of $N$ support images, we stack the relative features along one dimension. This matrix, whose rows correspond to support patches and columns to query patches, is used to derive the foreground and background similarity matrices by selectively retaining rows. Specifically, the support mask is max‐pooled to the patch-feature spatial dimension of the encoder. Only the rows corresponding to the regions covered by the support mask (or its complement) are preserved to form the foreground similarity matrix $S_{FG}$ and the background similarity matrix $S_{BG}$, respectively. We then compute the cost matrix $C_{FG}$ as:
\begin{equation}
    C_{FG} = \frac{1 - S_{FG}}{2}.
\end{equation}
The \textit{Global Visual Score} for a mask proposal $m_i$ is then computed as:
\begin{equation}
    GV(m_i) = 1 - \text{EMD}(C_{FG}, M_S, m_i),
\end{equation}
where $\text{EMD}$ denotes the Earth Mover's Distance, $M_S$ is the support mask, and $m_i$ is a mask proposal. Because of the nature of the EMD, the proposed $GC$ score considers both the magnitude of differences and the spatial relationships between the embedding distributions; it refers thus to the image part highlighted by the support mask as a whole.

The \textit{Local Visual Score} is based on a Refined Visual Alignment (RVA), which is derived from the combination of $S_{FG}$ and $S_{BG}$ as follows. For both $S_{FG}$ and $S_{BG}$, two preliminary saliency maps are computed using average and max aggregation on $S_{FG}$ and $S_{BG}$ column-wise, yielding to:
\begin{equation}
    S_{FG}^{avg}, \quad S_{FG}^{max}, \quad S_{BG}^{avg}, \quad S_{BG}^{max}.
\end{equation}

Following the findings in \cite{GFSAM}, the mean foreground similarity map captures global similarity to the reference object. Still, it may blur internal details, while the max foreground similarity map emphasizes highly similar regions, improving recall but increasing noise. To balance these effects, their Hadamard product is computed, obtaining the mixed similarity maps:
\begin{equation} 
    S_{FG}^{mix} = S_{FG}^{max} \odot S_{FG}^{avg}, \quad S_{BG}^{mix} = S_{BG}^{max} \odot S_{BG}^{avg}.
\end{equation}

Similarly to what is done through the softmax function in Section \ref{vta_sec}, we propose a background suppression mechanism in the visual domain. Specifically, the foreground similarity map is further polished by subtracting the background similarity map, effectively eliminating any residual noise and ensuring that misleading background spikes do not interfere with the subsequent scoring stage. Following these insights, the $RVA$ i computed as:

\begin{equation} 
    RVA^{'}= S_{FG}^{mix} - S_{BG}^{mix} ,
\end{equation}
$RVA^{'}$ is subsequently min-max normalized and processed through the PIR module, using all the attention maps $A_{DINO}$ extracted from DINOv2:
\begin{equation} 
    RVA = \text{PIR}(RVA^{'}, A_{DINO}) .
\end{equation}
Again a min-max normalized $RVA$ can be used to compute the score for each mask proposal $m_i$, adding the coverage:
\begin{equation}
\begin{split}
    LV(m_i) = \frac{\alpha}{\|m_i\|_1} \sum_{m_i(x,y) \neq 0} & RVA(x,y) \, m_i(x,y) + \\
    & + (1-\alpha)\text{Cov}(m_i),
\end{split}
\end{equation}
where $\|m_i\|_1$ denotes the sum of all elements in the mask, effectively normalizing the score by the area of the mask, and $\alpha=0.85$.

The design of the two scores ensures that both global alignment and localized feature correspondences are effectively captured, thereby complementing the textual cues provided by the other modules in our pipeline. A depiction of the model is provided in the bottom part of Fig.~\ref{fig:VTP_VVP}.

\subsection{Filtering-Merging Module}

Each mask proposal $m_i \in M_P(I_q, S(l))$, where $M_P(I_q, S(l))$ is the set of all mask proposals related to query image $I_q$, is evaluated using four distinct scores: $LC(m_i)$, $GC(m_i)$, $LV(m_i)$, $GV(m_i)$. To derive an overall confidence measure for each proposal, we compute the MARS score as the average of these four scores:
\begin{equation}  
    \begin{split}  
    \text{MARS}(m_i) =  \frac{1}{4} \Bigl(
            LC(m_i) +
            GC(m_i) + \\
            LV(m_i) +
            GV(m_i)    
            \Bigr).
    \end{split}
\end{equation}
We then apply a two-step filtering procedure. First, a static threshold $th_s = 0.55$ filters out lower-confidence proposals. 
\begin{equation}
    M_P^{th_s}(I_q) = \{ m_i \in M_P(I_q, S(l)) \mid \text{MARS}(m_i) \geq th_s \}.
\end{equation}
if $M_P^{th_s}(I_q) = \varnothing$, we apply a dynamic threshold $th_d = 0.95$ to $M_P(I_q, S(l))$:
\begin{equation}
\begin{split}
    M_P^{th_d}(I_q) = & \{ m_i \in M_P(I_q, S(l)) \mid \\
    & \text{MARS}(m_i) \geq \max_{m_j\in M_P(I_q, S(l))}\text{MARS}(m_j)*th_d \}.
\end{split}
\end{equation}

The fixed threshold, the dynamic threshold, and the coverage weight are the only hyperparameters introduced by the method. Finally, the predicted segmentation mask $\hat{M}_q$ is obtained by performing a logical OR operation over the filtered proposals:
\begin{equation}
    \hat{M}_q = \bigvee_{m_i \in M_P^{\text{filtered}}} m_i.
\end{equation}

\section{Experiments}
\label{sec:experiments}

In our experiments, we evaluate the performance of the proposed MARS on four widely used datasets for \ac{FSS}: COCO-20\textsuperscript{i} \cite{nguyen2019feature}, Pascal-5\textsuperscript{i} \cite{BMVC2017_167,hariharan2014simultaneous}, LVIS-92\textsuperscript{i} \cite{Matcher}, and FSS-1000 \cite{FSS1000}. We consider both 1-shot and 5-shot scenarios, where one or five support images and their corresponding segmentation masks are provided to guide the segmentation of a query image.

To assess the general applicability of MARS, we selected several state-of-the-art algorithms from the \ac{FSS} literature to generate mask proposals and coupled them with MARS. Specifically, we combined MARS with \textbf{Matcher}, \textbf{GF-SAM}, \textbf{VRP-SAM}, \textbf{PerSAM}, and \textbf{SegGPT}—all of which represent the top performers in \ac{FSS} task.

Segmentation quality is quantified using the mean \ac{mIoU} as the evaluation metric, computed following the procedure outlined in \cite{Matcher}. Detailed information about the prompts used for ViP-LLaVA to extract class names and descriptions, the exact specifications of the DinoV2 pre-trained ViT and associated feature maps, as well as the configurations for the ViP-LLaVA, AlphaClip, and CLIP models are provided in the Appendix of the Supplementary Materials. In addition, we adopted the coverage parameter $\alpha = 0.85$ and the values of the static threshold $th_s = 0.55$ and dynamic threshold $th_d = 0.95$ on every experiment.

\begin{table*}[htbp]
    \centering
    \small
    \begin{tabular}{c||c|c|c|c|c}
    Method & EMD & AlphaClip Score & \textit{RVA} ranking & \textit{VTA} ranking & mIoU \\
    \toprule  
      \rowcolor{shadecolor} Matcher + GV Score & \checkmark & & & & 35.2 \\
      \rowcolor{shadecolor2} Matcher + GC Score & & \checkmark & & & 38.0 \\
      \rowcolor{shadecolor} Matcher + LV Score & & & \checkmark & & 45.4 \\
      \rowcolor{shadecolor2} Matcher + LC Score & & & & \checkmark & 43.3 \\
      \rowcolor{shadecolor} Matcher + Global Score & \checkmark & \checkmark & & & 41.9 \\
      \rowcolor{shadecolor2} Matcher + Local Score & & & \checkmark & \checkmark & 48.7 \\
      \rowcolor{shadecolor} Matcher + Conceptual Score & & \checkmark & & \checkmark & 50.0 \\
      \rowcolor{shadecolor2} Matcher + Visual Score & \checkmark & & \checkmark & & \underline{56.0} \\
      \rowcolor{shadecolor} Matcher + MARS & \checkmark & \checkmark & \checkmark & \checkmark & \textbf{60.5} \\
    \bottomrule
    \end{tabular}
    \caption{Ablation studies on the COCO-20\textsuperscript{i} dataset. Expanding a single score along a specific dimension always lead to improvement. Cooperation between all four scores achieves the best performance, highlighted in bold. Second to best underlined.}
    \label{tab:ablation_studies}
\end{table*}
 
\subsection{Ablation Studies}
To evaluate the contribution of each component within the MARS pipeline, we conduct a series of ablation studies on the COCO-20\textsuperscript{i} dataset. Since our ranking system integrates four distinct types of information—visual, conceptual, local, and global—each forming mutually exclusive pairs, we systematically test all possible configurations. First, we reduce the ranking system to a single component (e.g. \textit{global visual}, \textit{global conceptual}, \textit{local visual}, \textit{local conceptual} scores) to analyze its individual impact. Next, we group components based on broader categories, such as metric scale (global-local) or conceptual relevance (visual-conceptual). All experiments are conducted on COCO-20\textsuperscript{i} using Matcher as the mask proposal system. Hyperparameters of the pipeline are the ones of the default settings.

Table~\ref{tab:ablation_studies} summarizes the performance of different configurations in terms of \ac{mIoU}. Notably, combining metrics either by scale (EMD + AlphaClip Score: Global, \textit{RVA} + \textit{VTA}: Local) or by information type (EMD + \textit{RVA}: Visual, AlphaClip Score + \textit{VTA}: Text) leads to significant improvements over using individual metrics alone. This result highlights the complementary nature of the four metrics in assessing the quality of proposed masks. Consequently, it is only through the integration of all four ranking components within MARS that we achieve best results on COCO-20\textsuperscript{i}.

We also observe that the vision-only ranking system attains the second-highest performance, underlining the dominant role of visual information.

\subsection{One-Shot and Five-Shot Segmentation Results}
As the core aspect of a \ac{FSS} pipeline based on a foundational model such as SAM is both the masks' proposition phase and the subsequent ranking, we further specify the MARS plug-in point for each tested method. For Matcher, we have collected all the unfiltered mask proposals generated after the promptable segmenter module. Similarly, for GF-SAM, we apply our ranking system after the mask generator step and before the point-mask clustering step. The two methods generate many proposals and our ranking system
can be tested in a scenario with much possible noise. With PerSAM, as the method proposes a three-steps-cascade-mask-generation refinement, we have collected all proposals produced in each step, for a total number of 9. VRP-SAM is even narrower with its mask proposition, leveraging only one single mask creation based on SAM. We thus chose to force the creation of three masks through SAM API and collect them. These methods prove the effectiveness of MARS even when few options are available.

Last, SegGPT is trained to produce a single prediction for a query image. Thus we developed a query system that leverages three easy data augmentation steps. We apply a random rotation between -30 and +30 degrees, a random horizontal flip, and a random vertical flip, both with 50\% probability. We use these pre-processing operations on the query image to retrieve nine different masks from the model, plus the one generated from the original query image. This scenario is exciting, as generating random transformations of the query image could potentially benefit or harm the target mask generation. Nevertheless, thanks to MARS, we can retrieve the correct prediction and even improve overall performance. All other specific hyperparameters are left as in the original work for each method.

Table~\ref{tab:one_shot_results} shows the performance of five state-of-the-art few-shot segmentation methods (PerSAM, VRP-SAM, GF-SAM, Matcher, SegGPT) on the one-shot segmentation task, together with their MARS-refined version. For each method, we downloaded the official code and re-executed or re-evaluated the performances on the datasets (Original column). VRP-SAM scores on LVIS-92\textsuperscript{i} are missing as the computational requirements for training the model in that scenario are over-demanding.

\begin{table*}[ht]
\small
\centering
\caption{One-Shot Segmentation Results. For each dataset, the table reports the original performance and the performance after applying  MARS in terms of \ac{mIoU}. In each column, we underlined the top performer method for a particular dataset.} 
\begin{tabular}{c|cc|cc|cc|cc}
\toprule
Method & \multicolumn{2}{c|}{COCO-20\textsuperscript{i}} & \multicolumn{2}{c|}{Pascal-5\textsuperscript{i}} & \multicolumn{2}{c|}{LVIS-92\textsuperscript{i}} & \multicolumn{2}{c}{FSS-1000} \\
 & Original & + MARS & Original & + MARS & Original & + MARS & Original & + MARS \\
\midrule
PerSAM    & 21.4 & \textbf{28.8} (\textbf{+7.4})  & 43.1 & \textbf{56.0} (\textbf{+12.9}) & 12.3 & \textbf{13.6} (\textbf{+1.3}) & 75.0 & \textbf{79.1} (\textbf{+4.1}) \\
VRP-SAM   & 50.1 & \bf{52.6 (+2.5)} & \bf{69.2} & 67.6 (-1.6) & - & - & \textbf{87.9} & 86.4 (-1.5) \\
SegGPT    & 54.5 &  \textbf{59.9} (\textbf{+5.4})& 83.2 &  \textbf{\underline{84.1}} (\textbf{+0.9}) & 20.8 & \textbf{24.0} (\textbf{+3.2}) & 83.3 & \textbf{84.3} (\textbf{+1.0})\\
Matcher   & 52.7 &  \textbf{60.5} (\textbf{+7.8}) & 68.1 & \textbf{77.2} (\textbf{+9.1}) & 33.0 & \textbf{36.9} (\textbf{+3.9}) & \textbf{87.0} & 85.4 (-1.6)\\
GF-SAM    & 58.7 & \textbf{\underline{61.9}} (\textbf{+3.2}) & 72.1 & \textbf{75.7} (\textbf{+3.6}) & 35.2 & \textbf{\underline{38.7}} (\textbf{+3.5}) & \textbf{\underline{88.0}} & 87.0 (-1.0) \\
\bottomrule
\end{tabular}
\label{tab:one_shot_results}
\end{table*}

\begin{table*}[ht]
\centering
\caption{Five-Shot Segmentation Results. For each dataset, the table reports the original performance in terms of \ac{mIoU} and the performance after applying MARS . In each column, we underlined the top performer method for a particular dataset.}
\begin{tabular}{c|cc|cc|cc|cc}
\toprule
Method & \multicolumn{2}{c|}{COCO-20\textsuperscript{i}} & \multicolumn{2}{c|}{Pascal-5\textsuperscript{i}} & \multicolumn{2}{c|}{LVIS-92\textsuperscript{i}} & \multicolumn{2}{c}{FSS-1000} \\
 & Original & + MARS & Original & + MARS & Original & + MARS & Original & + MARS \\
\midrule
SegGPT    & 61.2 & \textbf{64.3} (\textbf{+3.1}) & 86.8 & \textbf{\underline{87.8}} (\textbf{+1.0}) & 22.4 & \textbf{23.5} (\textbf{+0.9}) & 86.2 & \textbf{86.3} (\textbf{+0.1})\\
Matcher   & 60.7 & \textbf{63.6} (\textbf{+2.9}) & 74.0 & \textbf{80.7} (\textbf{+6.7}) & 40.0 & \textbf{40.5} (\textbf{+0.5}) & \textbf{\underline{89.6}} & 87.6 (-2.0) \\
GF-SAM    & 66.8 & \textbf{\underline{67.8}} (\textbf{+1.0}) & \textbf{82.6} & 81.5 (-1.1) & 44.0 & \textbf{\underline{46.7}} (\textbf{+2.7}) & \textbf{88.9} & 87.5 (-1.4) \\
\bottomrule
\end{tabular}
\label{tab:five_shot_results}
\end{table*}

As shown in Table~\ref{tab:one_shot_results}, MARS consistently enhances the performance of all methods and, on most datasets, even surpasses current state-of-the-art scores in 1-shot segmentation. However, for VRP-SAM, the improvement is less significant. We suspect that this is primarily because, for this method, we generate only three mask proposals, which may limit the effectiveness of our ranking system. The only exception is observed on FSS-1000, where our method yields slightly lower scores compared to top-performing models. We hypothesize that this discrepancy is primarily due to the characteristics of the FSS-1000 dataset, which predominantly contains images with large, clearly defined target objects set against simplistic backgrounds. In such oversimplified scenarios, segmentation scores are already near saturation and minor variations in mask predictions can lead to disproportionate fluctuations in the \ac{mIoU} metric. This limits the benefits of our approach, which is designed to excel in complex scenarios by leveraging subtle multimodal cues.

In Table~\ref{tab:five_shot_results}, we report the performance of methods in the 5-shot scenario. The overall performance gain is lower compared to the 1-shot case. Increasing the number of images in the support set refines visual information by enriching the similarity matrix with contributions from five distinct variations of the target object. This improves the model’s ability to generalize based on visual cues. On the other hand, the larger support set offers limited additional benefits for the conceptual components because the multiple support images are simply summarized within the Textual Information Retrieval module, not contributing to subsequent conceptual stages. Nevertheless, these gains are consistently observed across various mask generation methods and datasets, further validating MARS’s ability to enhance state-of-the-art performance even in the 5-shot scenario. We have reported in the Appendix of the Supplementary Materials several visualizations of the final mask predictions.

\subsection{A Simple MARS Pipeline}
In this scenario, we assess the autonomous potential of MARS in a single/few-shot segmentation pipeline that relies on minimal prior information. To isolate the effectiveness of our ranking system, we developed a simplified pipeline based on the SAM Automatic Mask Generator (AMG). In this setup, SAM generates mask proposals from a uniform random sampling grid of points, without any guidance regarding the target object. Further information about the SAM mask generation procedure used can be found in the Supplementary Materials. The main motivation of this experiment is to evaluate how well MARS can perform when the quality of mask proposals is deliberately suboptimal, thus removing any bias introduced by advanced FSS-specific proposal methods. This allows us to assess the intrinsic strength of our ranking and merging strategy in handling noisy or imprecise proposals.

Table~\ref{tab:combined_results} provides a comparative analysis between the proposed AMG + MARS pipeline and most relevant published methods. Despite the simplicity of the mask proposal stage, our approach has the highest score on COCO-20\textsuperscript{i} (1-shot) and remains competitive with top-performing methods in other settings. These results underscore the strength of our ranking system, asserting the dominance of the mask selection procedure over the mask proposal stage in \ac{FSS}.

\begin{table}[ht]
\centering
\caption{Performance of well-known methods on one-shot and five-shot segmentation tasks. The table reports \ac{mIoU} scores. We have highlighted in bold the top performer method and underlined the second-best.\\}

\resizebox{\columnwidth}{!}{
\begin{tabular}{l|cc|cc}
\toprule
 & \multicolumn{2}{c|}{COCO-20\textsuperscript{i}} & \multicolumn{2}{c}{Pascal-5\textsuperscript{i}} \\
Method              & 1-shot & 5-shot & 1-shot & 5-shot  \\
\midrule
HSNet \cite{hsnet21}               & 41.2  & 49.5  & 66.2  & 70.4    \\
VAT \cite{vat22}                 & 41.3  & 47.9  & 67.5  & 71.6    \\
PerSAM              & 21.4  &   -   & 43.1  &    -   \\
VRP-SAM             & 50.1  &    -  & 69.2  &    -   \\
SegGPT              & 54.5  & 61.2  & \textbf{83.2}  & \textbf{86.8}   \\
Matcher             & 52.7  & 60.7  & 68.1  & 74.0   \\
GF-SAM              & \underline{58.7}  & \textbf{66.8}  & 72.1  & \underline{82.6}   \\
\textbf{AMG + MARS (Ours)}   & \textbf{59.2}  & \underline{64.6}  & \underline{74.3}  & 80.9 \\
\bottomrule
\end{tabular}%
}
\label{tab:combined_results}
\end{table}

\section{Conclusions and Future Works}
In this paper, we presented MARS, a multimodal alignment and ranking system for \ac{FSS} that overcomes the limitations of relying solely on visual similarity. Our framework computes four complementary scores: \textit{Local Conceptual Score} ($LC$), \textit{Global Conceptual Score} ($GC$), \textit{Local Visual Score} ($LV$), and \textit{Global Visual Score} ($GV$) to evaluate mask proposals. The key innovation lies in the integration of conceptual scores derived from both the class name and a more generic textual description, extracted from resources such as WordNet and via ViP-LLaVA, which serves as a semantic anchor. This multimodal approach effectively mitigates challenges arising from significant visual differences between the support and query images by focusing on the inherent semantic attributes of the object.

Extensive experiments on datasets such as COCO-20\textsuperscript{i}, Pascal-5\textsuperscript{i}, LVIS-92\textsuperscript{i}, and FSS-1000 demonstrate that incorporating MARS into various segmentation methods (including Matcher, GF-SAM, VRP-SAM, PerSAM, and SegGPT) leads to substantial improvements over state of the art performance. Ablation studies further confirm that the fusion of local and global, as well as visual and conceptual information, is essential for robust segmentation. Notably, experiments on AMG + MARS, which represent a less optimal proposal generation scenario, validate the strengths of MARS and suggest that it can query a clustering model and transform it from an agnostic segmentation model into a \ac{FSS} model.


\begin{acronym}
    \acro{FSS}{Few Shot Segmentation}
    \acro{SS}{Semantic Segmentation}
    \acro{FSL}{Few Shot Learning}
    \acro{CD-FSS}{Cross Domain FSS}
    \acro{ZS3}{Zero Shot Semantic Segmentation}
    \acro{OVSS}{Open-Vocabulary Semantic Segmentation}
    \acro{CLIP}{Contrastive Language–Image Pre-training}
    \acro{SAM}{Segment Anything Model}
    \acro{FM}{Foundation Model}
    \acro{MLLM}{Multimodal Large Language Models}
    \acro{VLM}{Vision-Language Models}
    \acro{CV}{Computer Vision}
    \acro{NLP}{Natural Language Processing}
    \acro{DM}{Diffusion Model}
    \acro{DDPM}{Denoising Diffusion Probabilistic Models}
    \acro{ICL}{In-Context Learning}
    \acro{ICS}{In-Context Segmentation}
    \acro{mIoU}{mean Intersection Over Union}
\end{acronym}

\clearpage
{
     \small
    \bibliographystyle{plain}
    \bibliography{main}
}

\clearpage
\section{Supplementary Material} 

These supplemental materials provide additional details and qualitative results to complement the main manuscript.

\subsection{Text Prompt For ViP-LLaVa}
\label{vipllava_supp}
We retrieve the class name by prompting ViP-LLaVA  with the support image and the text prompt:
\begin{lstlisting}[caption={ViP-LLaVA Class Name Prompt}]
Human: <image> What is the name of the object inside the red mask contour in the image? Your output must be only the class name of the object. Do not add any additional text.
\end{lstlisting}

We retrieve the object description of the class name from ViP-LLaVa with the following prompt:
\begin{lstlisting}[caption={ViP-LLaVA Class Description Prompt}]
Human: <image> Given the image provided, identify and provide the definition of the {} inside the {} mask contour. Assistant:
\end{lstlisting}

\subsection{Visual Prompt For ViP-LLaVa}
\label{vipllava_supp_visual_prompts}

\begin{table*}[ht]
    \centering
    \caption{Comparison of visual prompt configurations for class name extraction on COCO-$20^i$.  
    “Acc. Exact Matching” reports the percentage of predictions that exactly match the ground-truth label;  
    “Acc. Matching Wordnet Synset” reports the percentage of predictions that match via shared WordNet synsets.  
    The best configuration is highlighted in bold.}
    \resizebox{\textwidth}{!}{
   \begin{tabular}{c c c c c c c}
    \toprule
    Prompt Type & Color & Alpha & Thickness & Zoom & Acc. Exact Matching (\%) & Acc. Matching Wordnet Synset(\%) \\
    \midrule
    mask overlay   & red   & 0.3 & 1  & 0    & 56.37 & 60.74 \\
    bounding box   & red   & 0.5 & 2  & 0    & 62.60 & 68.29 \\
    bounding box   & red   & 0.5 & 2  & 30   & 63.45 & 69.80 \\
    bounding box   & red   & 0.5 & 2  & 50   & 62.82 & 69.07 \\
    mask contour   & red   & 0.5 & 2  & 0    & 62.30 & 68.75 \\
    mask contour   & red   & 0.5 & 2  & 30   & 65.39 & 71.75 \\
    mask contour   & green & 0.5 & 2  & 50   & 65.70 & 72.23 \\
    \textbf{\underline{mask contour}}   & \textbf{\underline{red}}   & \textbf{\underline{0.5}} & \textbf{\underline{2}}  & \textbf{\underline{50}}   & \textbf{\underline{65.75}} & \textbf{\underline{72.25}} \\
    \bottomrule
  \end{tabular}
    }
    
    \label{table:prompt_types_vipllava_accuracy_exp}
\end{table*}

We prompt ViP-LLaVA on the support image using a mask contour highlight (red, $\alpha=0.5$, thickness=2) and a 50\% zoom. This configuration is compared with competing ones in Table~\ref{table:prompt_types_vipllava_accuracy_exp}) where we compare three prompt types (mask overlay, bounding box, mask contour), varied contour color, transparency (\texttt{alpha}), thickness and zoom level, and measuring a configuration  accuracy as: 

\[
    \text{Accuracy} \;=\;\frac{\#\{\text{matched names}\}}{\#\{\text{total predictions}\}}\,.
\]

Predicted and ground-truth labels are first lowercased and string-compared; only exact matches count as correct. To have a metric more resilient to synonyms, in the column ``{Acc. Matching Wordnet Synset(\%)" we report the Accuracy scores considering a positive match when the prediction and ground truth share the same set of synsets in WordNet; otherwise, they are incorrect.

\medskip
\noindent\textbf{Zoom‐based Cropping.} We compute the tightest bounding rectangle around all connected mask components, expand its width and height by the zoom percentage (clipped to image borders), crop, then resize back to the original frame as exemplified in Algorithm \autoref{algo:viplava_zoom}. When the object already fills most of the image, the result is nearly identical to the original; when the object is small or off-center, the expansion brings it closer into focus so that the vision-language model can attend to it more effectively.

\medskip
\begin{algorithm}[t]
  \caption{Zoom‐based Cropping for ViP-LLaVA Prompting}
  \label{alg:vipllava_zoom}
  \begin{algorithmic}[1]
    \Require image \(I\), mask \(M\), zoom \(\,Z\%\)
    \Ensure zoomed crop \(C\)
    \State \(\text{bbox}\gets\text{get\_bbox}(M)\)
    \State \((\text{bbox.x},\text{bbox.y}) \gets (\text{bbox.x}_{\text{top-left}},\text{bbox.y}_{\text{top-left}})\)
    \State \((w,h)\gets(\text{bbox.width},\,\text{bbox.height})\)
    \State \(\text{bbox.x}_{\text{center}}\gets\text{bbox.x}_{\text{top-left}}+w/2,\)
    \State \(\text{bbox.y}_{\text{center}}\gets\text{bbox.y}_{\text{top-left}} + h/2\)
    \State \(w'\gets \min(w\times(100/Z),\;I.\text{width})\)
    \State \(h'\gets \min(h\times(100/Z),\;I.\text{height})\)
    \State \(\text{bbox.x'}\gets\max(0,\;\text{bbox.x}_{\text{center}}-w'/2)\)
    \State \(\text{bbox.y'}\gets\max(0,\;\text{bbox.y}_{\text{center}}-h'/2)\)
    \State \(C_{\text{raw}}\gets\text{crop}(I,\;\text{bbox})\)
    \State \(C\gets\text{resize}(C_{\text{raw}},\,I.\text{width},\,I.\text{height})\)
    \State \Return \(C\)
  \end{algorithmic}
  
    \label{algo:viplava_zoom}
\end{algorithm}

\subsection{Prompt For Visual-Text Alignment Module}

The $VTA$ serves as a saliency map that highlights the regions in the query image where the object of interest, described by textual information, is likely located. It leverages the visual-textual alignment capabilities of the CLIP model \cite{radford2021learningtransferablevisualmodels}.

The construction of $VTA$ begins with the class name of the entity of interest extracted by ViP-LLaVa. Two text prompts are then generated: a positive prompt $t_{FG}$

 \begin{lstlisting}[caption={ViP-LLaVA $t_{FG}$ Prompt}] 
 a {predicted_class_name}.
 \end{lstlisting}

 and a negative prompt $t_{BG}$:
 
 \begin{lstlisting}[caption={ViP-LLaVA $t_{BG}$ Prompt}]
 a photo without {predicted_class_name}.
\end{lstlisting}

\subsection{MARS Default Configuration}
The configuration of the models used in MARS and all the experiments is as follows:
\begin{itemize}
    \item Text-extraction Module: this component uses a ViP-LLaVA vision-language model based on LLaVA-7B with 4-bit quantization. The prompting strategies employed to extract relevant information from the support set are detailed in \autoref{vipllava_supp}.
    \item Visual-Text Alignment Module: the proposed method is used to generate $VTA$, computed with a pre-trained CLIP-B/16 model. The threshold parameter in the PIR module, which is applied to refine the initial $VTA$, is set to 0.4 and employs the attention maps extracted from the last 8 self-attention layers (out of the 12) of the encoder, as reported in PI-CLIP \cite{wang2024rethinking}.
    \item Visual-Visual Alignment Module: this module employs a Vision Transformer ViT-L/14 pre-trained with DINOv2, specifically the variant using four register tokens, to extract $RVA$. The PIR module threshold used to refine $RVA$ is set to 0.85 and employs the attention maps extracted from all 24 self-attention layers of the encoder.
    \item AlphaClip Module: a pre-trained AlphaCLIP-L/14@336 model is used to compute the global-conceptual score for each mask proposal. 
    \item Filtering-Merging Module: the fixed threshold $\text{threshold}{\text{static}}$ in the filtering component is set to 0.55. The dynamic threshold $\text{threshold}{\text{dynamic}}$ is set to 0.95, ensuring that if no proposals exceed the fixed threshold, only those achieving at least 95\% of the score of the best proposal are retained.
\end{itemize}
We have always adopted original weights available from official repositories.

\subsection{Computational overhead and resource requirements}
We assessed the computational overhead introduced by MARS and found it to be minimal when compared to similar pipelines. In particular, we benchmarked the time required for each component of MARS and the competing method Matcher using the COCO-20\textsuperscript{i} dataset, where hundreds of mask proposals are ranked for a single query. All experiments were run on an NVIDIA Quadro RTX 6000 (24GB), with MARS requiring approximately 18 GB of VRAM to operate.

The initial mask proposal step, performed by SAM, takes approximately $12.1 \pm 1.31$ seconds, which is consistent across both Matcher and MARS. The main distinction lies in the ranking and merging phase. Matcher completes this step in $0.27 \pm 0.25$ seconds, while MARS requires $2.04 \pm 0.61$ seconds. Despite being slower, this additional cost remains negligible relative to the overall computation time dominated by mask generation.

Additionally, in MARS, class name prediction is required only once per support set. This step takes $4.72 \pm 0.59$ seconds and is not repeated for subsequent queries referencing the same support set. Therefore, in the typical scenario where the support set has already been processed, MARS requires approximately $13$--$14$ seconds per prediction, only slightly more than Matcher’s $12$--$13$ seconds. 

In the worst-case scenario, where the support set is encountered for the first time, the total inference time increases to $18$--$19$ seconds. However, this case occurs only once per novel support set and does not represent the standard operational cost. 

\subsection{SAM Automatic Mask Generation Hyperparameters}

In this section, we report the parameters of the \textit{SamAutomaticMaskGenerator} class (taken from the official Matcher repository) used to generate the mask proposals. 

\begin{verbatim}
points_per_side = 64,
points_per_batch = 64,
pred_iou_thresh = 0.88,
sel_pred_iou_thresh = 0.85,
stability_score_thresh = 0.95,
stability_score_offset = 1.0,
sel_stability_score_thresh = 0.90,
sel_stability_score_offset = 1.0,
box_nms_thresh = 0.65,
crop_n_layers = 1,
crop_nms_thresh = 0.7,
crop_overlap_ratio = 512 / 1500,
crop_n_points_downscale_factor = 2,
point_grids = None,
min_mask_region_area = 14,
output_mode = "binary_mask",
multimask_output = True,
sel_multimask_output = True,
output_layer = -1,
sel_output_layer = -1,
dense_pred = True
\end{verbatim}

\onecolumn
\subsection{Qualitative Results}
\renewcommand{\LTcapwidth}{\textwidth}

\begin{longtable}{>{\centering\arraybackslash}p{2cm} >{\centering\arraybackslash}p{2cm} >{\centering\arraybackslash}p{2cm} >{\centering\arraybackslash}p{2cm} >{\centering\arraybackslash}p{2cm} >{\centering\arraybackslash}p{2cm} >{\centering\arraybackslash}p{2cm}}
\caption{\textbf{Qualitative Results:} This table presents each sample as a two-part row. The upper part displays the Support Set, Query Image, Visual Prior, Text Prior, Matcher Prediction, Matcher with MARS Prediction, and the Ground Truth. The lower part provides the corresponding textual information: the dataset's class name, the predicted class name, a description of the subject class inferred by MARS using WordNet, followed by the IoU values for the Matcher prediction and the Matcher with MARS prediction.}
\label{tab:qualitative}\\

\toprule
Support Set & Query Image & Text Prior & Visual Prior & Matcher & Matcher + MARS & Ground Truth \\
\midrule
\endfirsthead

\multicolumn{7}{c}{\tablename\ \thetable\ -- continued from previous page} \\
\toprule
Support Set & Query Image & Text Prior & Visual Prior & Matcher & Matcher + MARS & Ground Truth \\
\midrule
\endhead

\midrule \multicolumn{7}{r}{Continued on next page} \\
\endfoot

\bottomrule
\endlastfoot

\begin{minipage}[t]{\linewidth}\centering \fixedimg{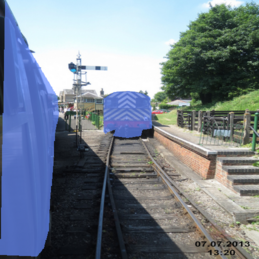}\\\end{minipage} & 
\begin{minipage}[t]{\linewidth}\centering \fixedimg{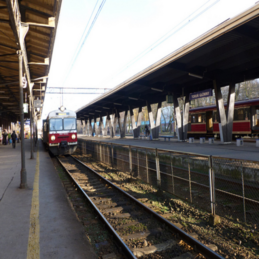}\\\end{minipage} & 
\begin{minipage}[t]{\linewidth}\centering \fixedimg{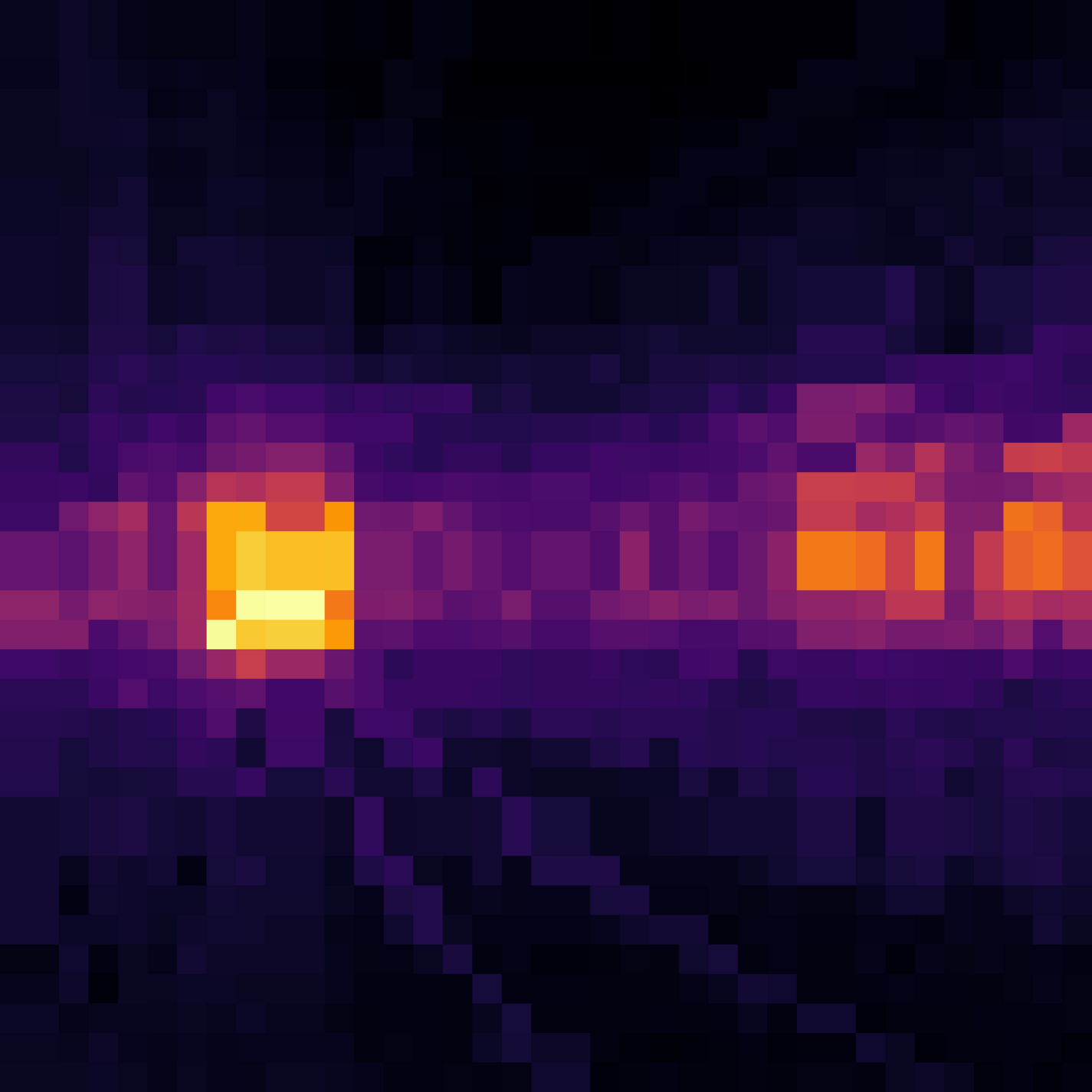}\end{minipage} & 
\begin{minipage}[t]{\linewidth}\centering \fixedimg{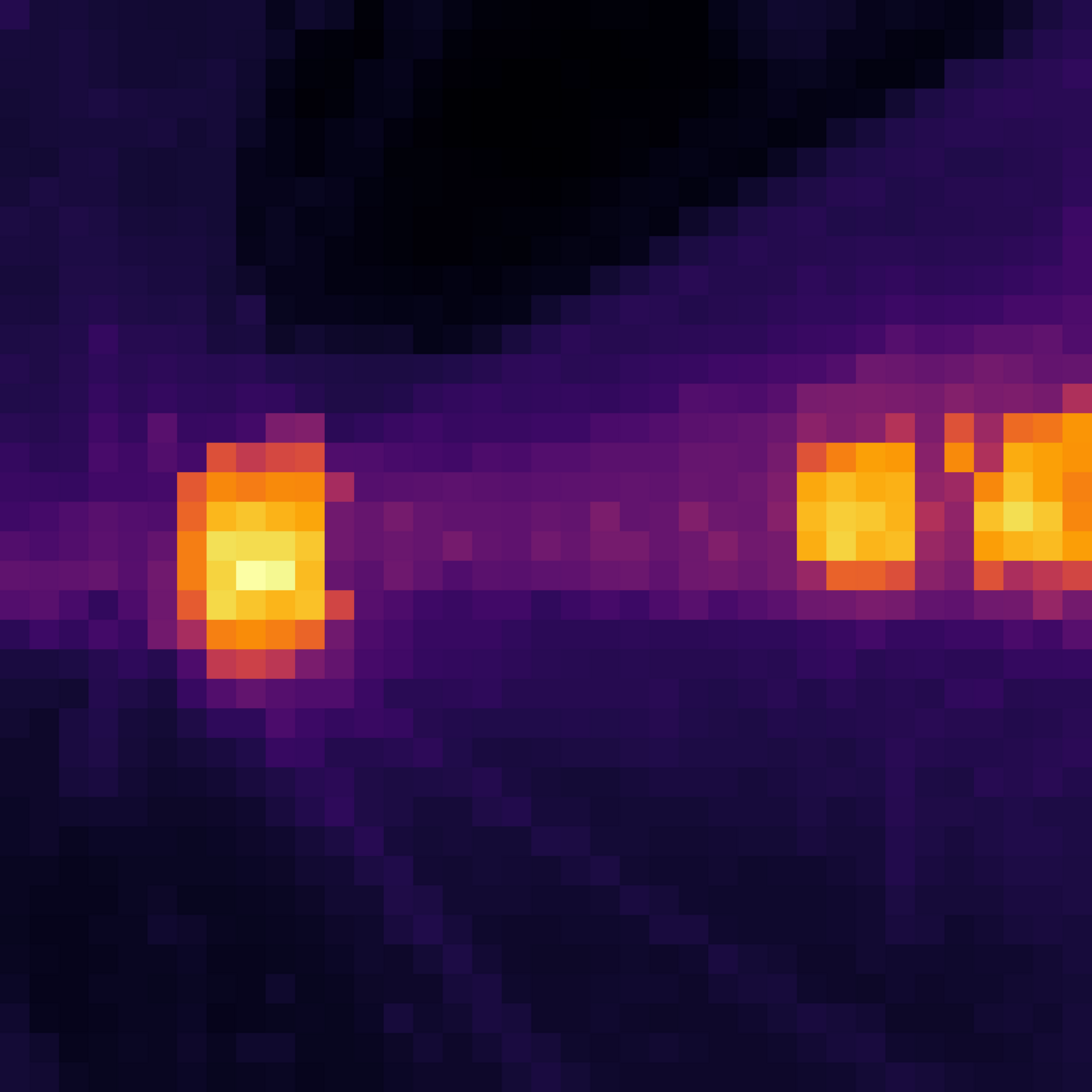}\end{minipage} & 
\begin{minipage}[t]{\linewidth}\centering \fixedimg{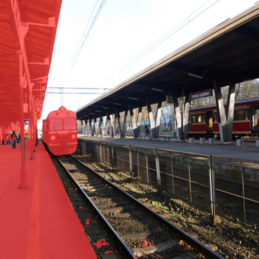}\\\end{minipage} & 
\begin{minipage}[t]{\linewidth}\centering \fixedimg{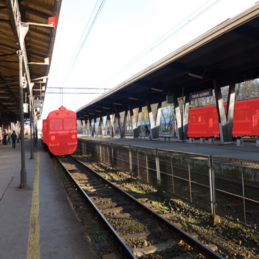}\end{minipage} & 
\begin{minipage}[t]{\linewidth}\centering \fixedimg{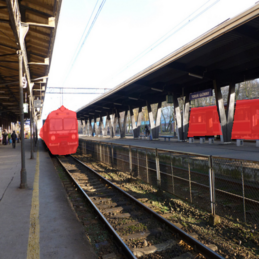}\end{minipage} \\
\centering train & \centering Train & \multicolumn{2}{>{\centering\arraybackslash}p{4cm}}{\textit{*Description not found*}} & \centering 7.38 mIoU & \centering 87.11 mIoU &  \\

\begin{minipage}[t]{\linewidth}\centering \fixedimg{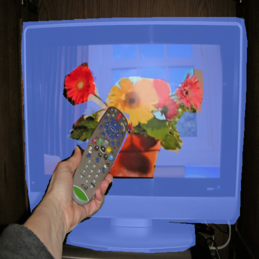}\\\end{minipage} & \begin{minipage}[t]{\linewidth}\centering \fixedimg{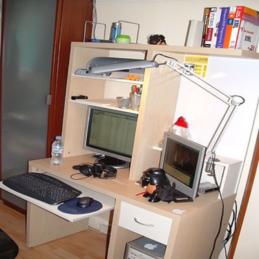}\\\end{minipage} & \begin{minipage}[t]{\linewidth}\centering \fixedimg{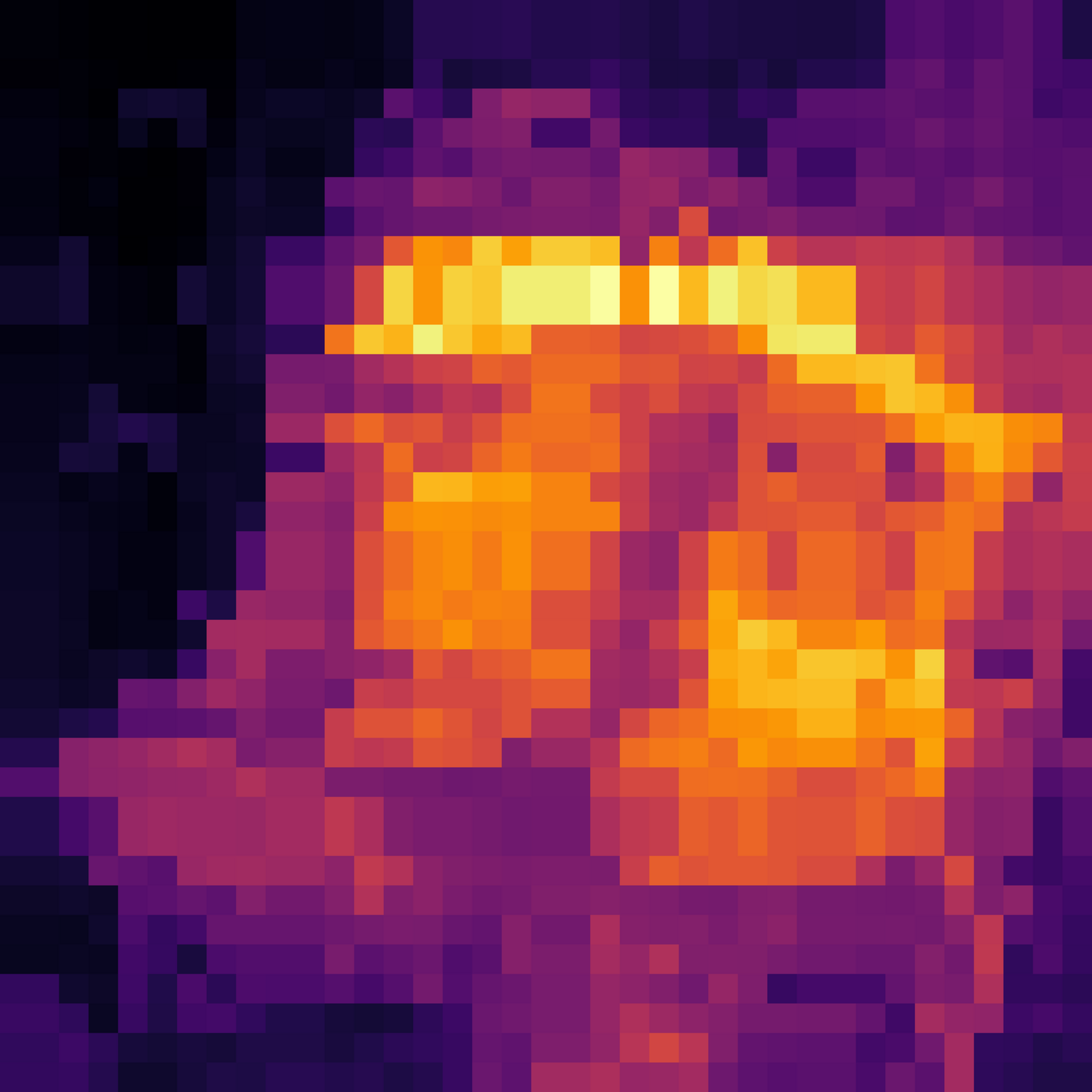}\end{minipage} & \begin{minipage}[t]{\linewidth}\centering \fixedimg{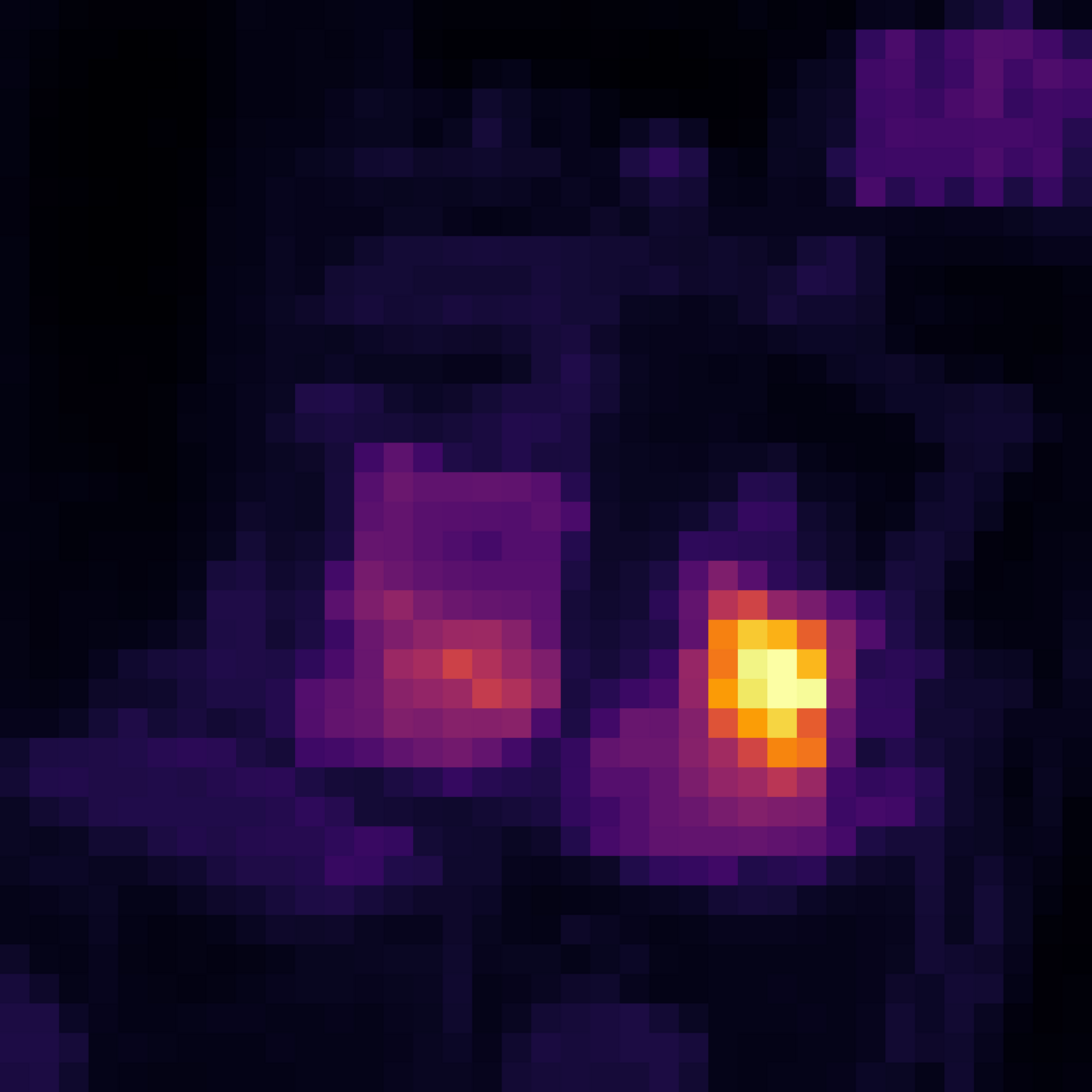}\end{minipage} & \begin{minipage}[t]{\linewidth}\centering \fixedimg{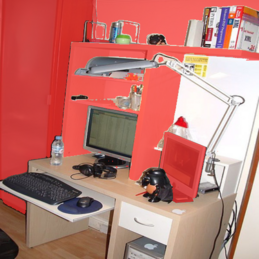}\\\end{minipage} & \begin{minipage}[t]{\linewidth}\centering \fixedimg{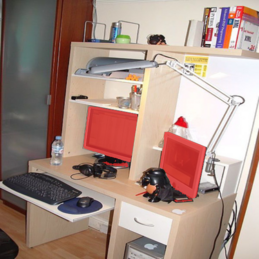}\end{minipage} & \begin{minipage}[t]{\linewidth}\centering \fixedimg{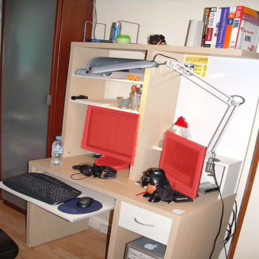}\end{minipage} \\
\centering tv & \centering Television & \multicolumn{2}{>{\centering\arraybackslash}p{4cm}}{\textit{an electronic device that receives television signals and displays them on a screen}} & \centering 7.59 mIoU & \centering 86.63 mIoU &  \\

\begin{minipage}[t]{\linewidth}\centering \fixedimg{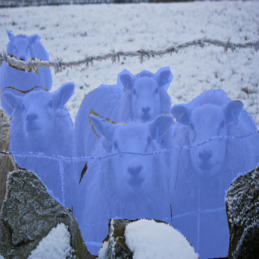}\\\end{minipage} & \begin{minipage}[t]{\linewidth}\centering \fixedimg{}\\\end{minipage} & \begin{minipage}[t]{\linewidth}\centering \fixedimg{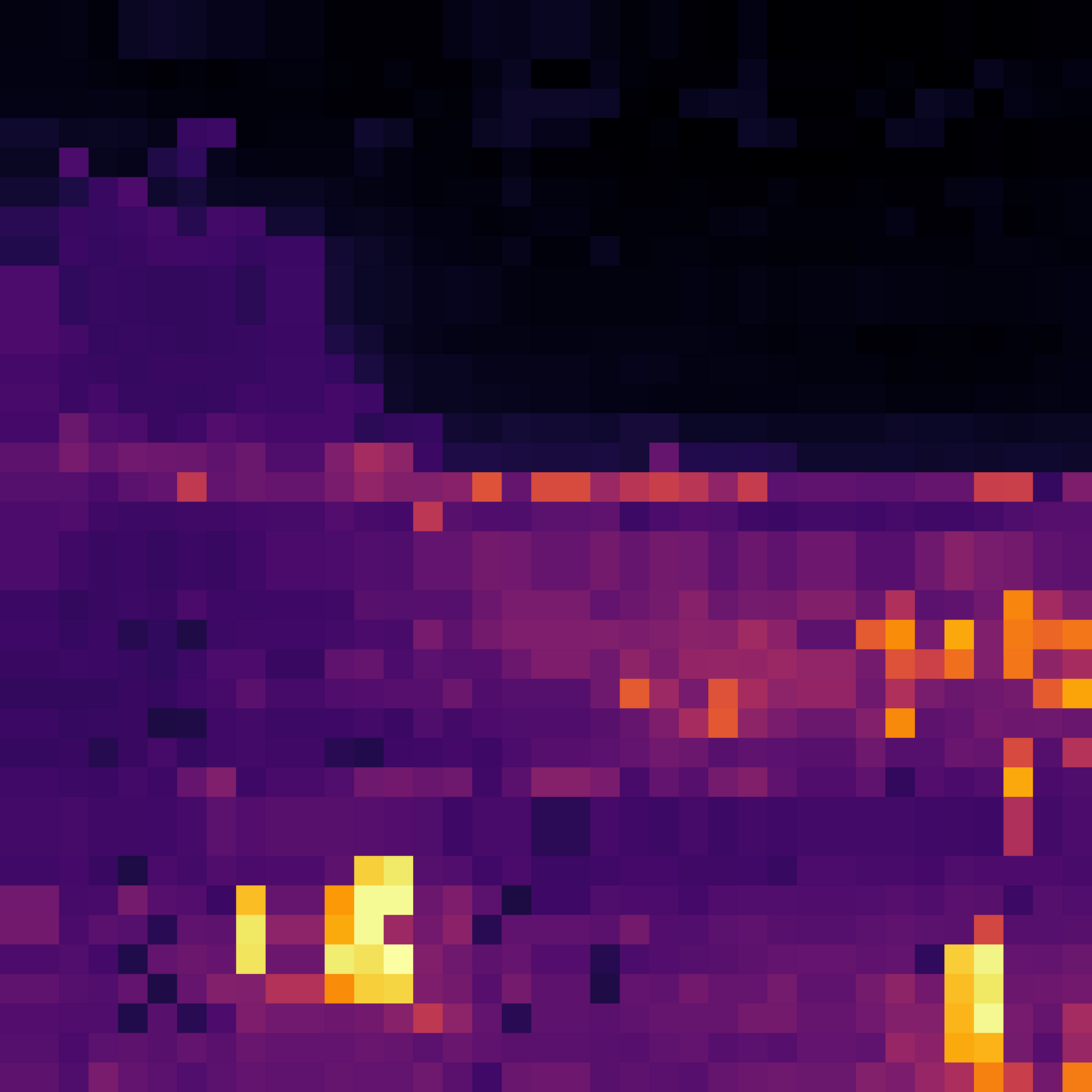}\end{minipage} & \begin{minipage}[t]{\linewidth}\centering \fixedimg{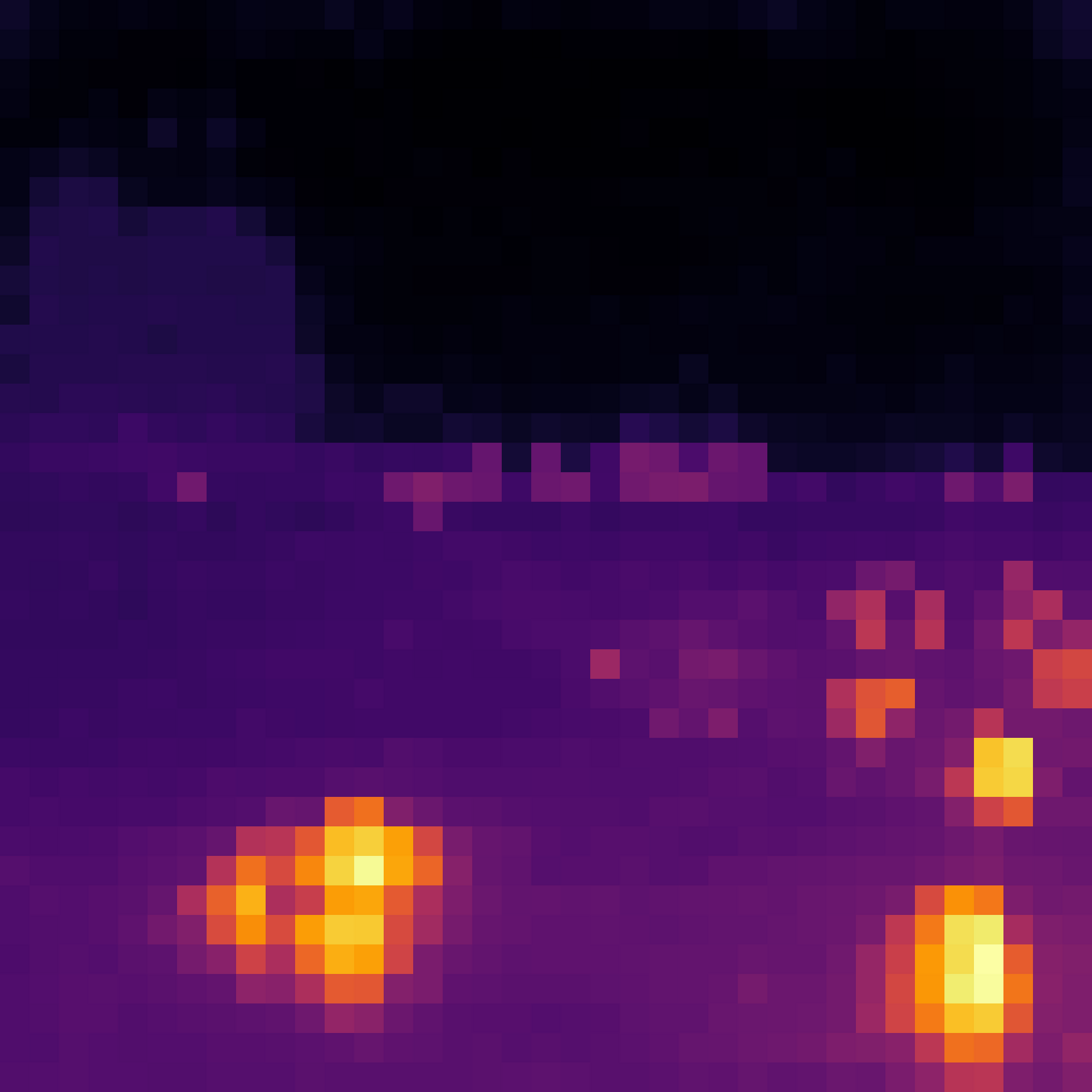}\end{minipage} & \begin{minipage}[t]{\linewidth}\centering \fixedimg{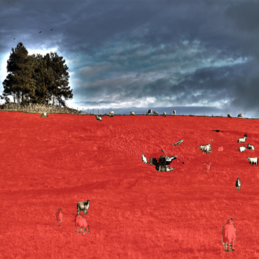}\\\end{minipage} & \begin{minipage}[t]{\linewidth}\centering \fixedimg{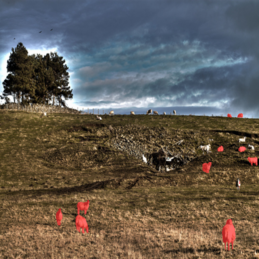}\end{minipage} & \begin{minipage}[t]{\linewidth}\centering \fixedimg{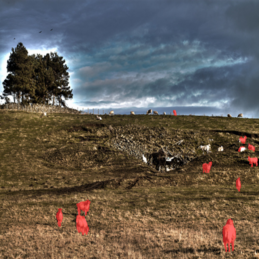}\end{minipage} \\
\centering sheep & \centering Sheep & \multicolumn{2}{>{\centering\arraybackslash}p{4cm}}{\textit{woolly usually horned ruminant mammal related to the goat}} & \centering 1.67 mIoU & \centering 64.91 mIoU &  \\

\begin{minipage}[t]{\linewidth}\centering \fixedimg{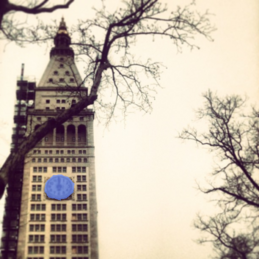}\\\end{minipage} & \begin{minipage}[t]{\linewidth}\centering \fixedimg{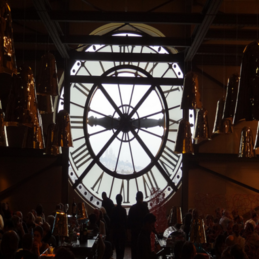}\\\end{minipage} & \begin{minipage}[t]{\linewidth}\centering \fixedimg{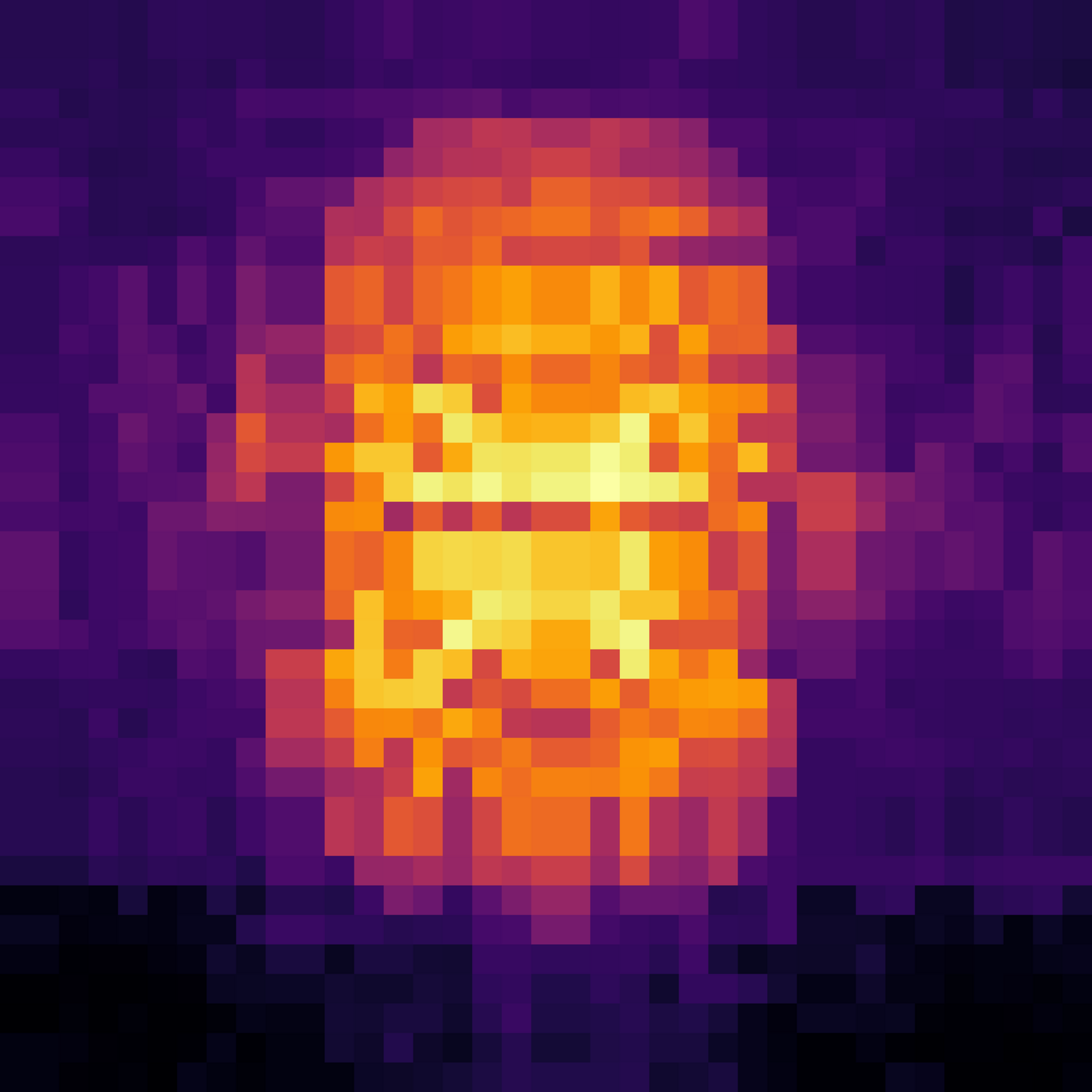}\end{minipage} & \begin{minipage}[t]{\linewidth}\centering \fixedimg{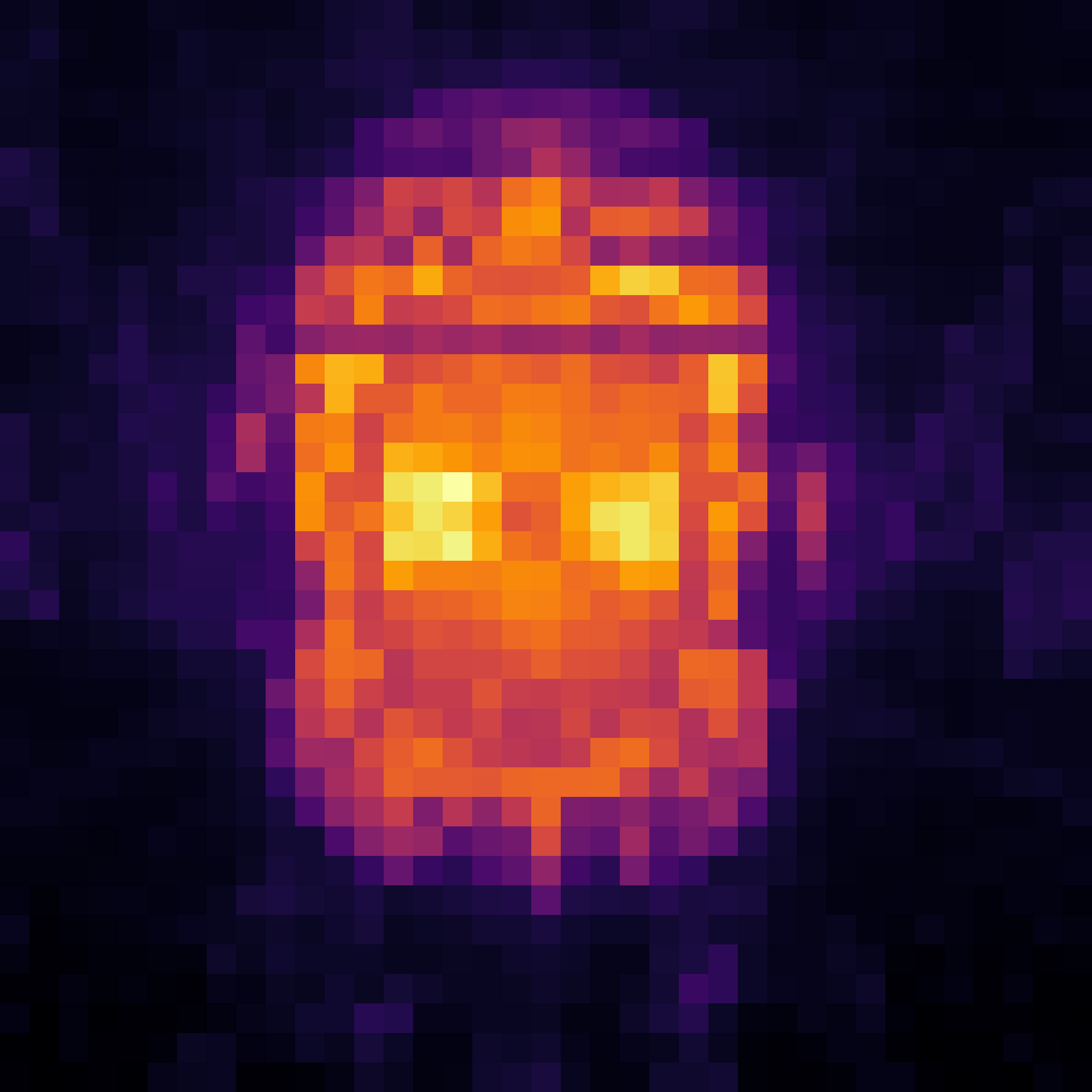}\end{minipage} & \begin{minipage}[t]{\linewidth}\centering \fixedimg{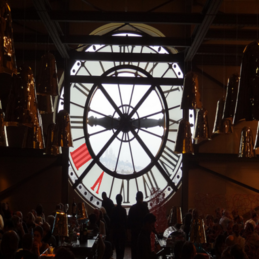}\\\end{minipage} & \begin{minipage}[t]{\linewidth}\centering \fixedimg{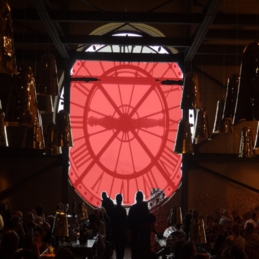}\end{minipage} & \begin{minipage}[t]{\linewidth}\centering \fixedimg{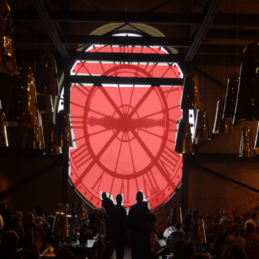}\end{minipage} \\
\centering clock & \centering Clock & \multicolumn{2}{>{\centering\arraybackslash}p{4cm}}{\textit{a timepiece that shows the time of day}} & \centering 2.94 mIoU & \centering 85.34 mIoU &  \\

\begin{minipage}[t]{\linewidth}\centering \fixedimg{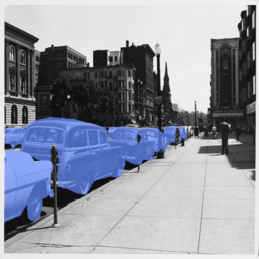}\\\end{minipage} & \begin{minipage}[t]{\linewidth}\centering \fixedimg{}\\\end{minipage} & \begin{minipage}[t]{\linewidth}\centering \fixedimg{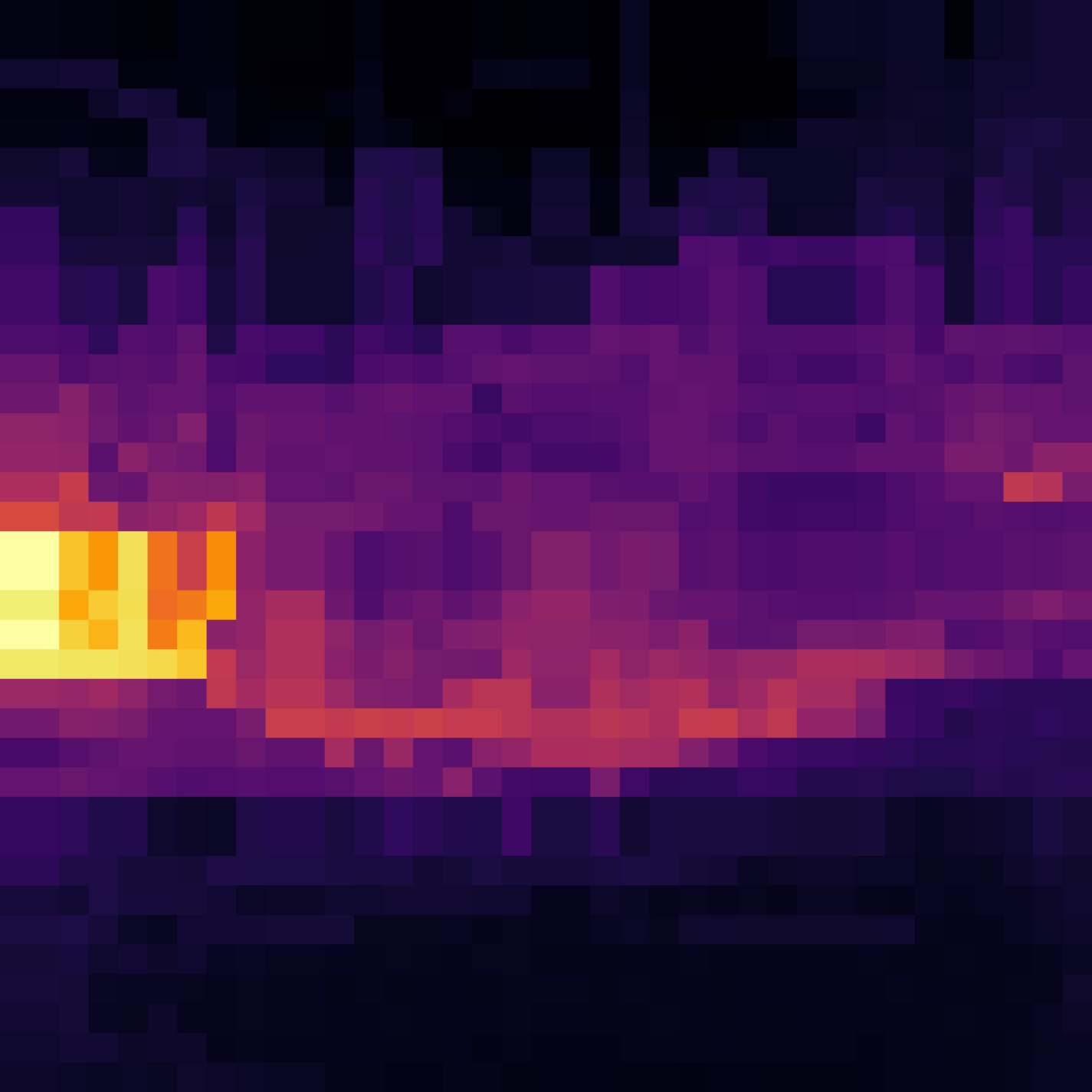}\end{minipage} & \begin{minipage}[t]{\linewidth}\centering \fixedimg{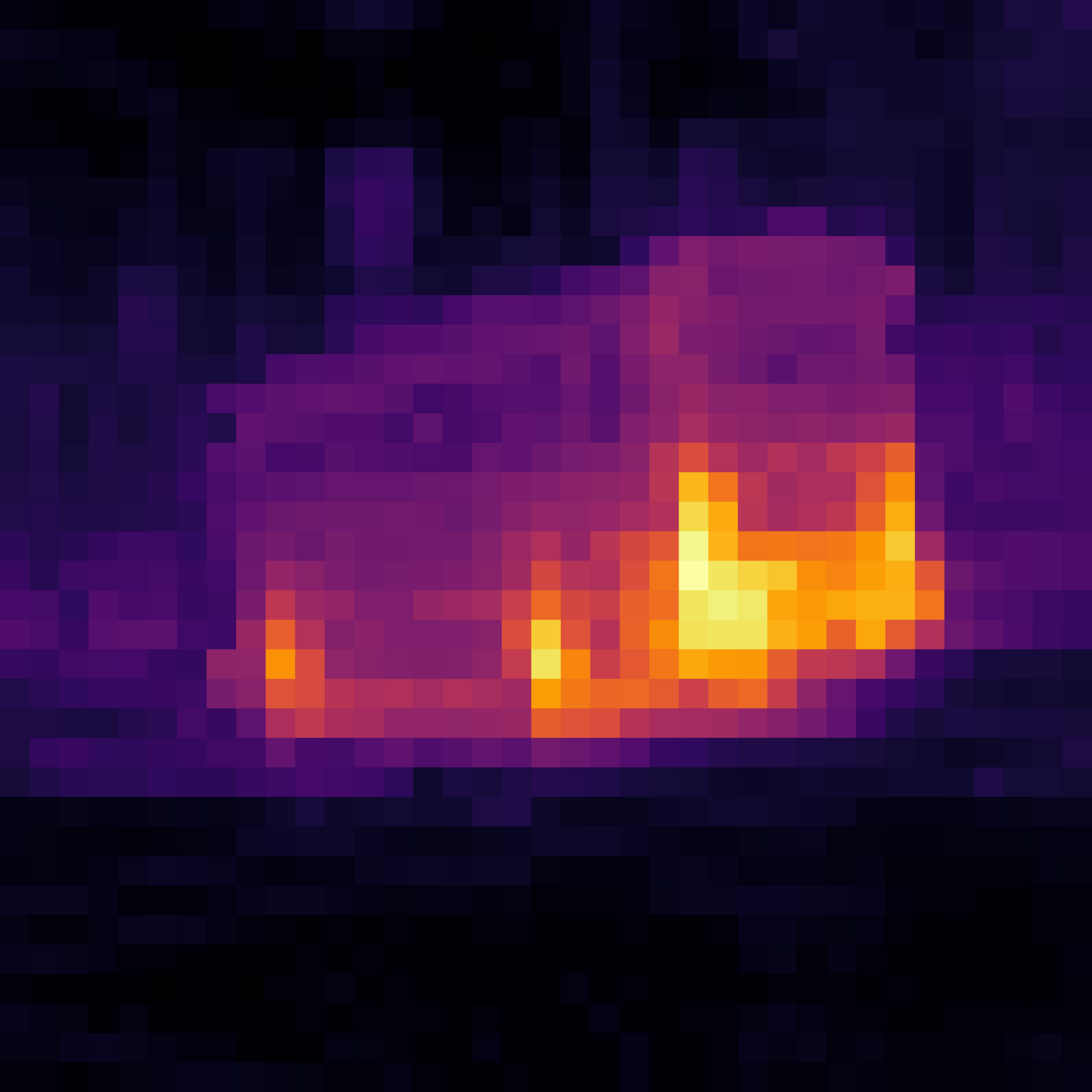}\end{minipage} & \begin{minipage}[t]{\linewidth}\centering \fixedimg{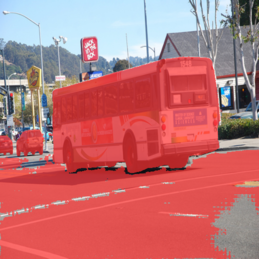}\\\end{minipage} & \begin{minipage}[t]{\linewidth}\centering \fixedimg{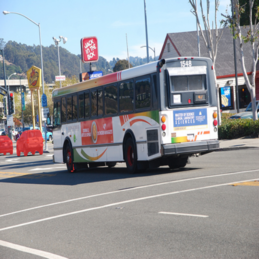}\end{minipage} & \begin{minipage}[t]{\linewidth}\centering \fixedimg{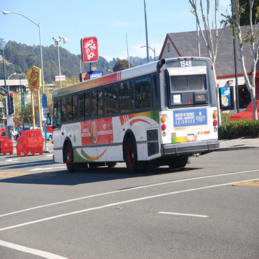}\end{minipage} \\
\centering car & \centering Car & \multicolumn{2}{>{\centering\arraybackslash}p{4cm}}{\textit{*Description not found*}} & \centering 1.96 mIoU & \centering 65.84 mIoU &  \\

\begin{minipage}[t]{\linewidth}\centering \fixedimg{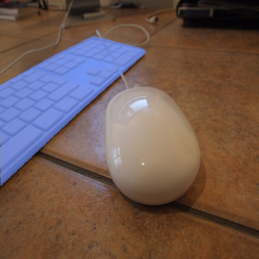}\\\end{minipage} & \begin{minipage}[t]{\linewidth}\centering \fixedimg{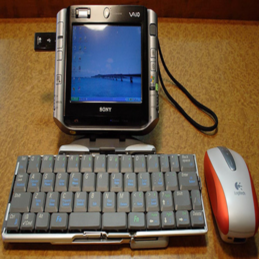}\\\end{minipage} & \begin{minipage}[t]{\linewidth}\centering \fixedimg{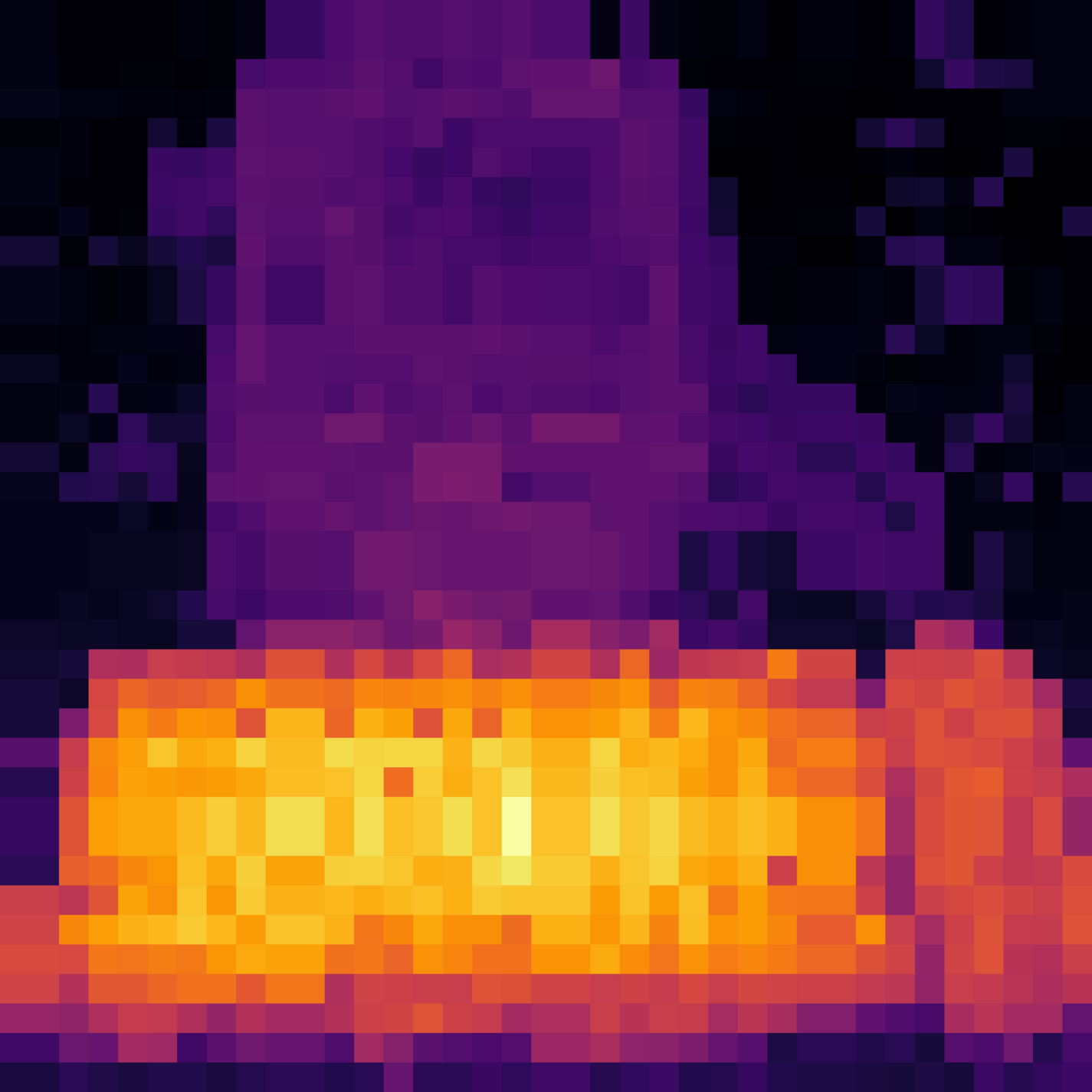}\end{minipage} & \begin{minipage}[t]{\linewidth}\centering \fixedimg{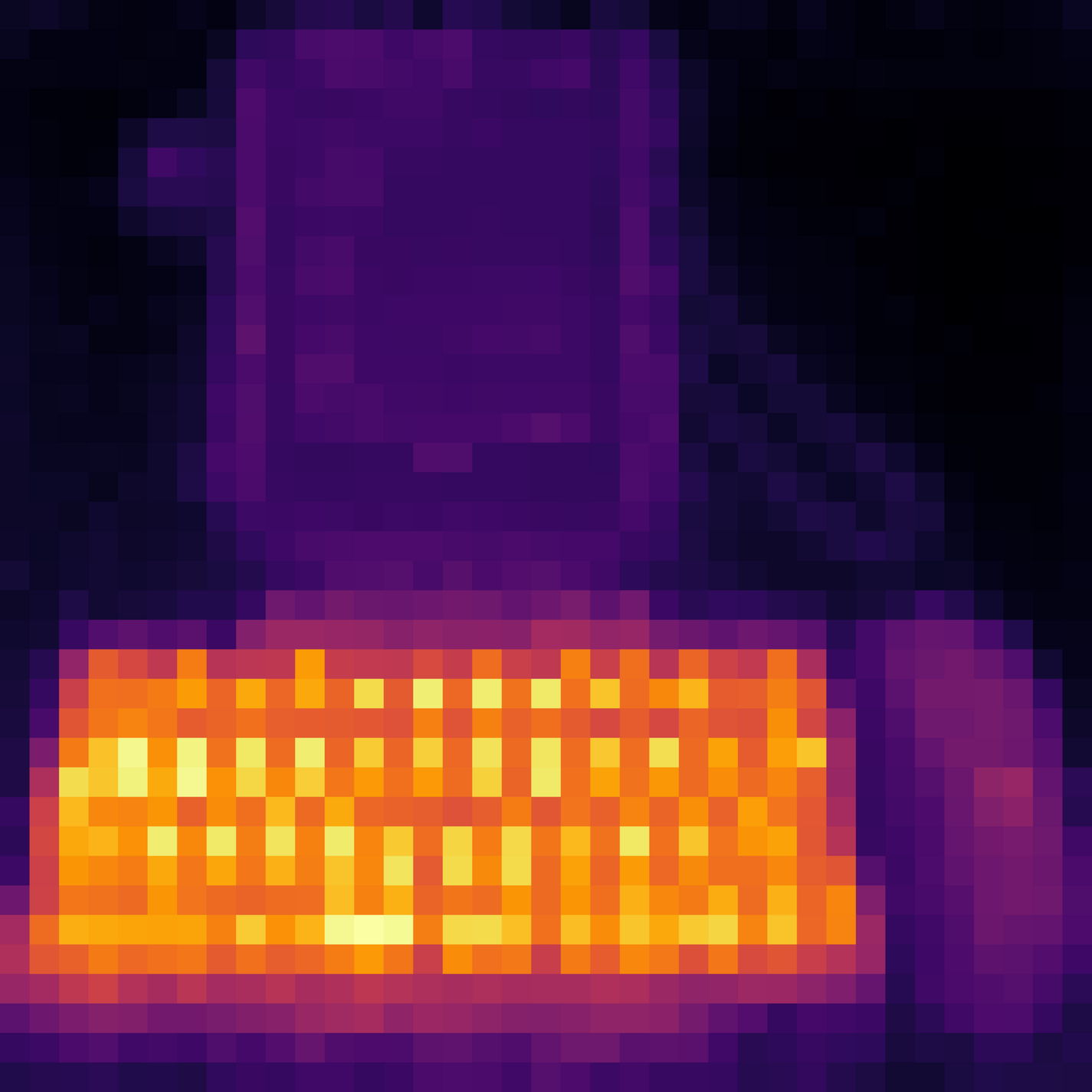}\end{minipage} & \begin{minipage}[t]{\linewidth}\centering \fixedimg{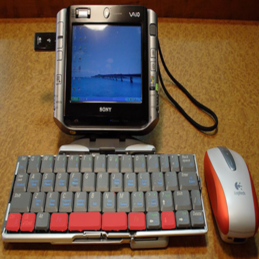}\\\end{minipage} & \begin{minipage}[t]{\linewidth}\centering \fixedimg{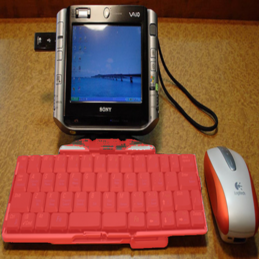}\end{minipage} & \begin{minipage}[t]{\linewidth}\centering \fixedimg{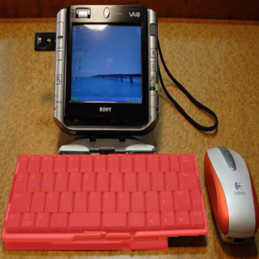}\end{minipage} \\
\centering keyboard & \centering Keyboard & \multicolumn{2}{>{\centering\arraybackslash}p{4cm}}{\textit{*Description not found*}} & \centering 12.76 mIoU & \centering 89.25 mIoU &  \\

\begin{minipage}[t]{\linewidth}\centering \fixedimg{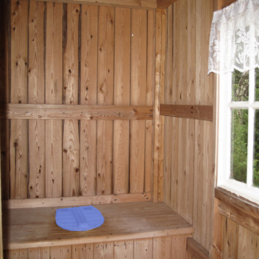}\\\end{minipage} & \begin{minipage}[t]{\linewidth}\centering \fixedimg{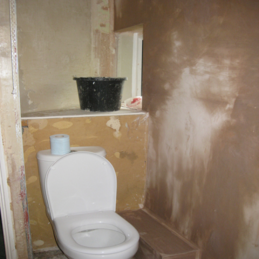}\\\end{minipage} & \begin{minipage}[t]{\linewidth}\centering \fixedimg{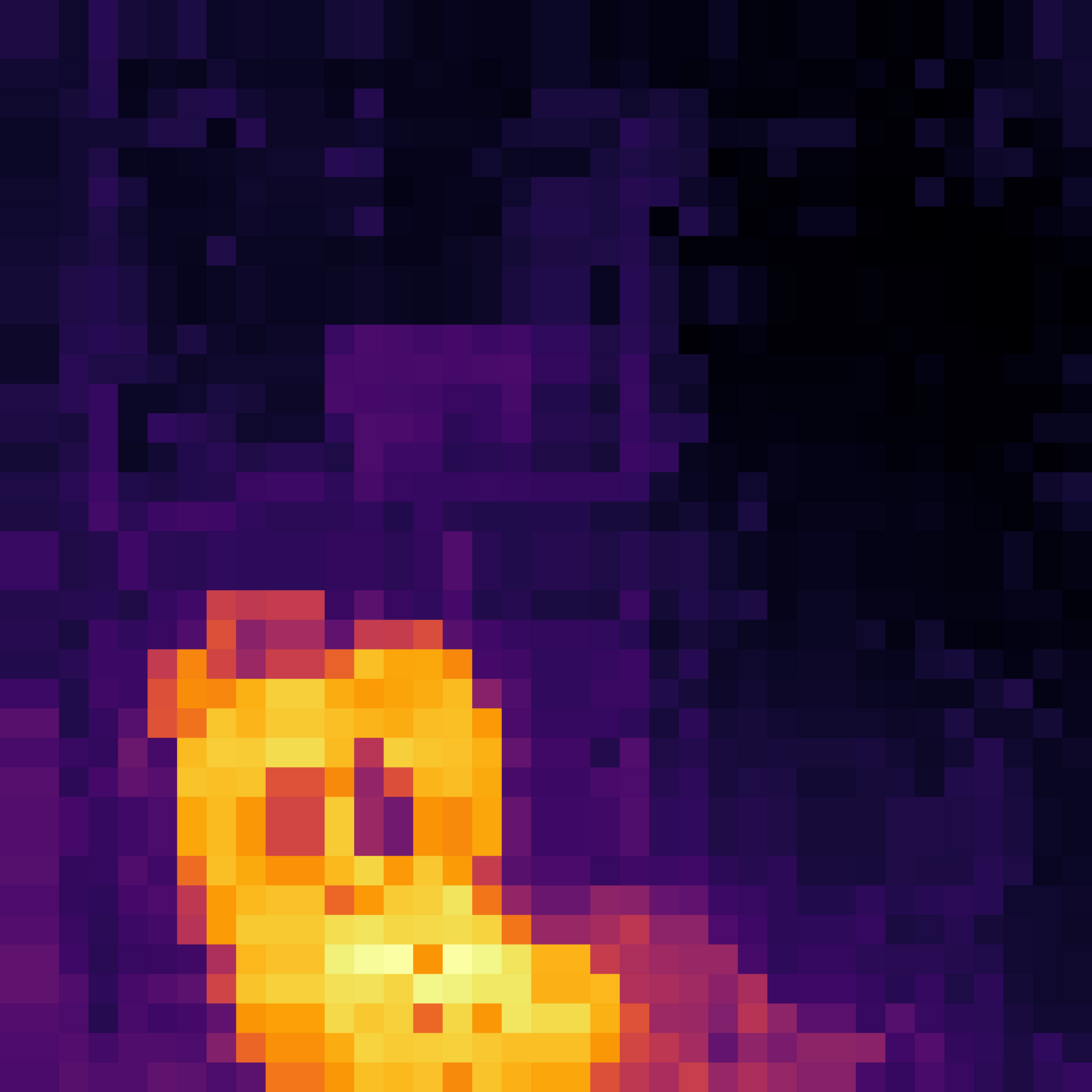}\end{minipage} & \begin{minipage}[t]{\linewidth}\centering \fixedimg{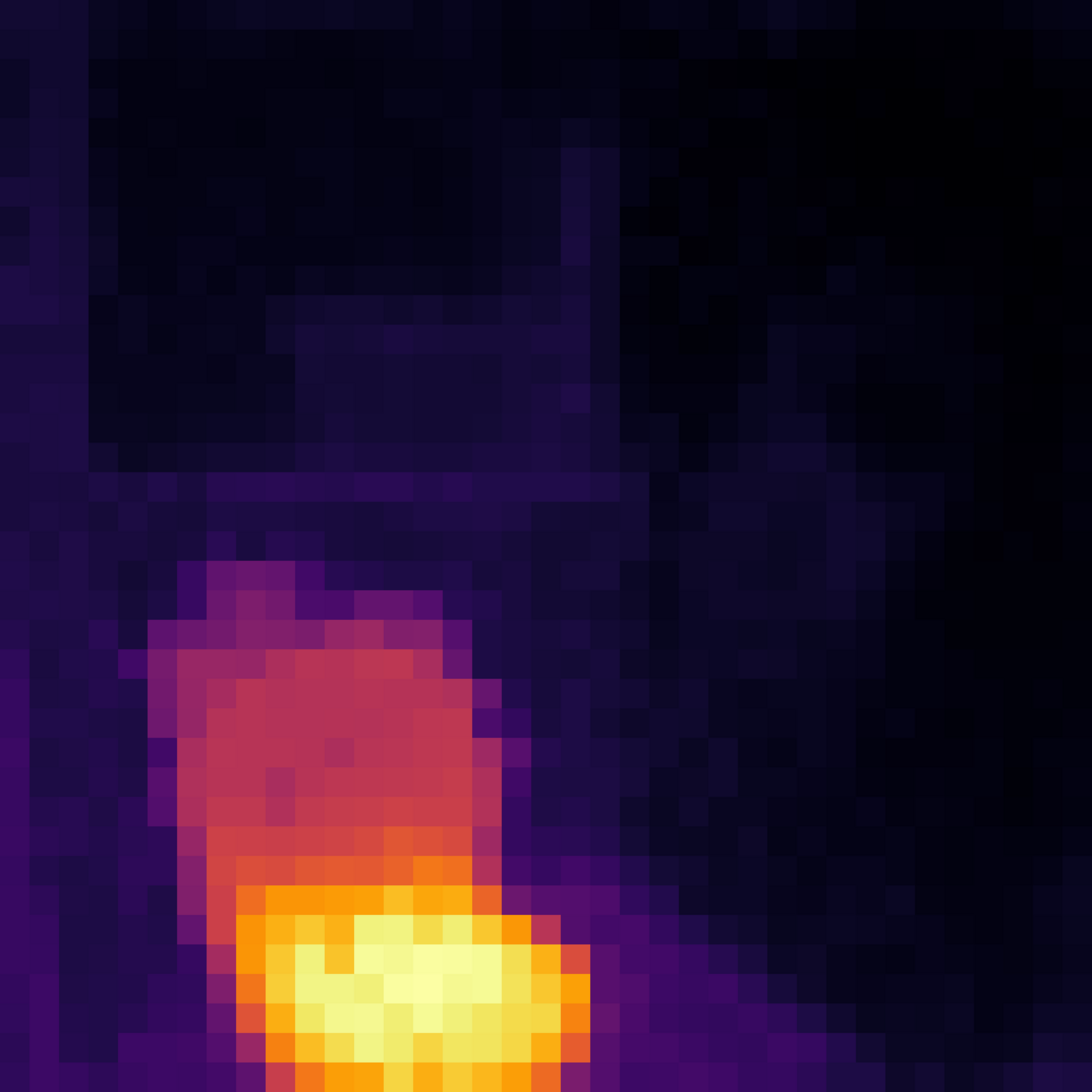}\end{minipage} & \begin{minipage}[t]{\linewidth}\centering \fixedimg{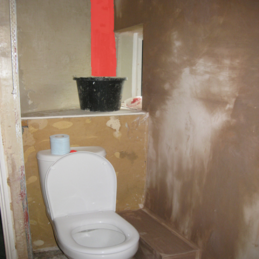}\\\end{minipage} & \begin{minipage}[t]{\linewidth}\centering \fixedimg{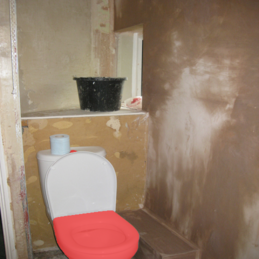}\end{minipage} & \begin{minipage}[t]{\linewidth}\centering \fixedimg{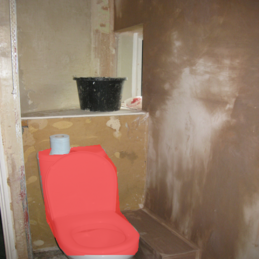}\end{minipage} \\
\centering toilet & \centering Chair & \multicolumn{2}{>{\centering\arraybackslash}p{4cm}}{\textit{a seat for one person, with a support for the back}} & \centering 0.13 mIoU & \centering 38.24 mIoU &  \\

\begin{minipage}[t]{\linewidth}\centering \fixedimg{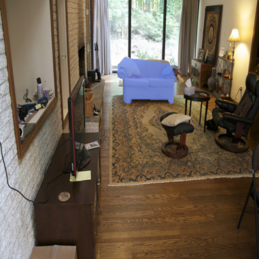}\\\end{minipage} & \begin{minipage}[t]{\linewidth}\centering \fixedimg{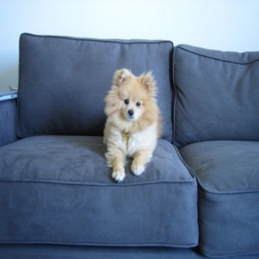}\\\end{minipage} & \begin{minipage}[t]{\linewidth}\centering \fixedimg{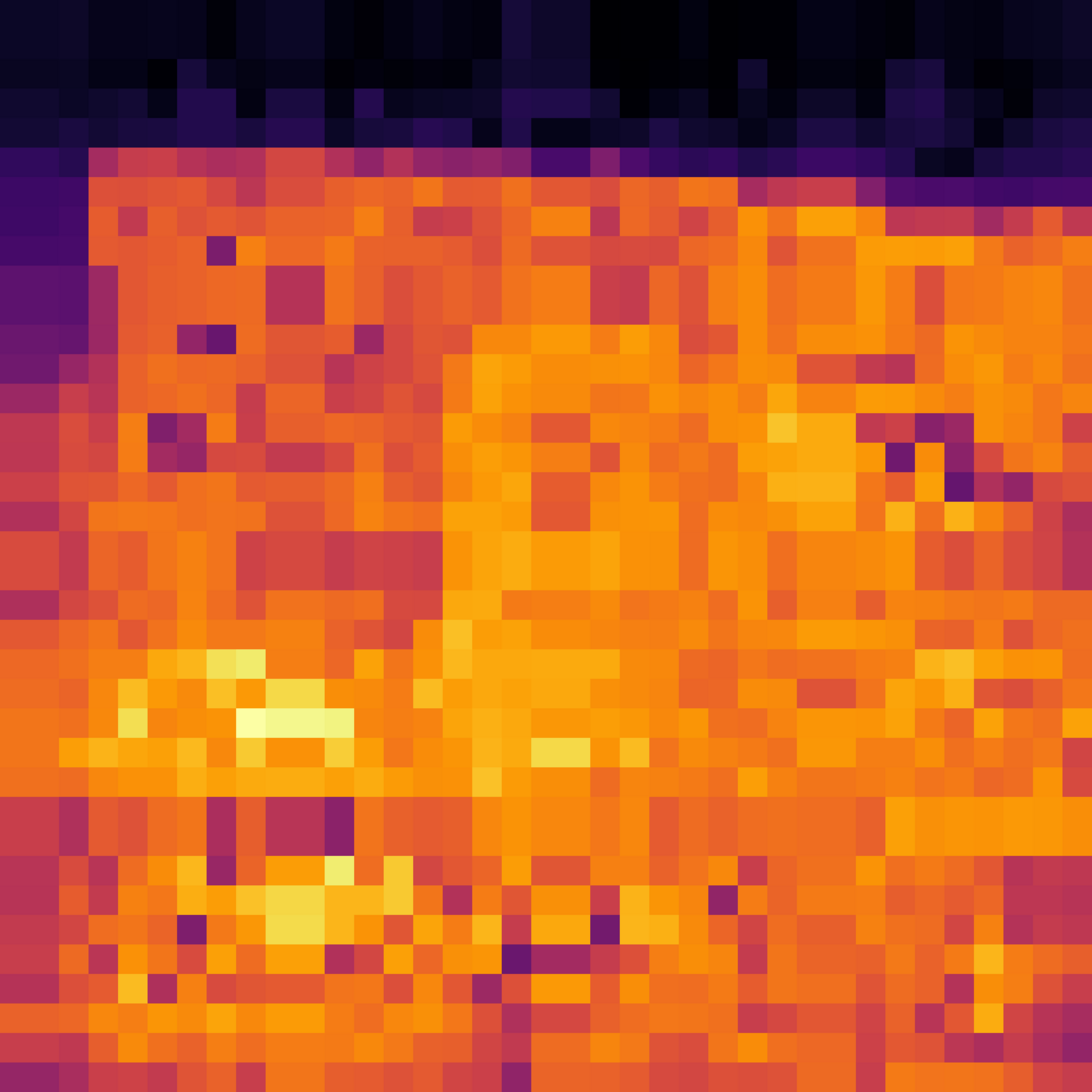}\end{minipage} & \begin{minipage}[t]{\linewidth}\centering \fixedimg{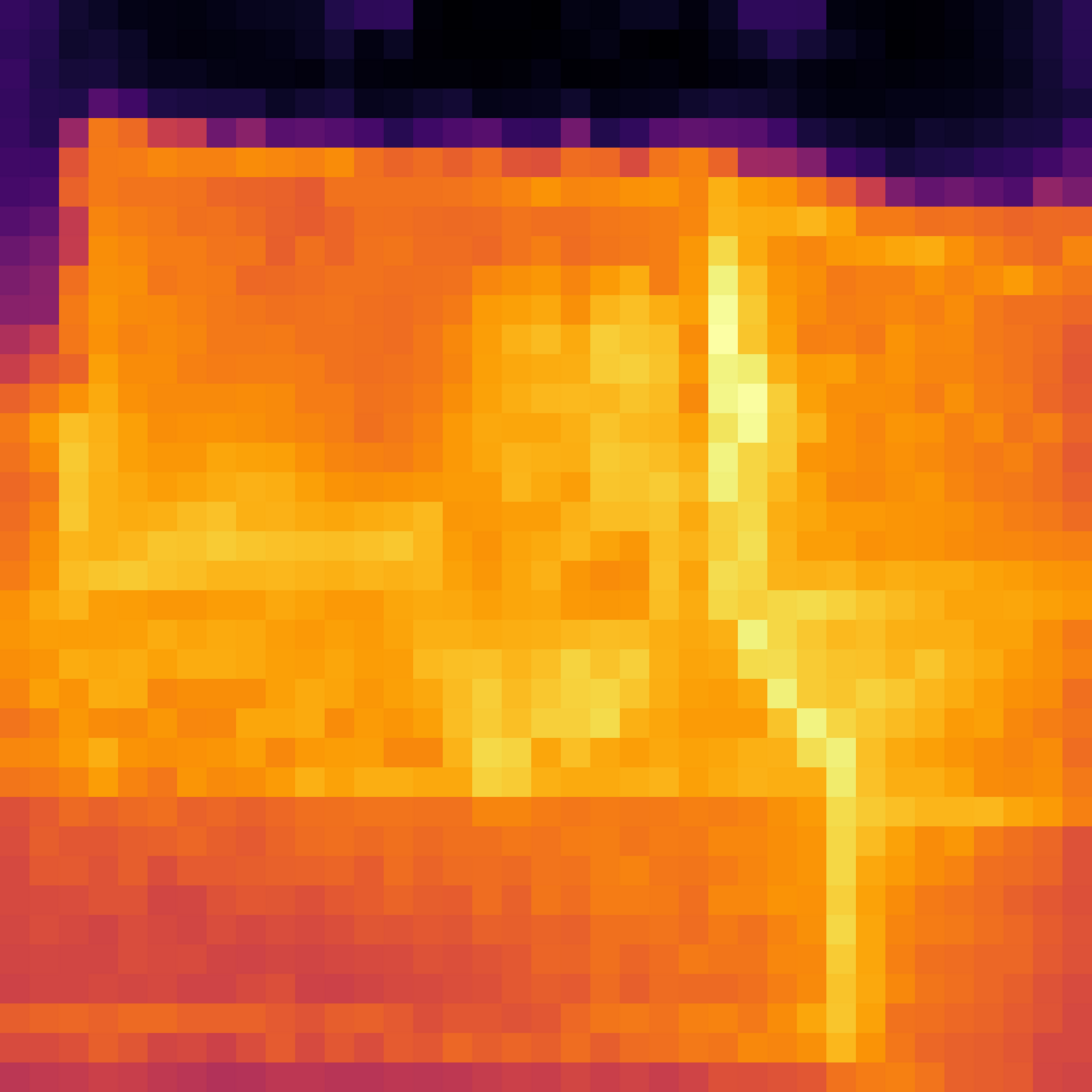}\end{minipage} & \begin{minipage}[t]{\linewidth}\centering \fixedimg{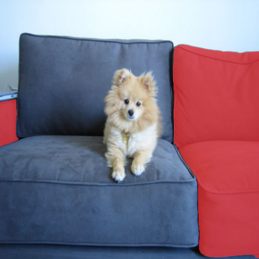}\\\end{minipage} & \begin{minipage}[t]{\linewidth}\centering \fixedimg{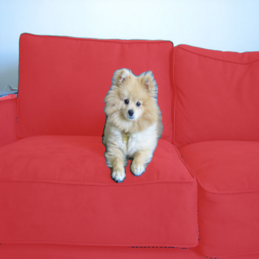}\end{minipage} & \begin{minipage}[t]{\linewidth}\centering \fixedimg{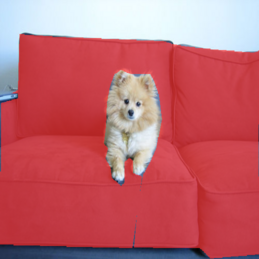}\end{minipage} \\
\centering couch & \centering Couch & \multicolumn{2}{>{\centering\arraybackslash}p{4cm}}{\textit{*Description not found*}} & \centering 33.31 mIoU & \centering 92.98 mIoU &  \\

\begin{minipage}[t]{\linewidth}\centering \fixedimg{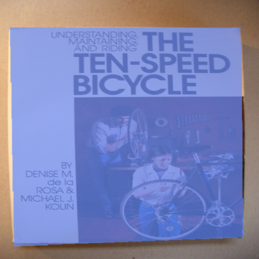}\\\end{minipage} & \begin{minipage}[t]{\linewidth}\centering \fixedimg{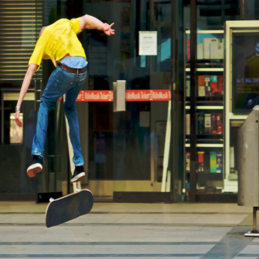}\\\end{minipage} & \begin{minipage}[t]{\linewidth}\centering \fixedimg{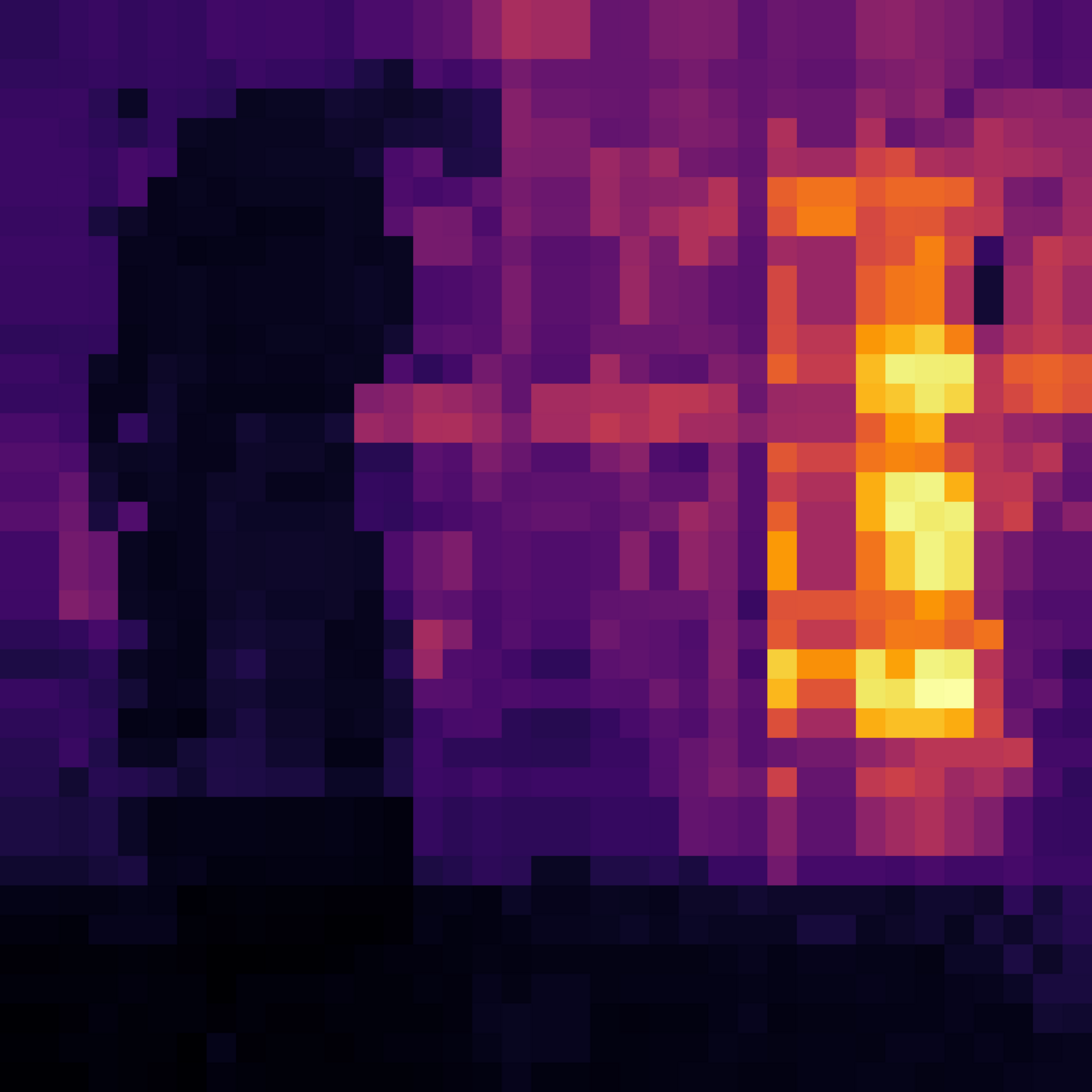}\end{minipage} & \begin{minipage}[t]{\linewidth}\centering \fixedimg{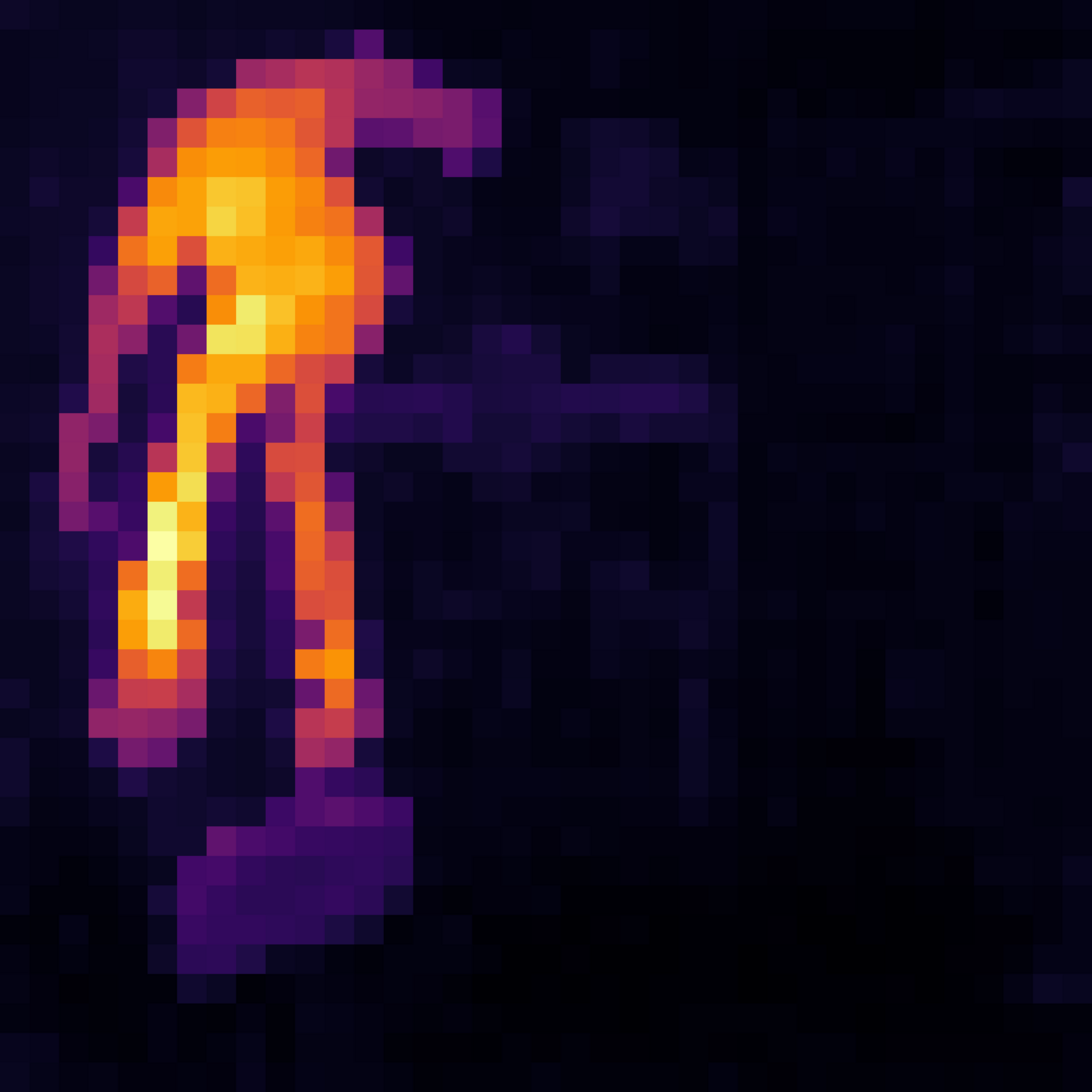}\end{minipage} & \begin{minipage}[t]{\linewidth}\centering \fixedimg{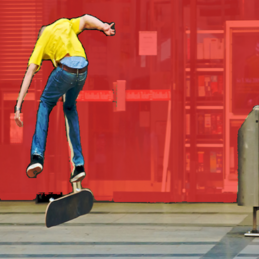}\\\end{minipage} & \begin{minipage}[t]{\linewidth}\centering \fixedimg{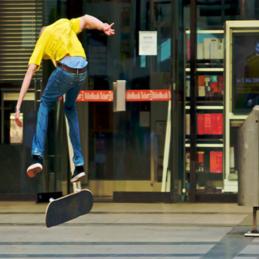}\end{minipage} & \begin{minipage}[t]{\linewidth}\centering \fixedimg{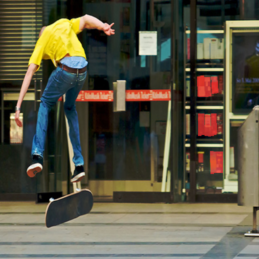}\end{minipage} \\
\centering book & \centering Book & \multicolumn{2}{>{\centering\arraybackslash}p{4cm}}{\textit{a written work or composition that has been published (printed on pages bound together)}} & \centering 2.47 mIoU & \centering 51.30 mIoU &  \\

\begin{minipage}[t]{\linewidth}\centering \fixedimg{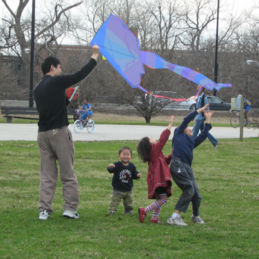}\\\end{minipage} & \begin{minipage}[t]{\linewidth}\centering \fixedimg{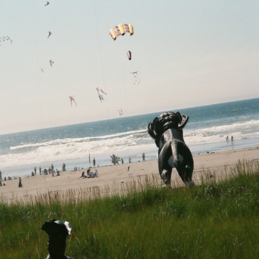}\\\end{minipage} & \begin{minipage}[t]{\linewidth}\centering \fixedimg{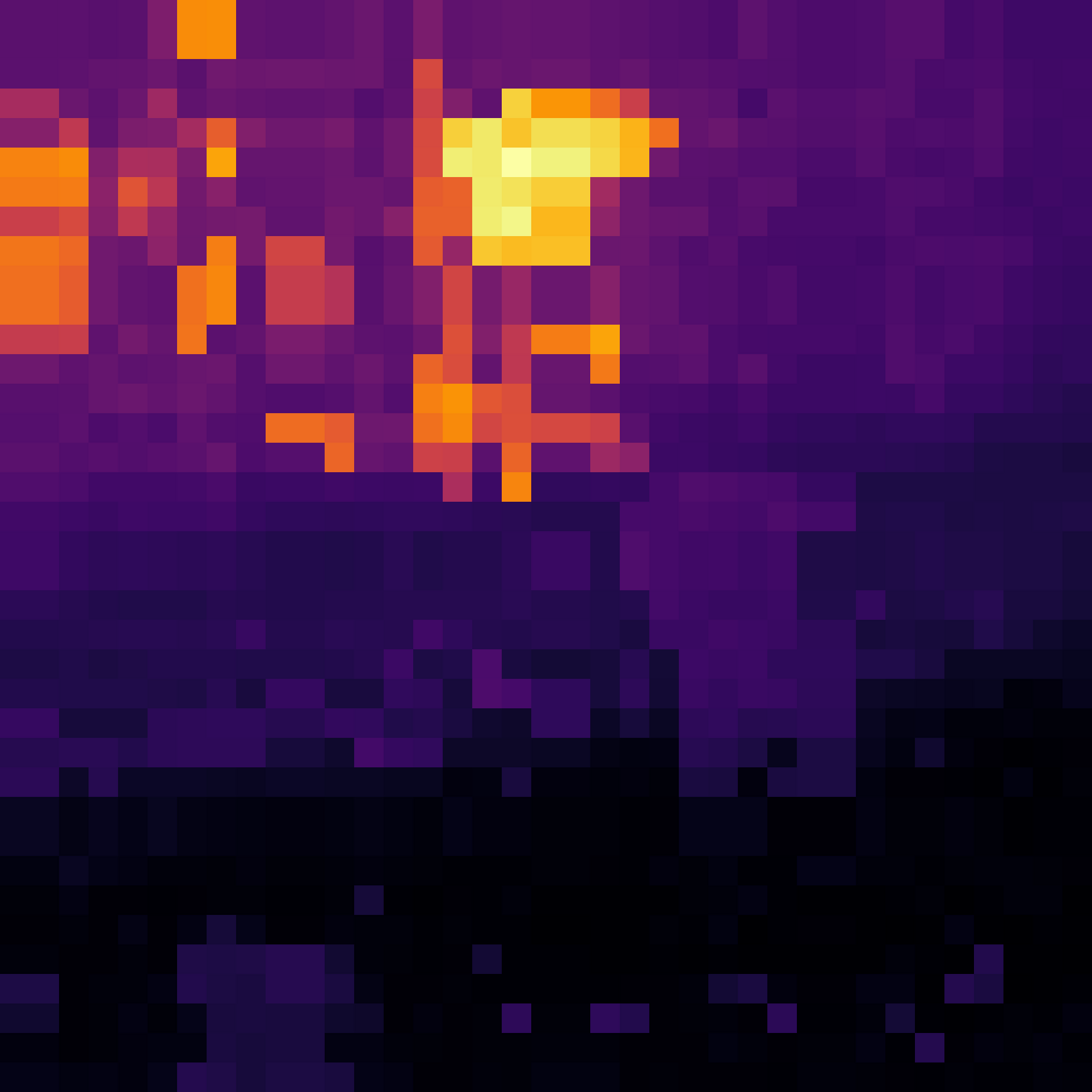}\end{minipage} & \begin{minipage}[t]{\linewidth}\centering \fixedimg{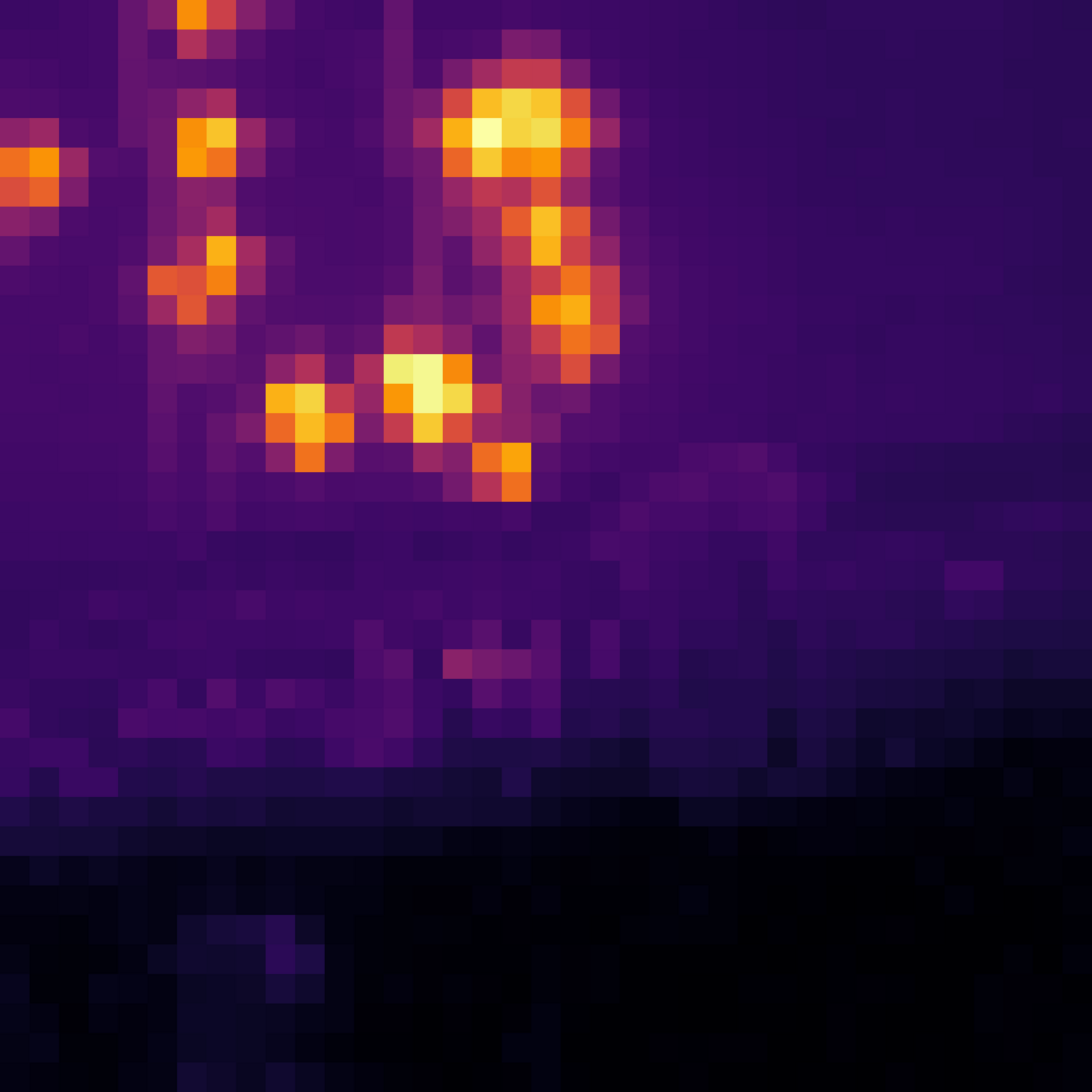}\end{minipage} & \begin{minipage}[t]{\linewidth}\centering \fixedimg{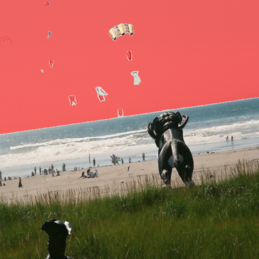}\\\end{minipage} & \begin{minipage}[t]{\linewidth}\centering \fixedimg{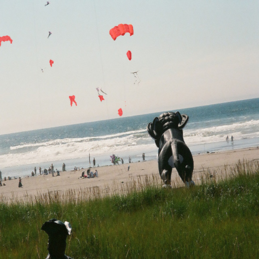}\end{minipage} & \begin{minipage}[t]{\linewidth}\centering \fixedimg{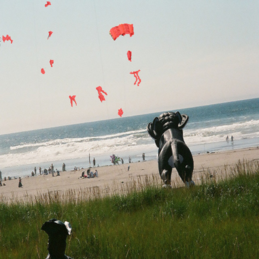}\end{minipage} \\
\centering kite & \centering Kite & \multicolumn{2}{>{\centering\arraybackslash}p{4cm}}{\textit{any of several small graceful hawks of the family Accipitridae having long pointed wings and feed...}} & \centering 0.74 mIoU & \centering 59.23 mIoU &  \\

\begin{minipage}[t]{\linewidth}\centering \fixedimg{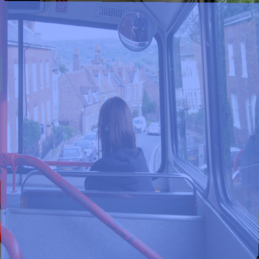}\\\end{minipage} & \begin{minipage}[t]{\linewidth}\centering \fixedimg{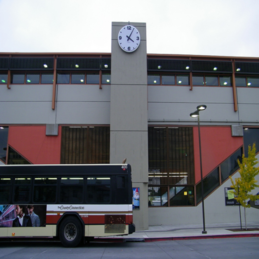}\\\end{minipage} & \begin{minipage}[t]{\linewidth}\centering \fixedimg{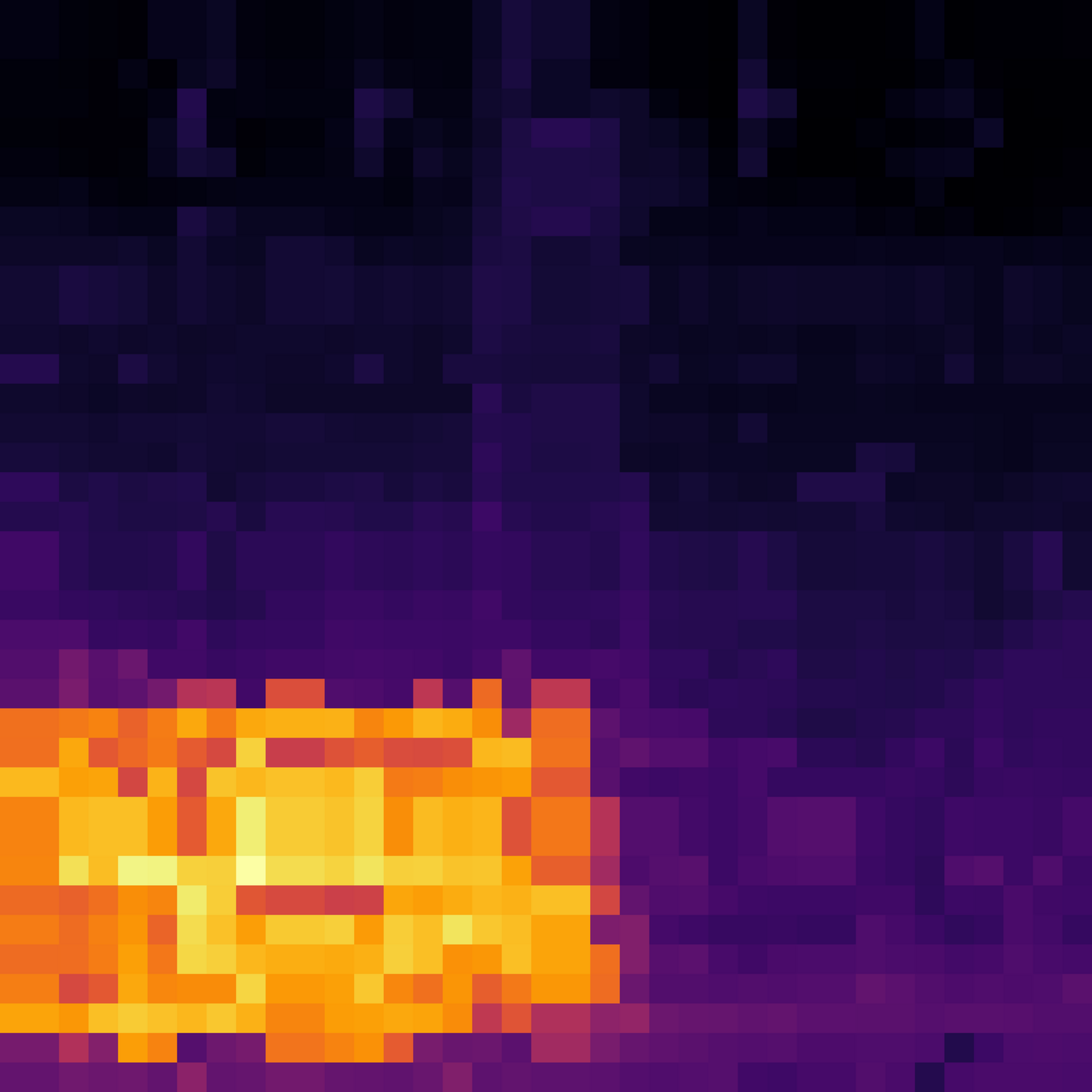}\end{minipage} & \begin{minipage}[t]{\linewidth}\centering \fixedimg{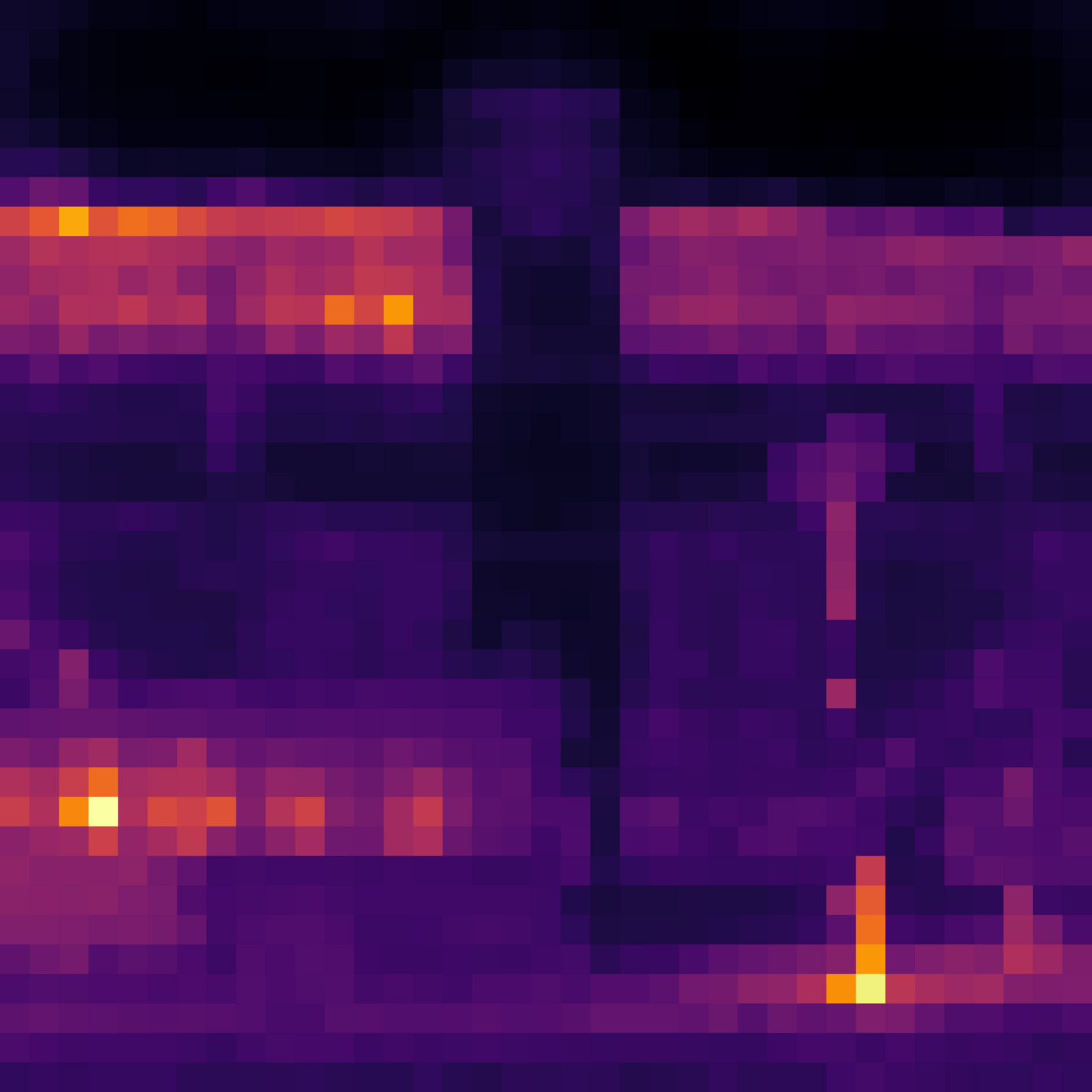}\end{minipage} & \begin{minipage}[t]{\linewidth}\centering \fixedimg{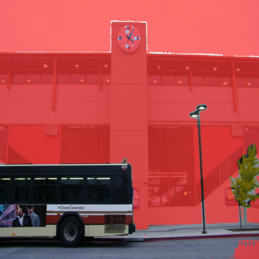}\\\end{minipage} & \begin{minipage}[t]{\linewidth}\centering \fixedimg{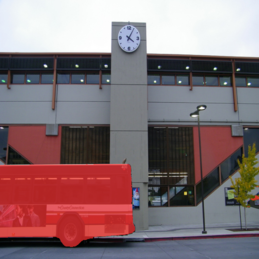}\end{minipage} & \begin{minipage}[t]{\linewidth}\centering \fixedimg{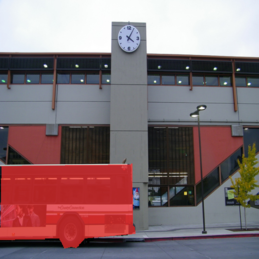}\end{minipage} \\
\centering bus & \centering Bus & \multicolumn{2}{>{\centering\arraybackslash}p{4cm}}{\textit{a vehicle carrying many passengers; used for public transport}} & \centering 0.02 mIoU & \centering 95.65 mIoU &  \\

\begin{minipage}[t]{\linewidth}\centering \fixedimg{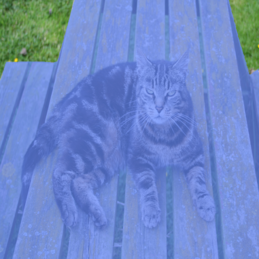}\\\end{minipage} & \begin{minipage}[t]{\linewidth}\centering \fixedimg{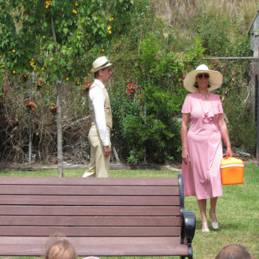}\\\end{minipage} & \begin{minipage}[t]{\linewidth}\centering \fixedimg{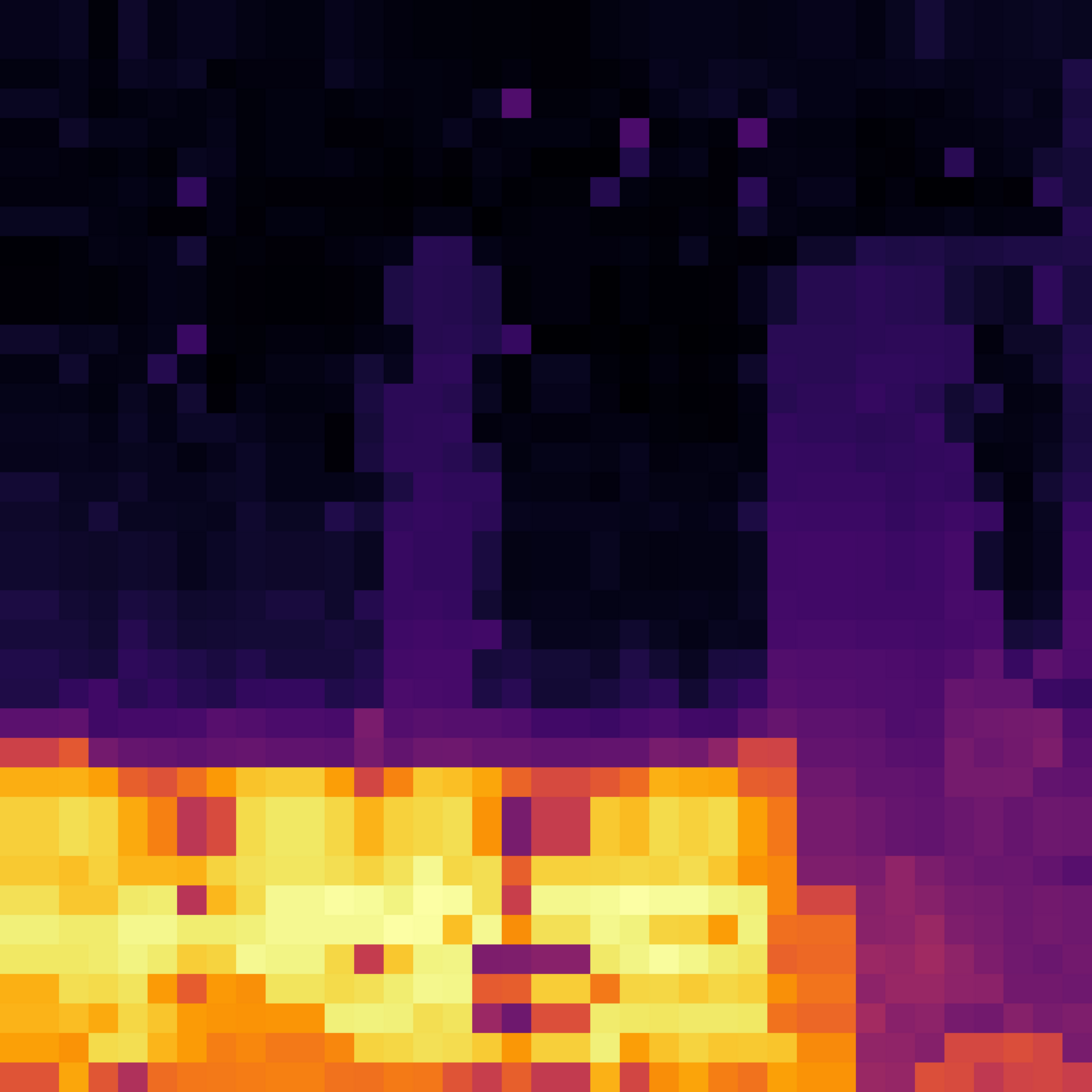}\end{minipage} & \begin{minipage}[t]{\linewidth}\centering \fixedimg{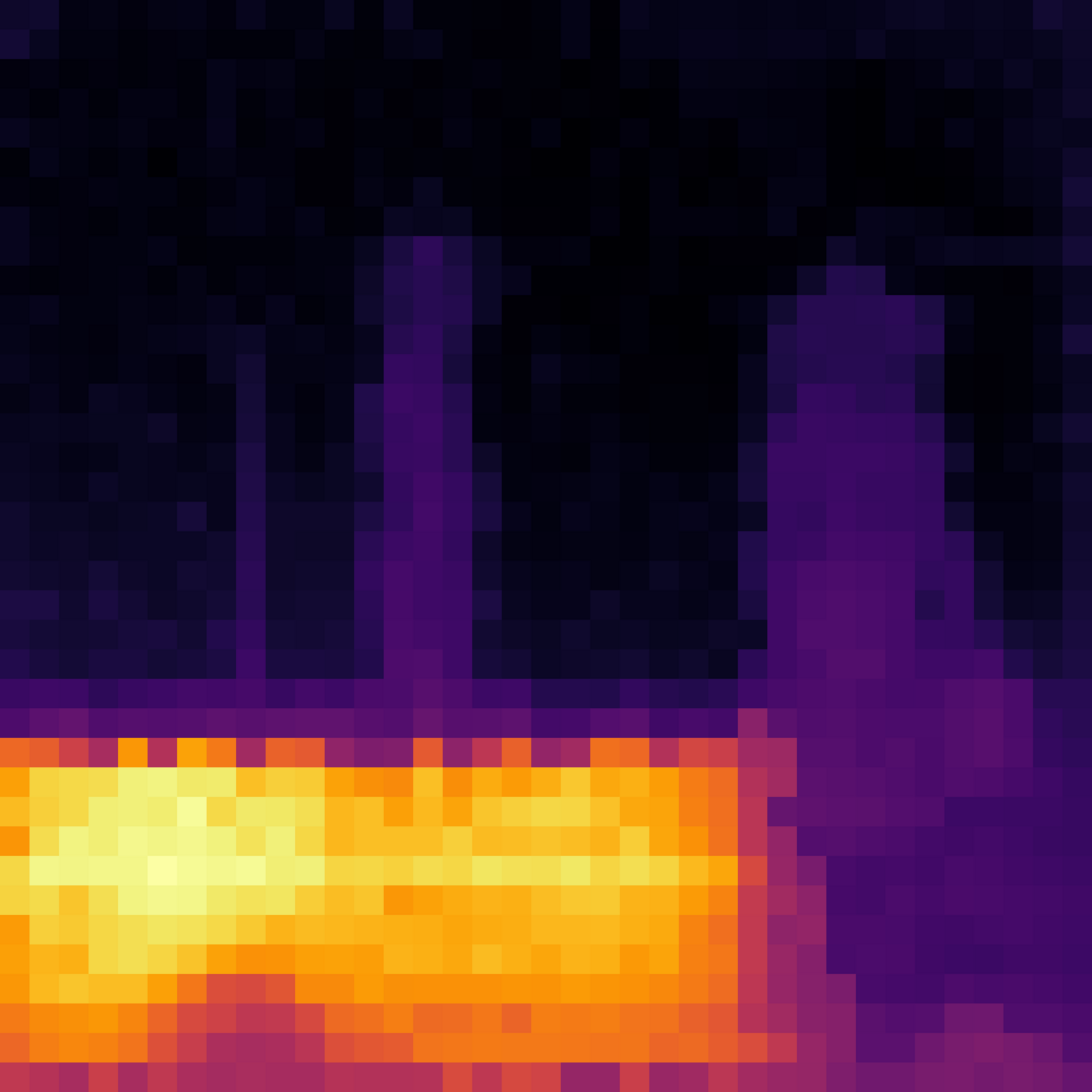}\end{minipage} & \begin{minipage}[t]{\linewidth}\centering \fixedimg{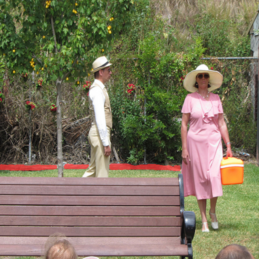}\\\end{minipage} & \begin{minipage}[t]{\linewidth}\centering \fixedimg{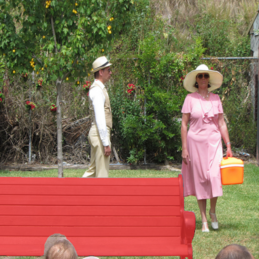}\end{minipage} & \begin{minipage}[t]{\linewidth}\centering \fixedimg{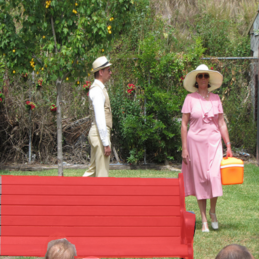}\end{minipage} \\
\centering bench & \centering Bench & \multicolumn{2}{>{\centering\arraybackslash}p{4cm}}{\textit{the magistrate or judge or judges sitting in court in judicial capacity to compose the court coll...}} & \centering 0.00 mIoU & \centering 94.98 mIoU &  \\

\begin{minipage}[t]{\linewidth}\centering \fixedimg{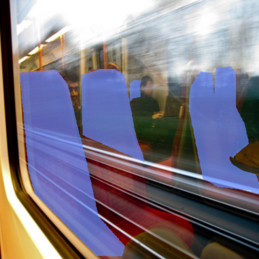}\\\end{minipage} & \begin{minipage}[t]{\linewidth}\centering \fixedimg{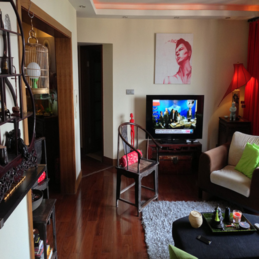}\\\end{minipage} & \begin{minipage}[t]{\linewidth}\centering \fixedimg{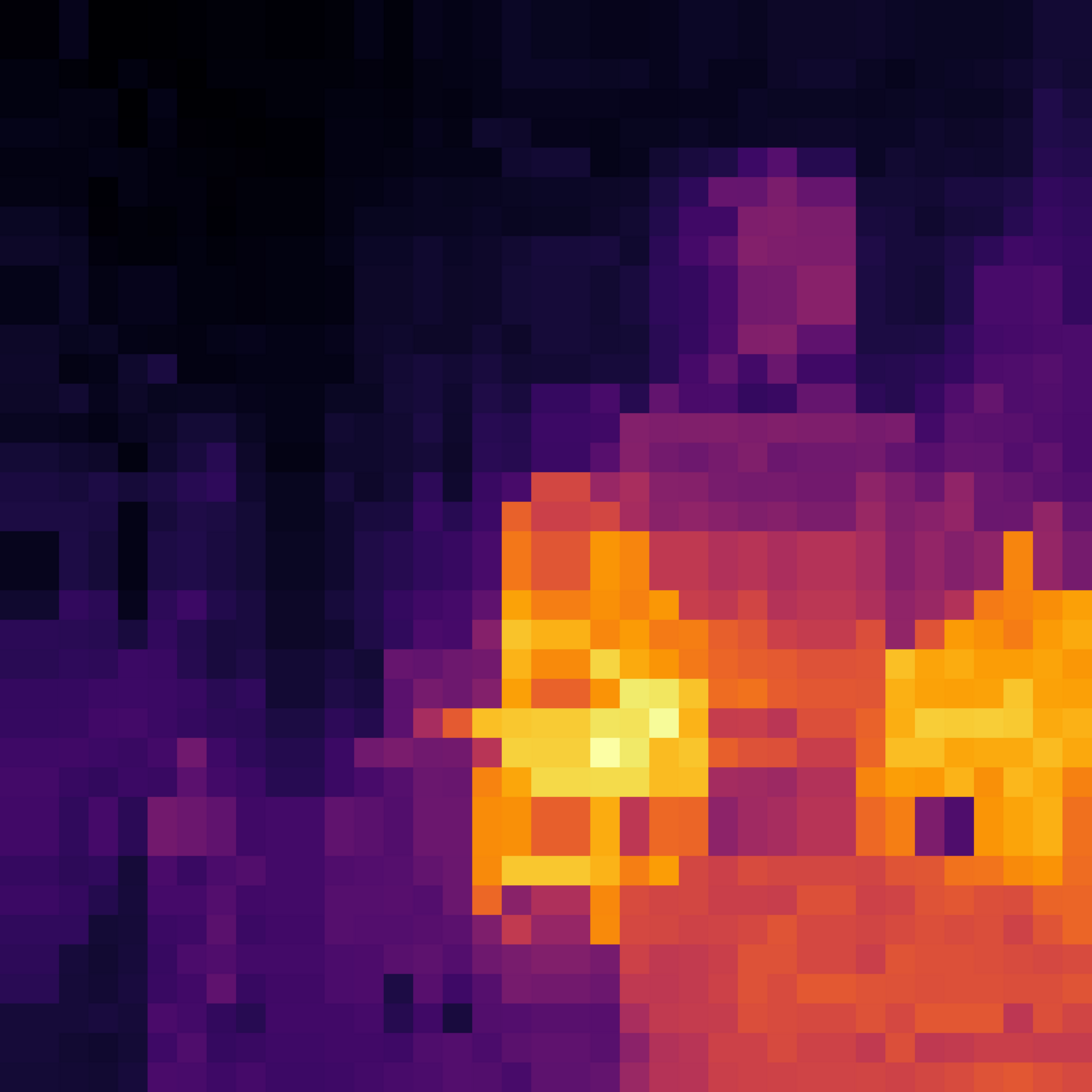}\end{minipage} & \begin{minipage}[t]{\linewidth}\centering \fixedimg{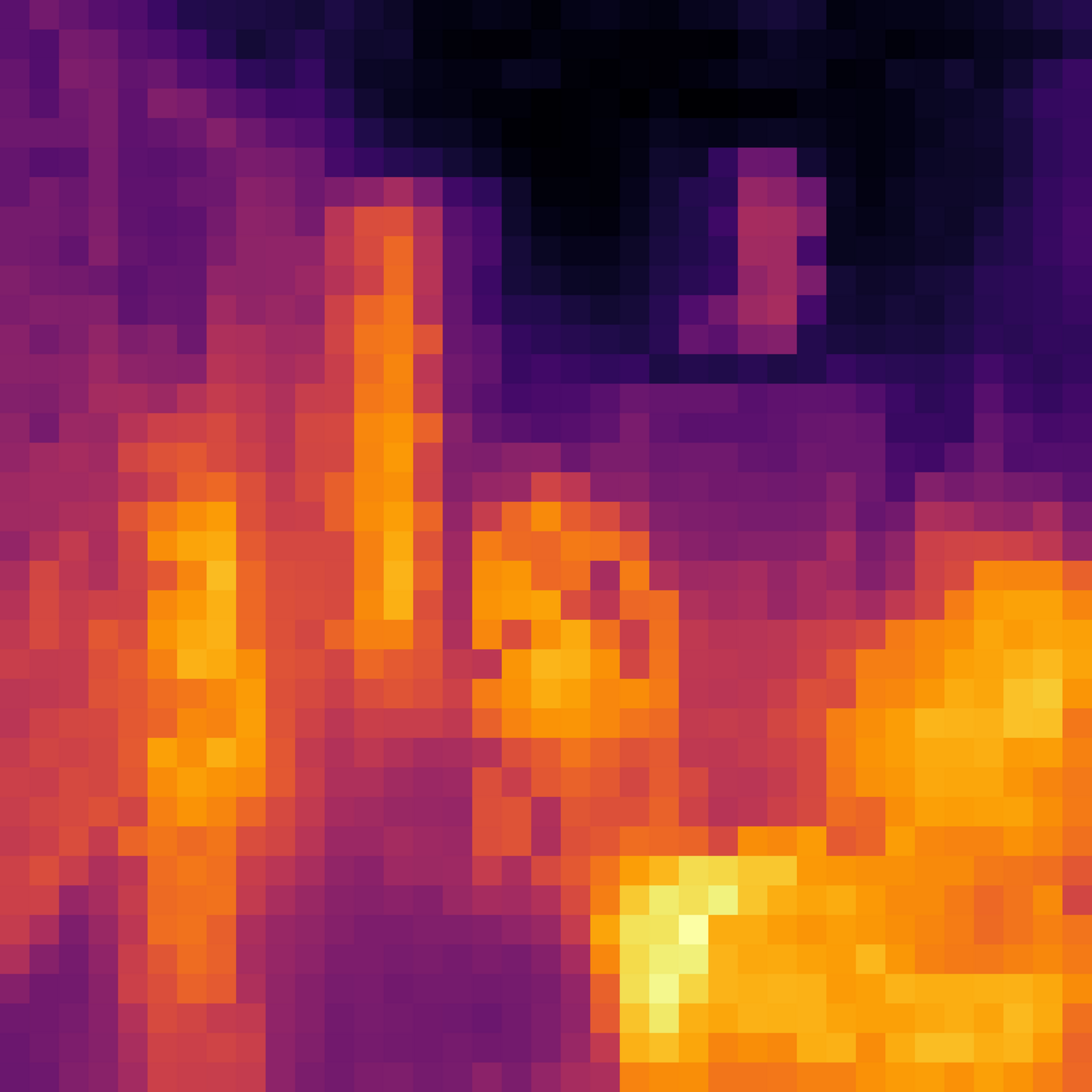}\end{minipage} & \begin{minipage}[t]{\linewidth}\centering \fixedimg{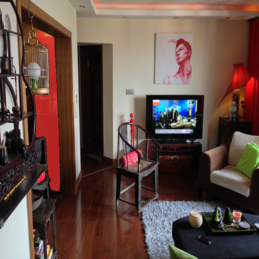}\\\end{minipage} & \begin{minipage}[t]{\linewidth}\centering \fixedimg{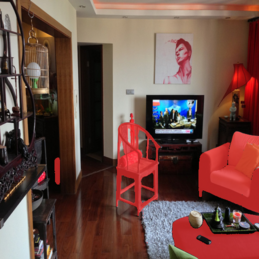}\end{minipage} & \begin{minipage}[t]{\linewidth}\centering \fixedimg{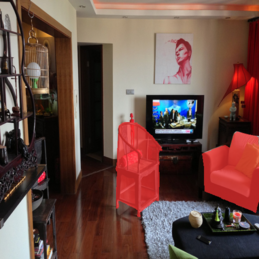}\end{minipage} \\
\centering chair & \centering Seat & \multicolumn{2}{>{\centering\arraybackslash}p{4cm}}{\textit{any support where you can sit (especially the part of a chair or bench etc. on which you sit)}} & \centering 0.00 mIoU & \centering 57.79 mIoU &  \\

\begin{minipage}[t]{\linewidth}\centering \fixedimg{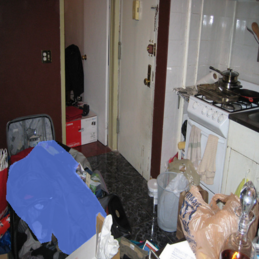}\\\end{minipage} & \begin{minipage}[t]{\linewidth}\centering \fixedimg{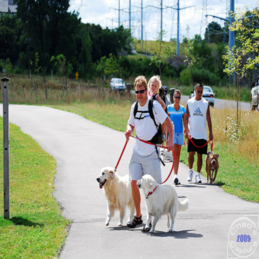}\\\end{minipage} & \begin{minipage}[t]{\linewidth}\centering \fixedimg{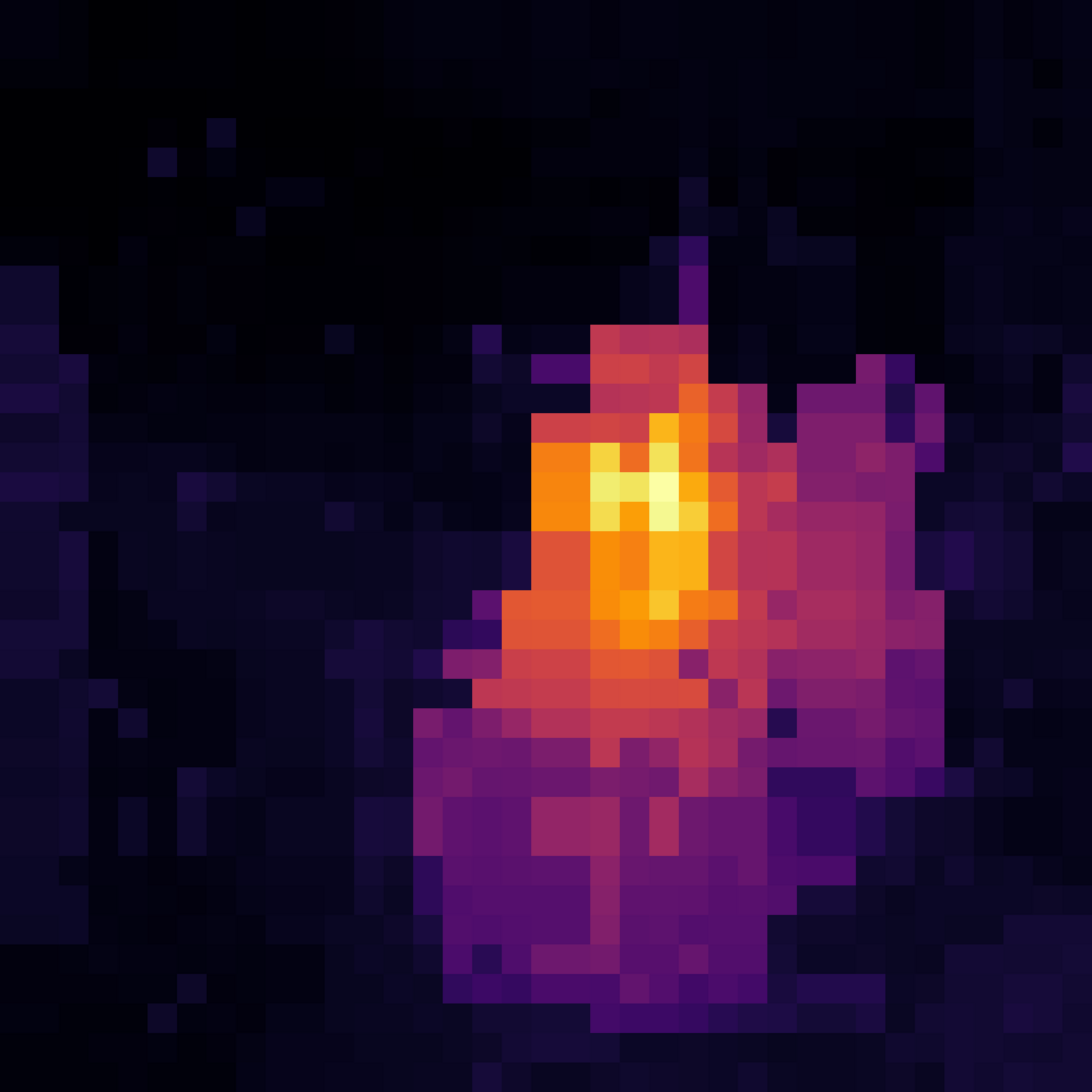}\end{minipage} & \begin{minipage}[t]{\linewidth}\centering \fixedimg{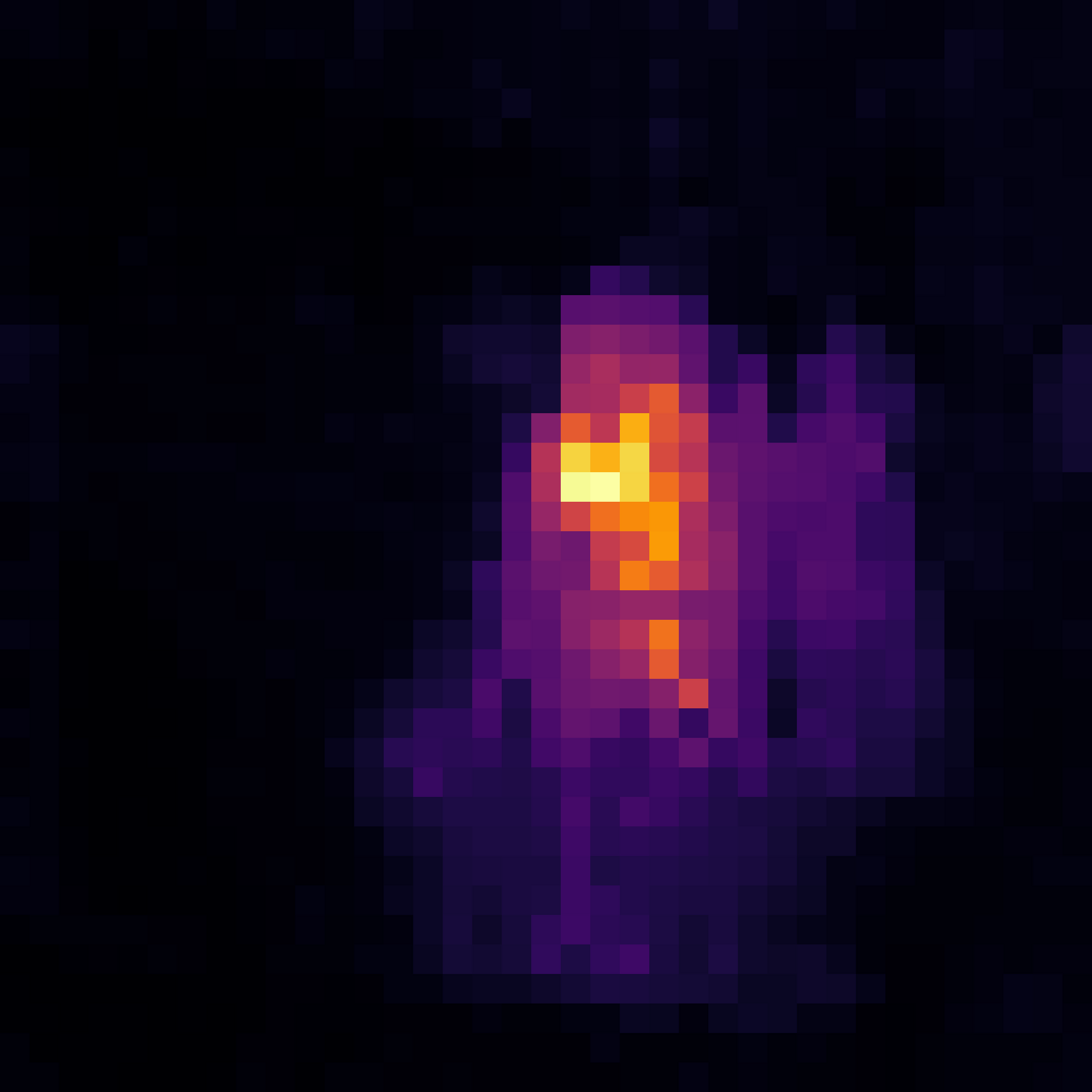}\end{minipage} & \begin{minipage}[t]{\linewidth}\centering \fixedimg{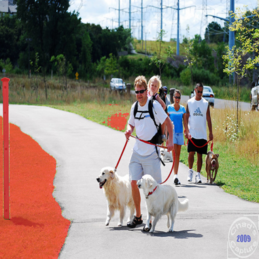}\\\end{minipage} & \begin{minipage}[t]{\linewidth}\centering \fixedimg{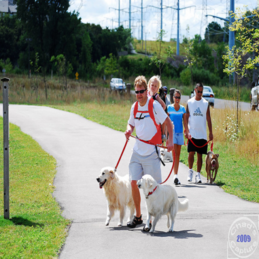}\end{minipage} & \begin{minipage}[t]{\linewidth}\centering \fixedimg{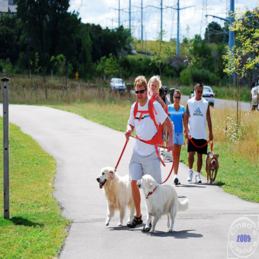}\end{minipage} \\
\centering backpack & \centering Backpack & \multicolumn{2}{>{\centering\arraybackslash}p{4cm}}{\textit{a bag carried by a strap on your back or shoulder}} & \centering 0.00 mIoU & \centering 58.18 mIoU &  \\

\begin{minipage}[t]{\linewidth}\centering \fixedimg{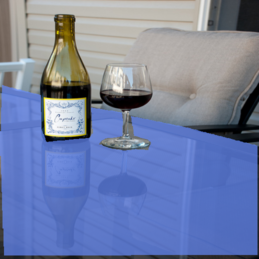}\\\end{minipage} & \begin{minipage}[t]{\linewidth}\centering \fixedimg{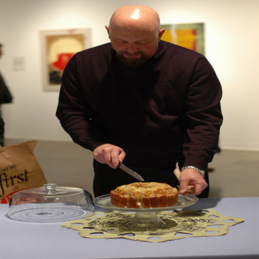}\\\end{minipage} & \begin{minipage}[t]{\linewidth}\centering \fixedimg{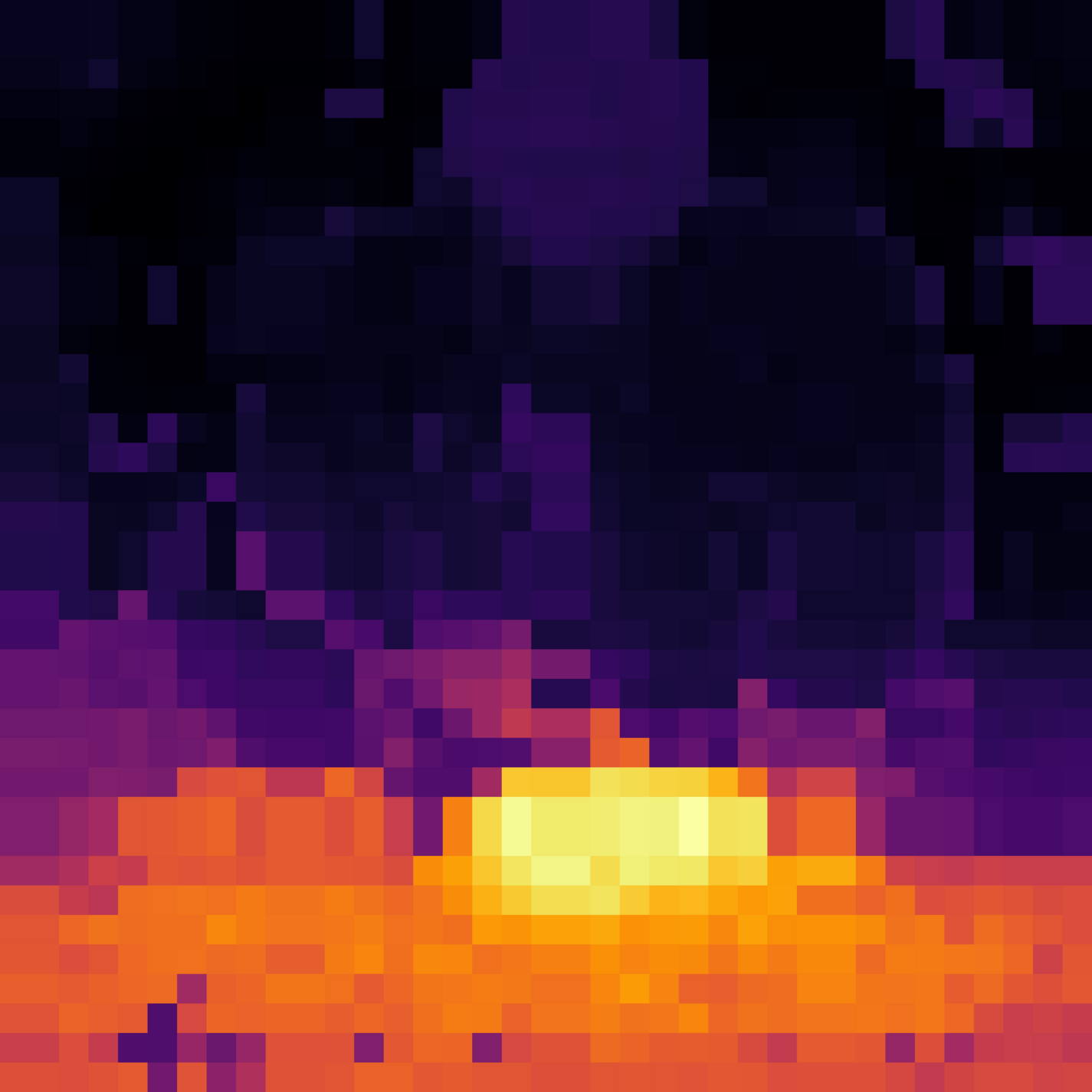}\end{minipage} & \begin{minipage}[t]{\linewidth}\centering \fixedimg{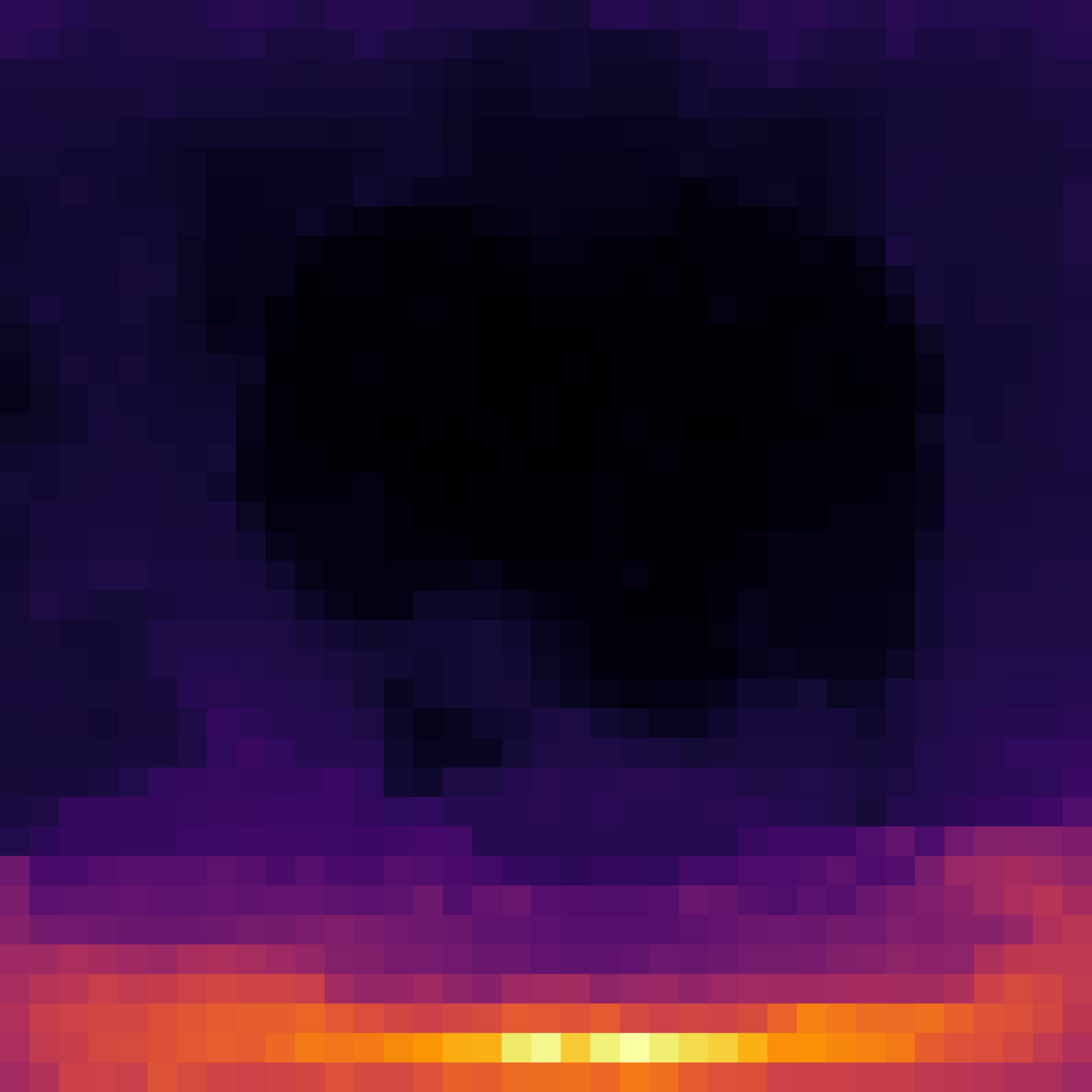}\end{minipage} & \begin{minipage}[t]{\linewidth}\centering \fixedimg{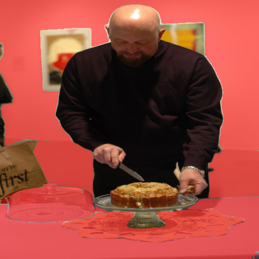}\\\end{minipage} & \begin{minipage}[t]{\linewidth}\centering \fixedimg{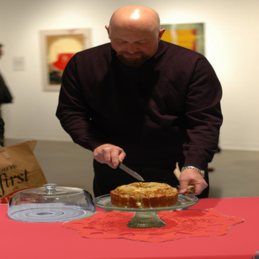}\end{minipage} & \begin{minipage}[t]{\linewidth}\centering \fixedimg{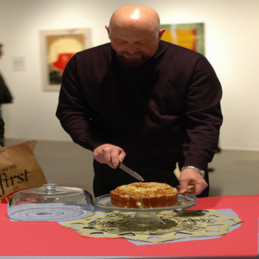}\end{minipage} \\
\centering dining table & \centering Table & \multicolumn{2}{>{\centering\arraybackslash}p{4cm}}{\textit{a company of people assembled at a table for a meal or game}} & \centering 19.90 mIoU & \centering 58.23 mIoU &  \\

\begin{minipage}[t]{\linewidth}\centering \fixedimg{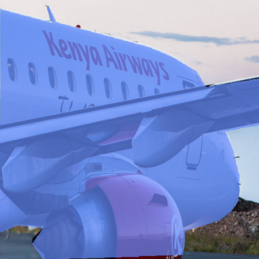}\\\end{minipage} & \begin{minipage}[t]{\linewidth}\centering \fixedimg{}\\\end{minipage} & \begin{minipage}[t]{\linewidth}\centering \fixedimg{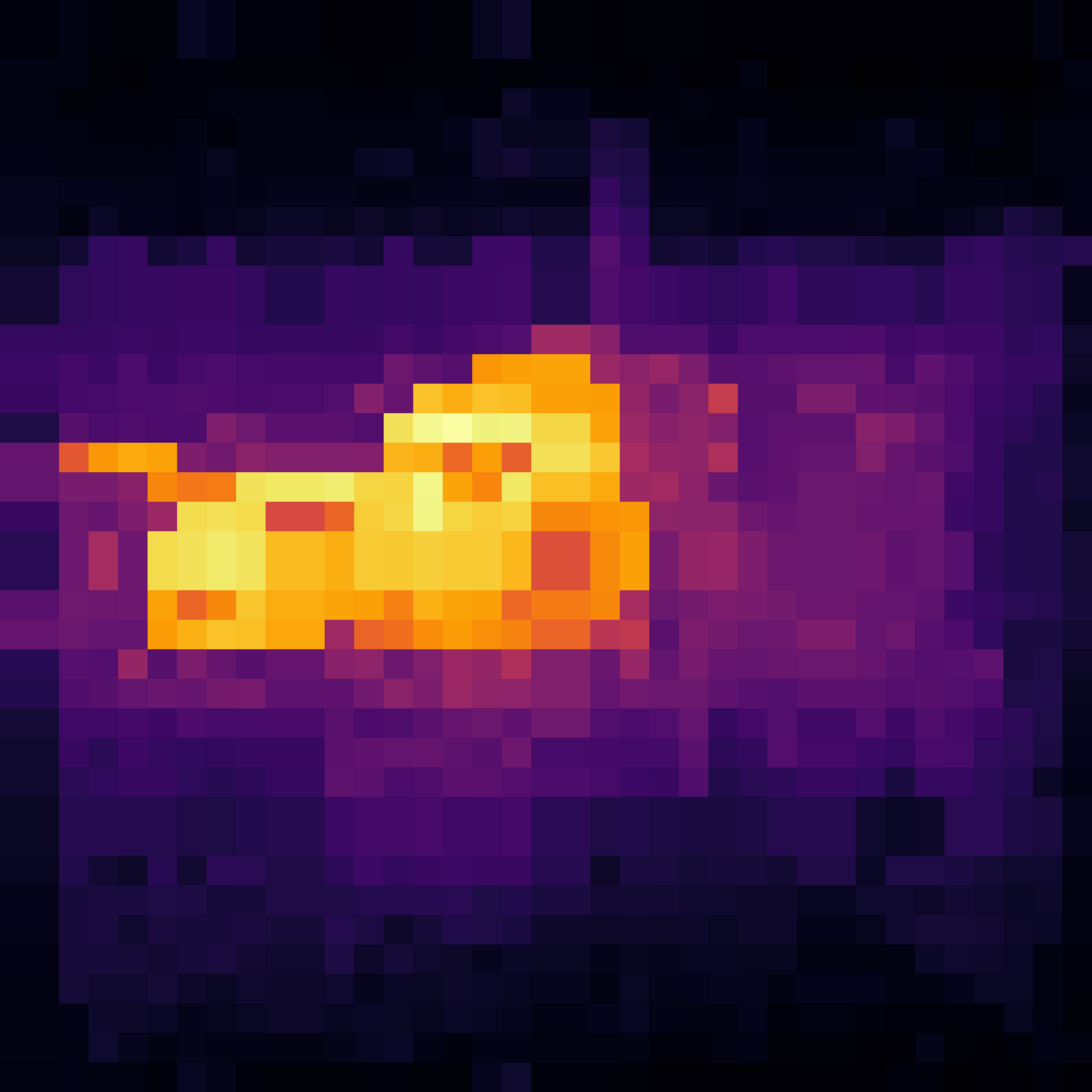}\end{minipage} & \begin{minipage}[t]{\linewidth}\centering \fixedimg{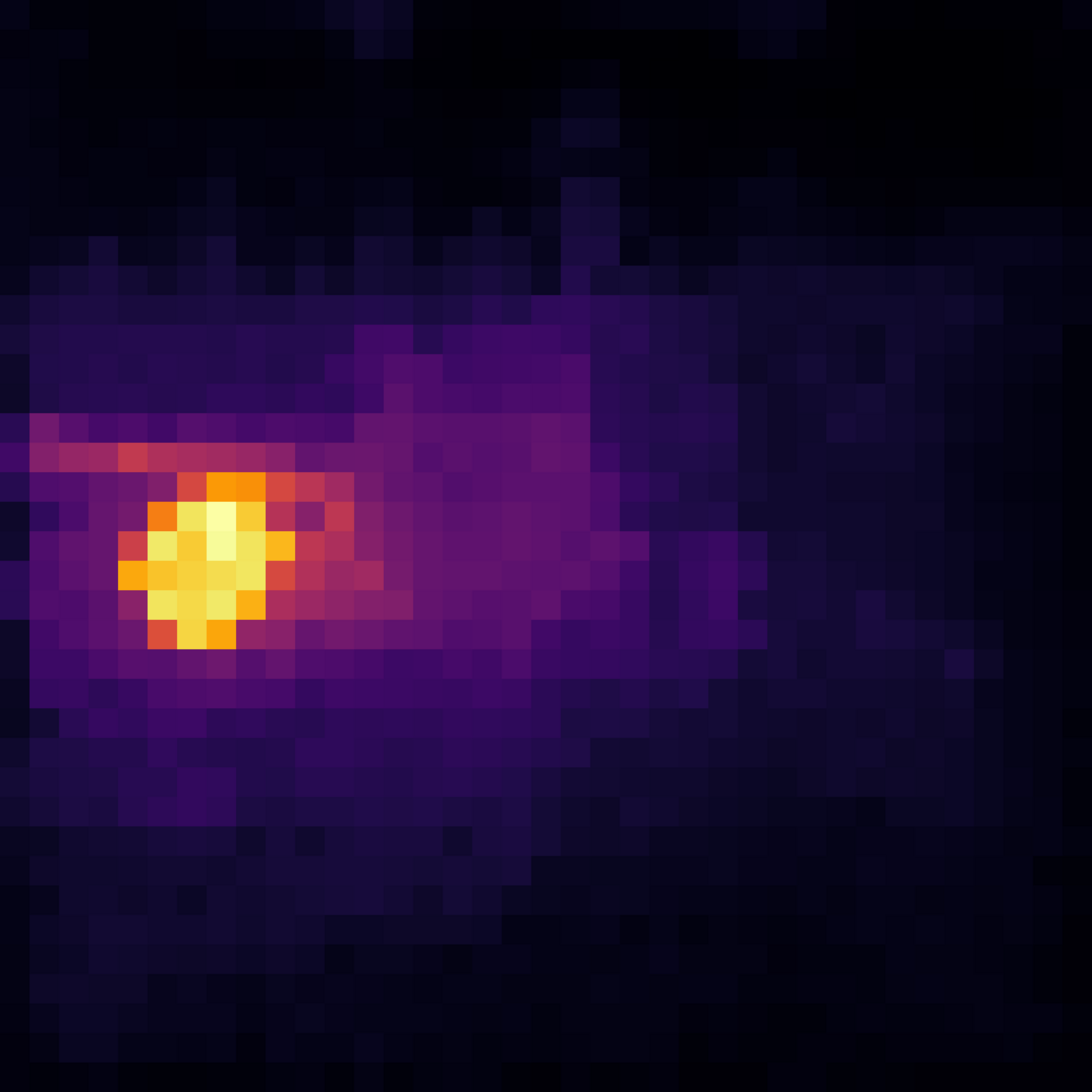}\end{minipage} & \begin{minipage}[t]{\linewidth}\centering \fixedimg{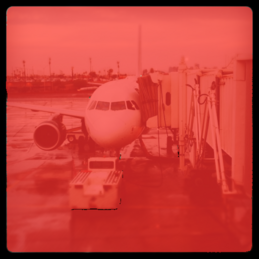}\\\end{minipage} & \begin{minipage}[t]{\linewidth}\centering \fixedimg{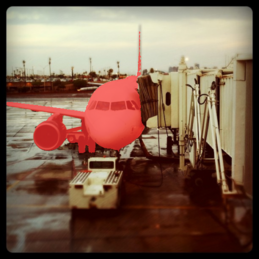}\end{minipage} & \begin{minipage}[t]{\linewidth}\centering \fixedimg{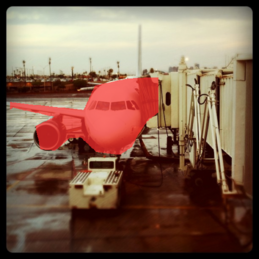}\end{minipage} \\
\centering airplane & \centering Airplane & \multicolumn{2}{>{\centering\arraybackslash}p{4cm}}{\textit{an aircraft that has a fixed wing and is powered by propellers or jets}} & \centering 10.36 mIoU & \centering 72.47 mIoU &  \\

\begin{minipage}[t]{\linewidth}\centering \fixedimg{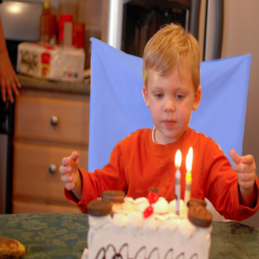}\\\end{minipage} & \begin{minipage}[t]{\linewidth}\centering \fixedimg{}\\\end{minipage} & \begin{minipage}[t]{\linewidth}\centering \fixedimg{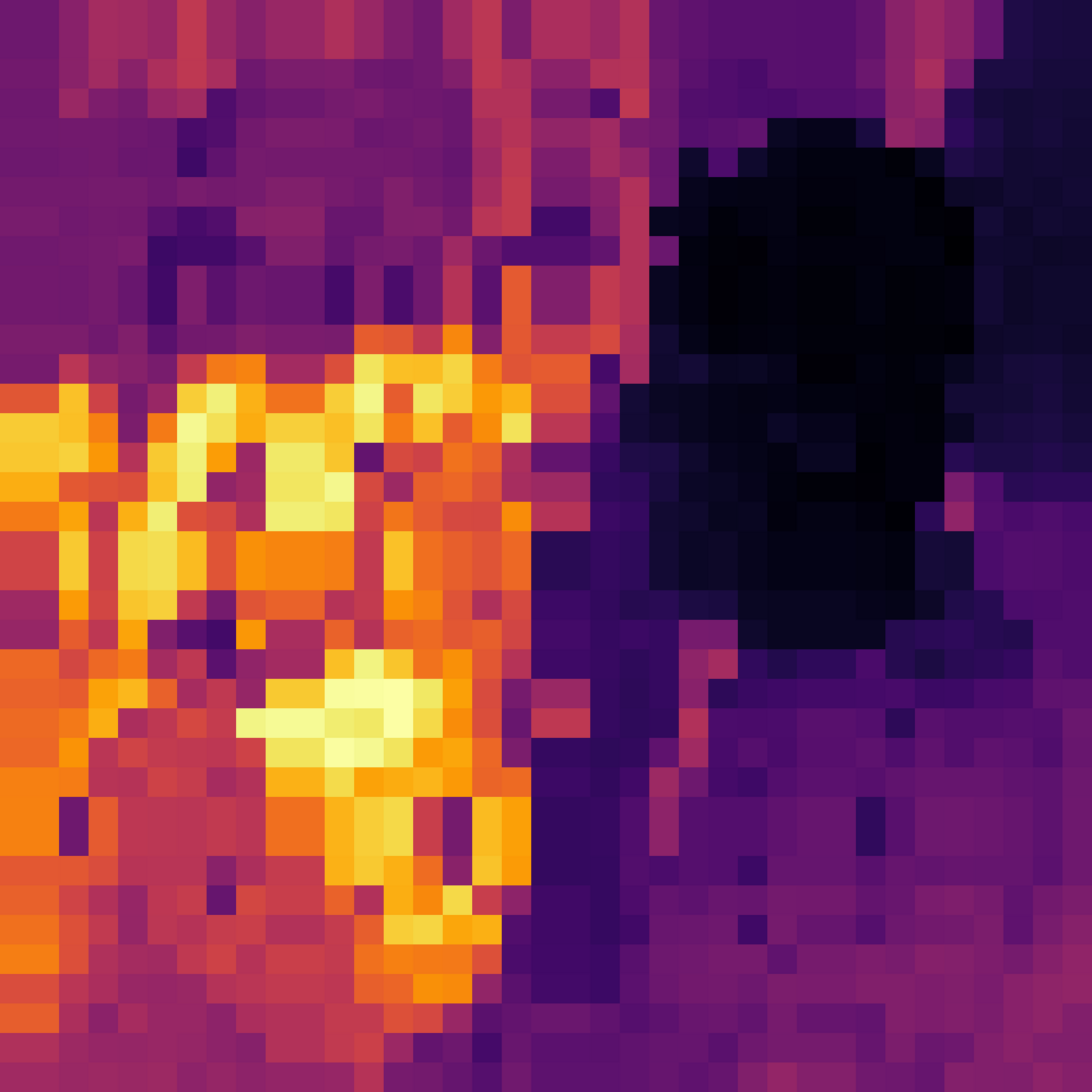}\end{minipage} & \begin{minipage}[t]{\linewidth}\centering \fixedimg{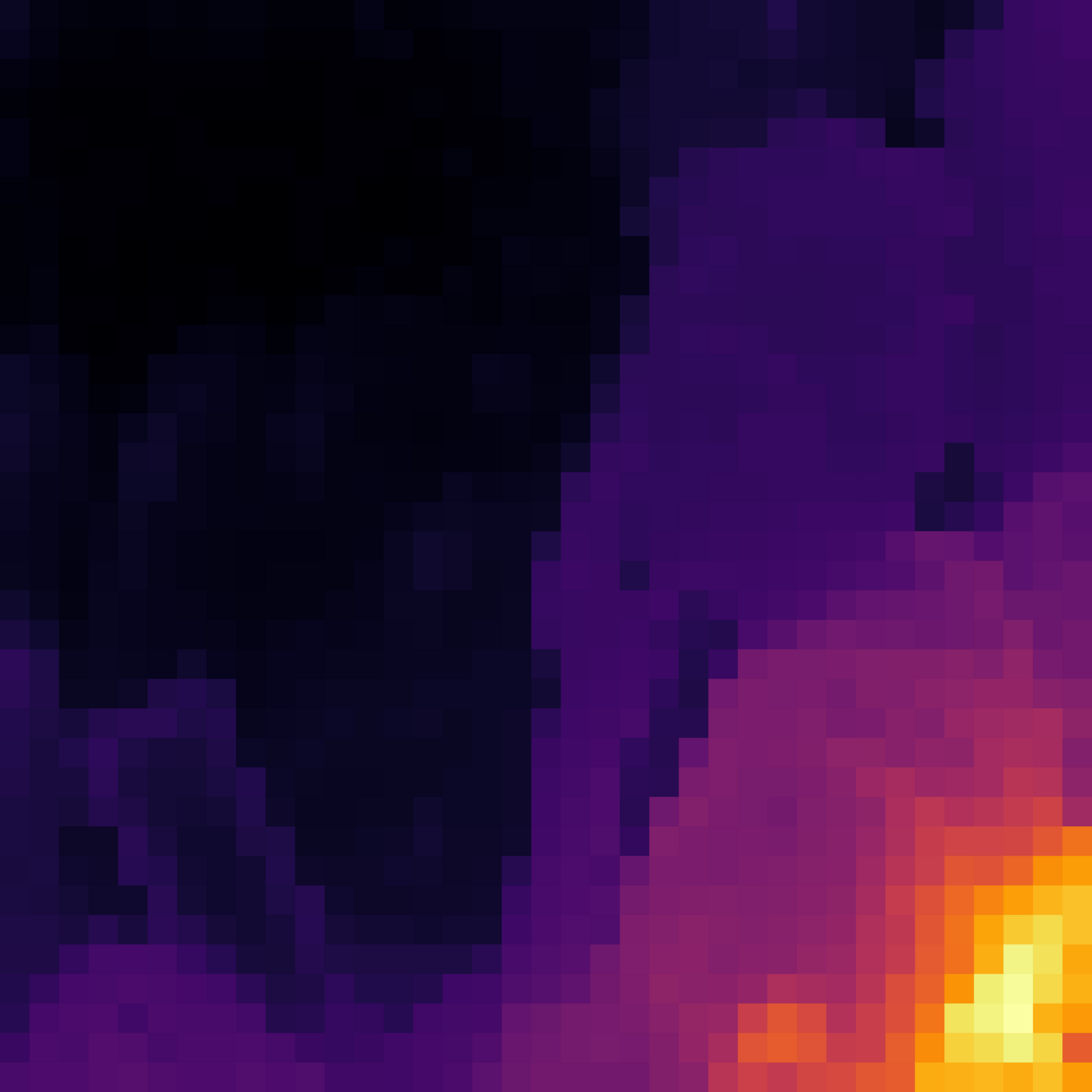}\end{minipage} & \begin{minipage}[t]{\linewidth}\centering \fixedimg{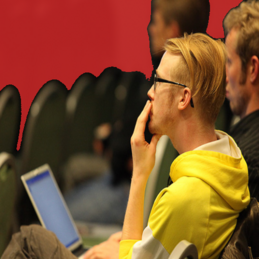}\\\end{minipage} & \begin{minipage}[t]{\linewidth}\centering \fixedimg{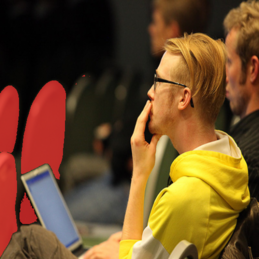}\end{minipage} & \begin{minipage}[t]{\linewidth}\centering \fixedimg{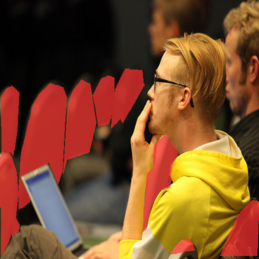}\end{minipage} \\
\centering chair & \centering Chair & \multicolumn{2}{>{\centering\arraybackslash}p{4cm}}{\textit{*Description not found*}} & \centering 0.17 mIoU & \centering 40.51 mIoU &  \\

\begin{minipage}[t]{\linewidth}\centering \fixedimg{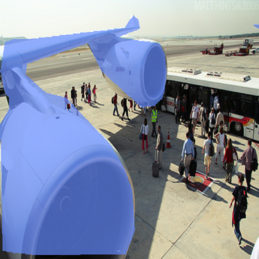}\\\end{minipage} & \begin{minipage}[t]{\linewidth}\centering \fixedimg{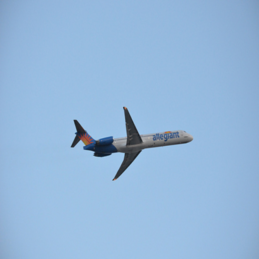}\\\end{minipage} & \begin{minipage}[t]{\linewidth}\centering \fixedimg{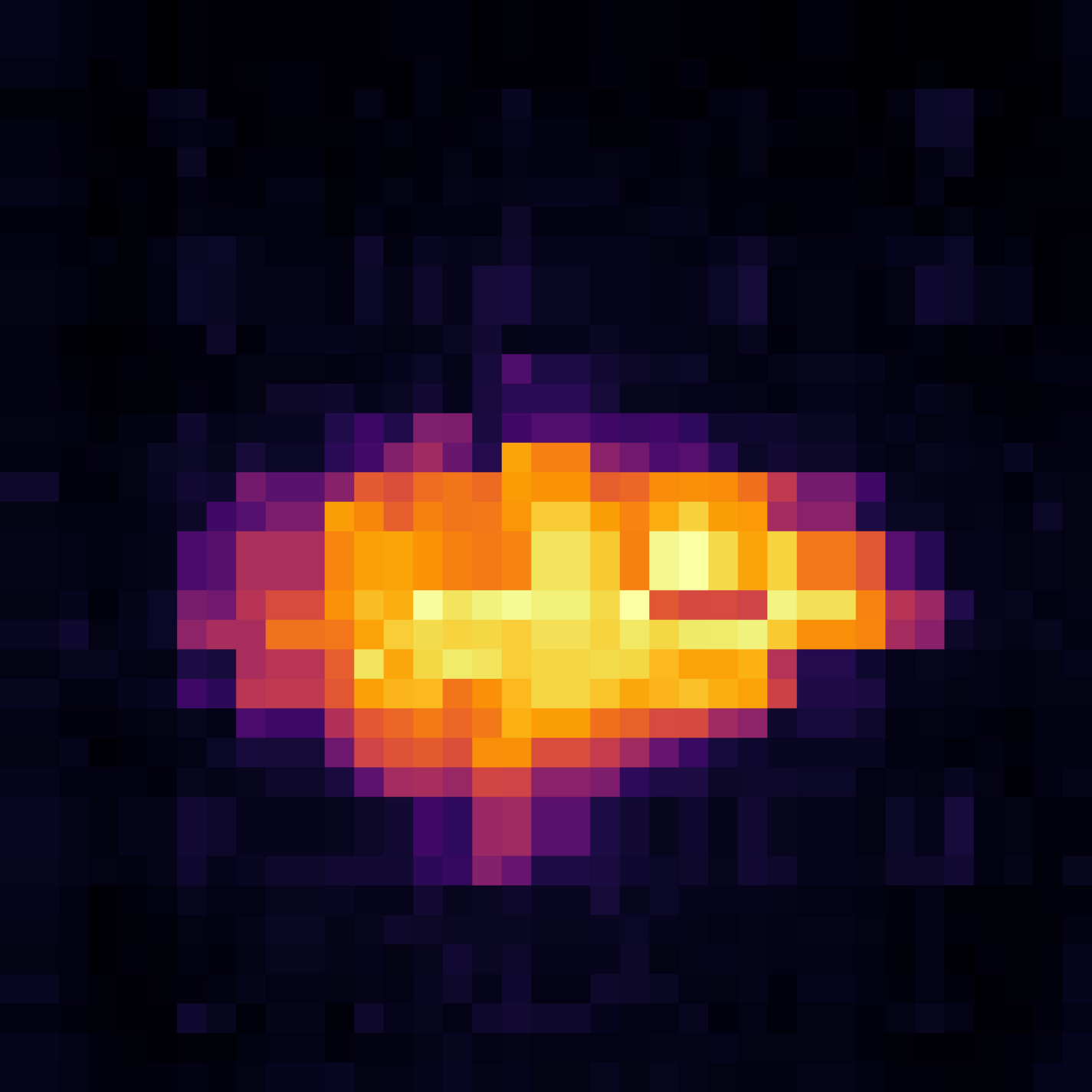}\end{minipage} & \begin{minipage}[t]{\linewidth}\centering \fixedimg{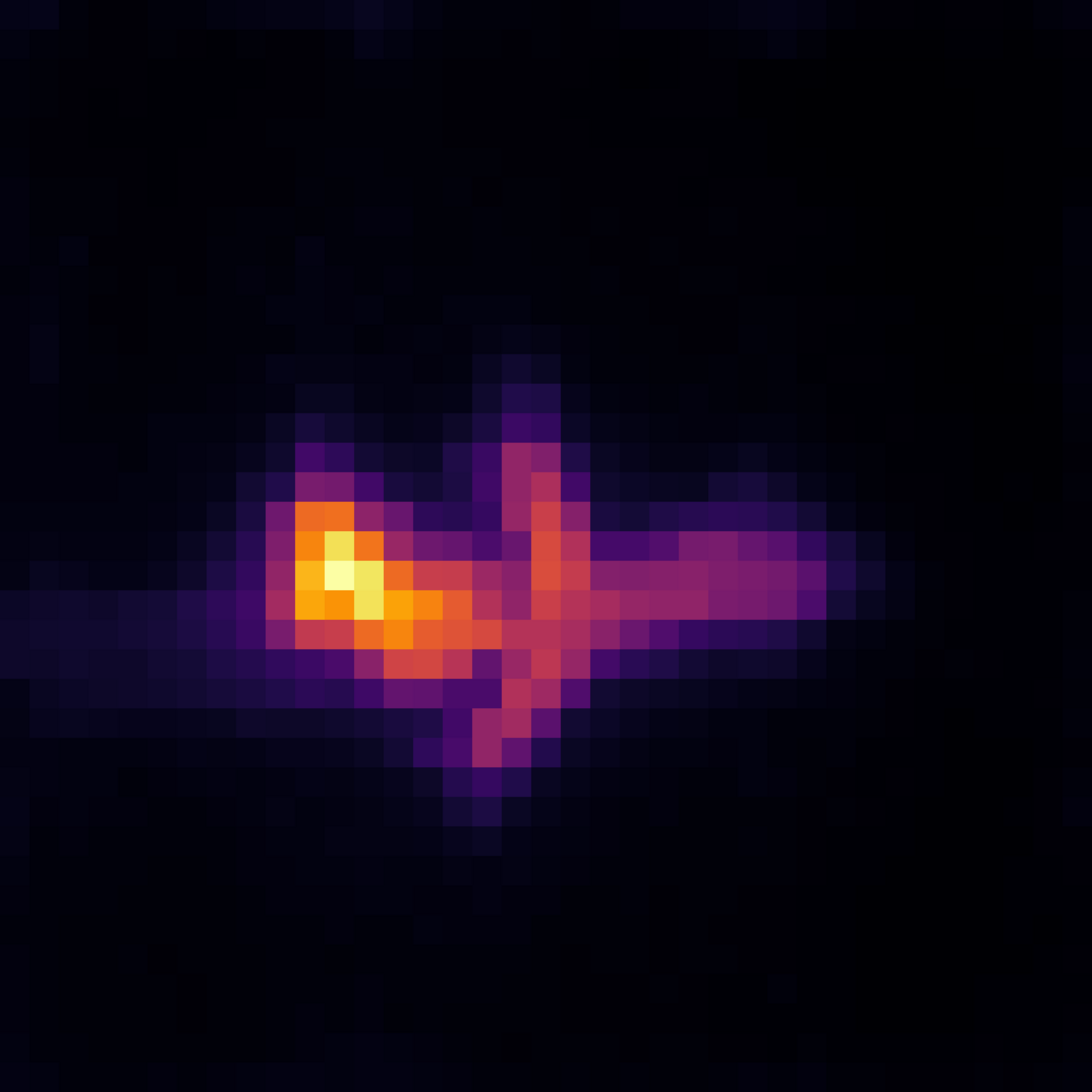}\end{minipage} & \begin{minipage}[t]{\linewidth}\centering \fixedimg{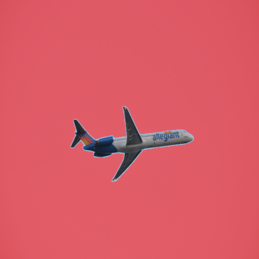}\\\end{minipage} & \begin{minipage}[t]{\linewidth}\centering \fixedimg{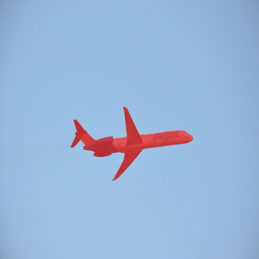}\end{minipage} & \begin{minipage}[t]{\linewidth}\centering \fixedimg{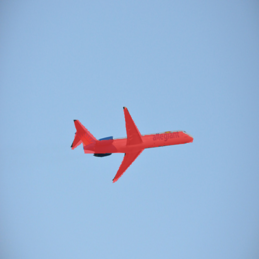}\end{minipage} \\
\centering airplane & \centering Jet engine & \multicolumn{2}{>{\centering\arraybackslash}p{4cm}}{\textit{a gas turbine produces a stream of hot gas that propels a jet plane by reaction propulsion}} & \centering 0.12 mIoU & \centering 83.25 mIoU &  \\

\begin{minipage}[t]{\linewidth}\centering \fixedimg{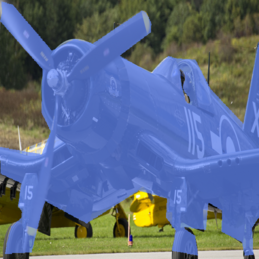}\\\end{minipage} & \begin{minipage}[t]{\linewidth}\centering \fixedimg{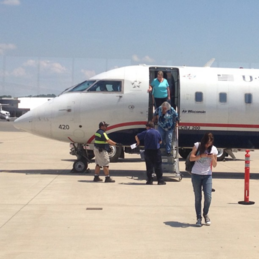}\\\end{minipage} & \begin{minipage}[t]{\linewidth}\centering \fixedimg{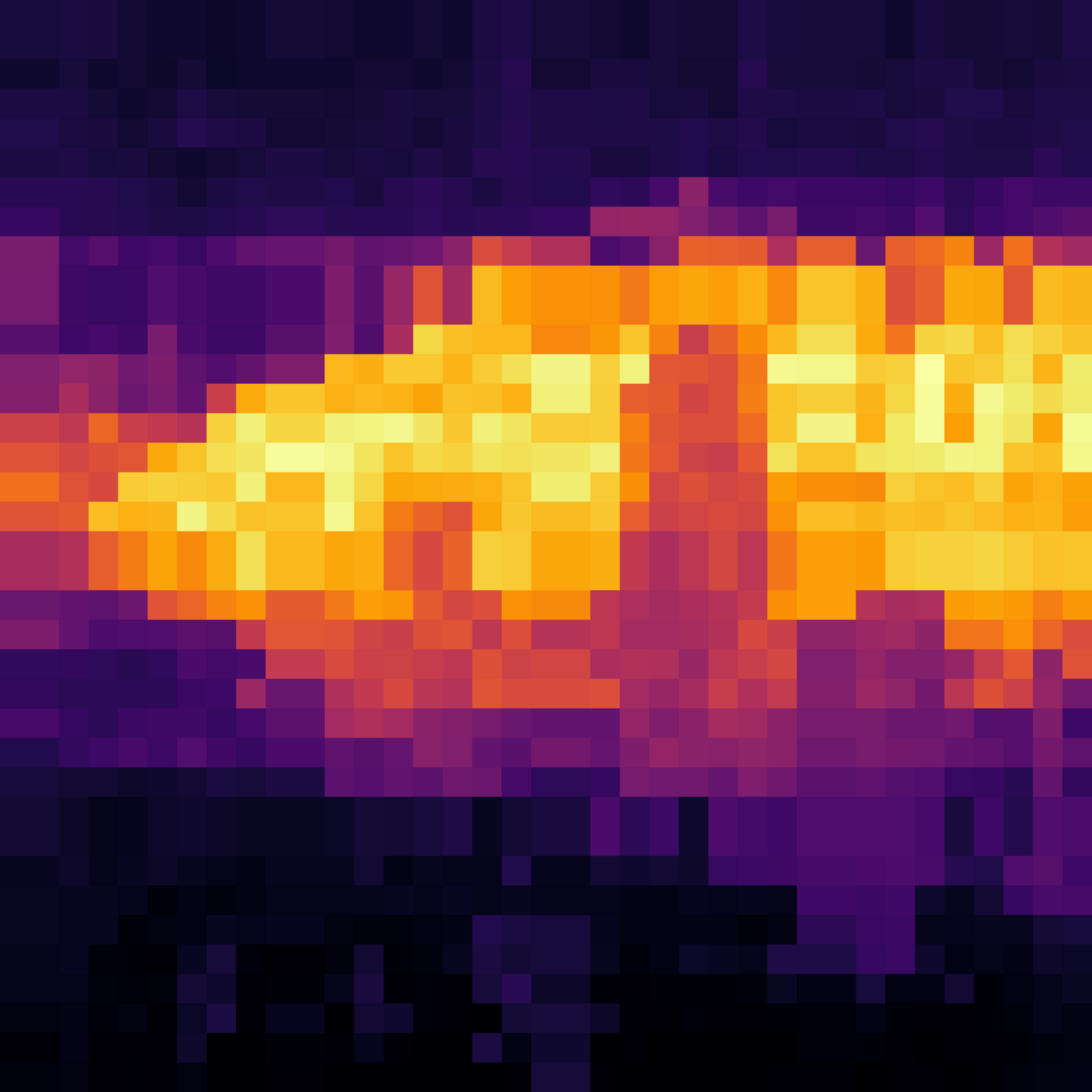}\end{minipage} & \begin{minipage}[t]{\linewidth}\centering \fixedimg{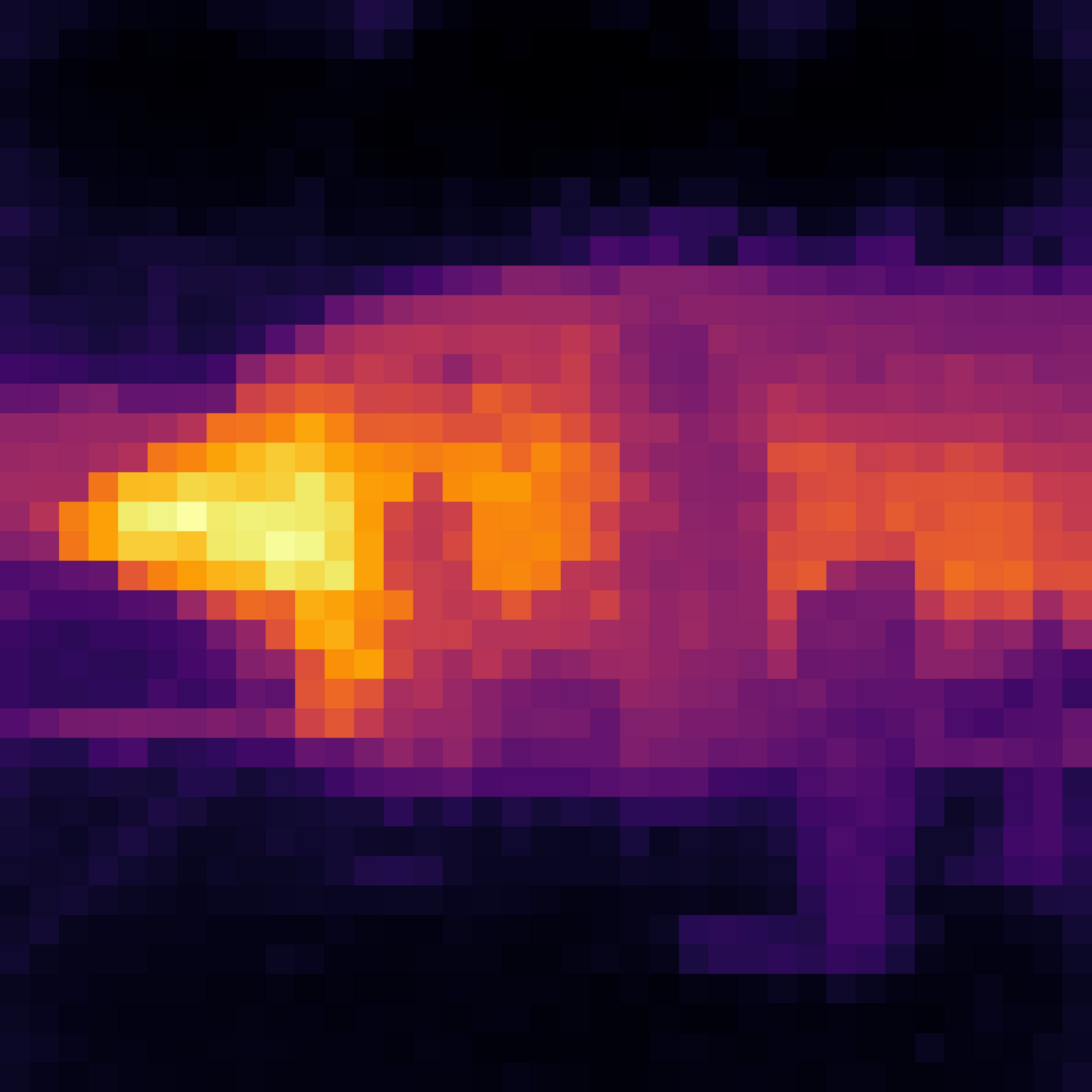}\end{minipage} & \begin{minipage}[t]{\linewidth}\centering \fixedimg{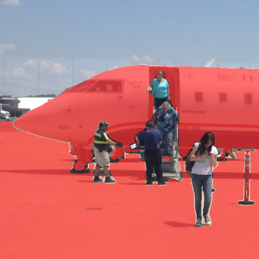}\\\end{minipage} & \begin{minipage}[t]{\linewidth}\centering \fixedimg{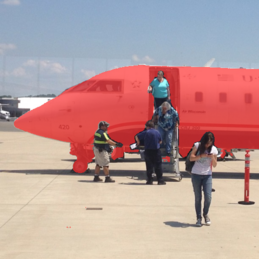}\end{minipage} & \begin{minipage}[t]{\linewidth}\centering \fixedimg{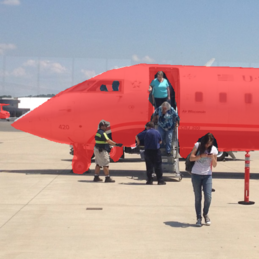}\end{minipage} \\
\centering airplane & \centering Airplane & \multicolumn{2}{>{\centering\arraybackslash}p{4cm}}{\textit{an aircraft that has a fixed wing and is powered by propellers or jets}} & \centering 35.28 mIoU & \centering 91.19 mIoU &  \\

\begin{minipage}[t]{\linewidth}\centering \fixedimg{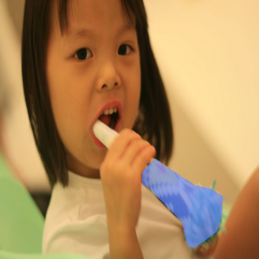}\\\end{minipage} & \begin{minipage}[t]{\linewidth}\centering \fixedimg{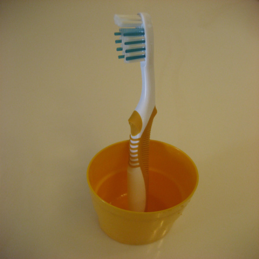}\\\end{minipage} & \begin{minipage}[t]{\linewidth}\centering \fixedimg{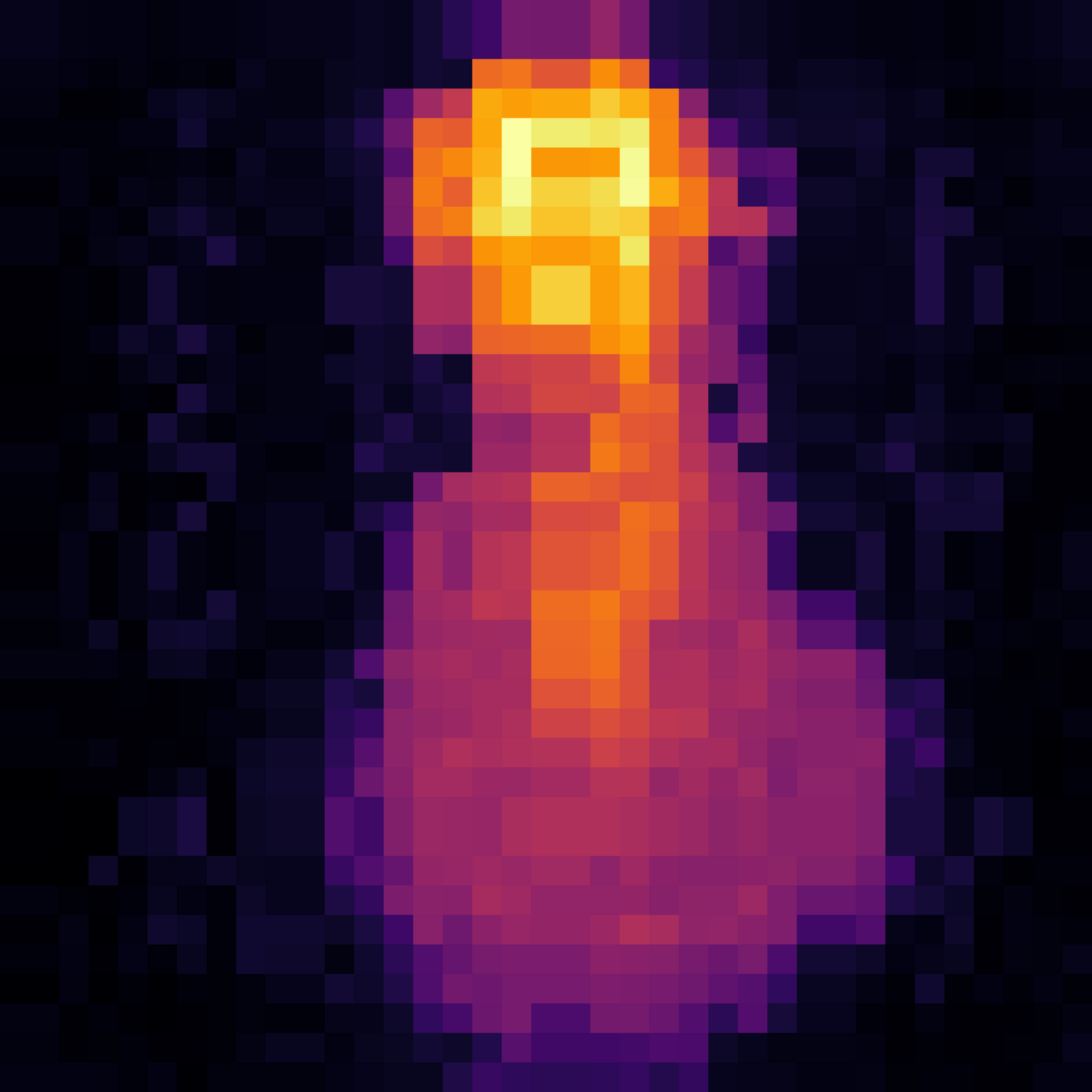}\end{minipage} & \begin{minipage}[t]{\linewidth}\centering \fixedimg{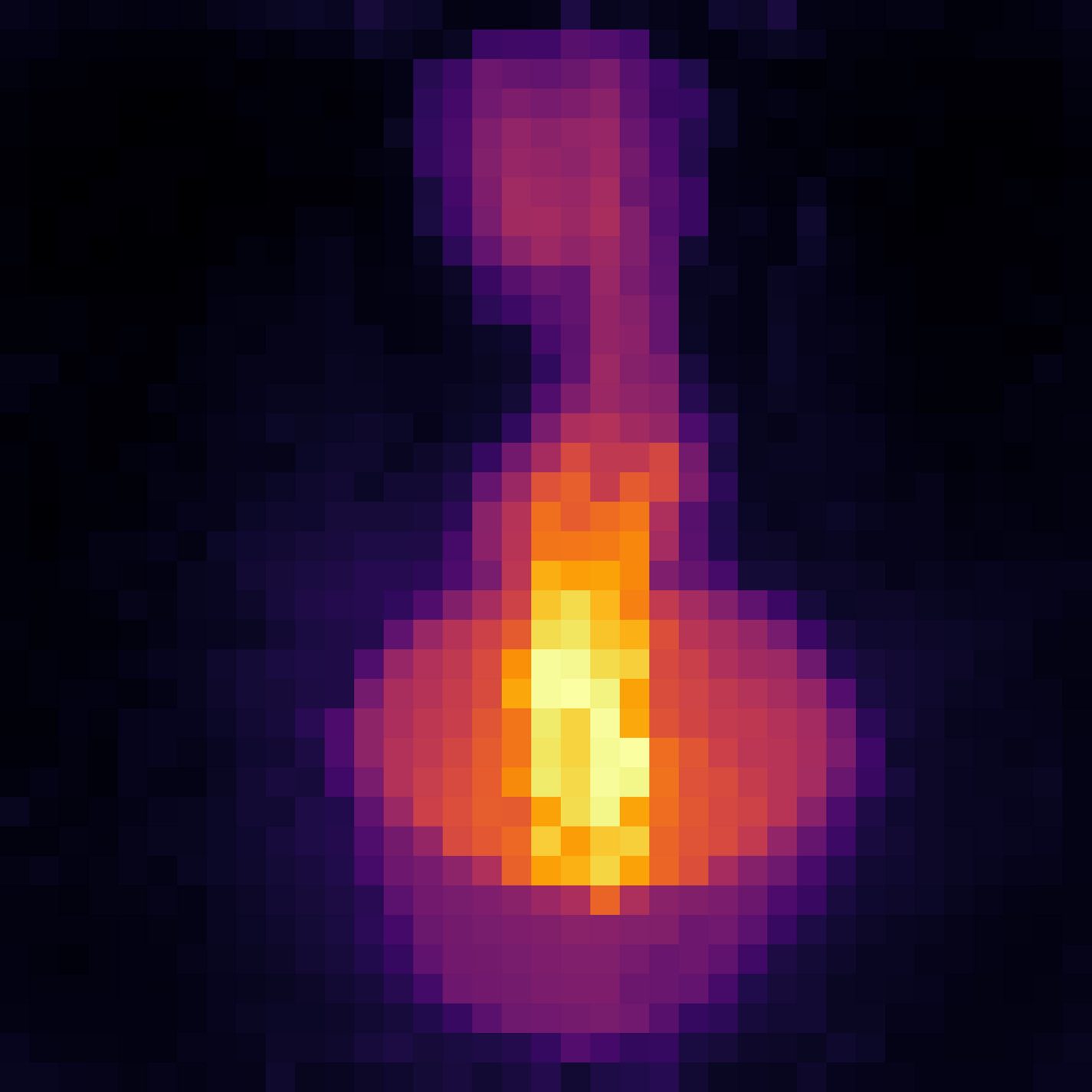}\end{minipage} & \begin{minipage}[t]{\linewidth}\centering \fixedimg{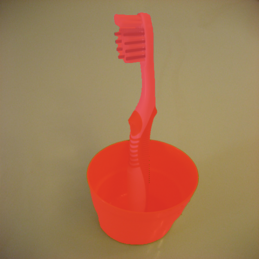}\\\end{minipage} & \begin{minipage}[t]{\linewidth}\centering \fixedimg{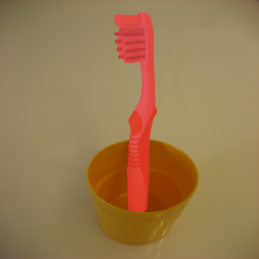}\end{minipage} & \begin{minipage}[t]{\linewidth}\centering \fixedimg{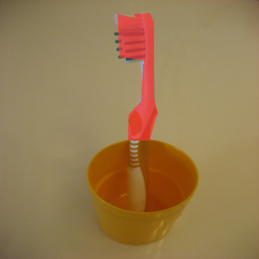}\end{minipage} \\
\centering toothbrush & \centering Toothbrush & \multicolumn{2}{>{\centering\arraybackslash}p{4cm}}{\textit{small brush; has long handle; used to clean teeth}} & \centering 23.30 mIoU & \centering 59.88 mIoU &  \\


\begin{minipage}[t]{\linewidth}\centering \fixedimg{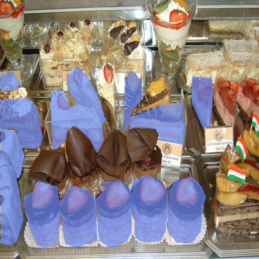}\\\end{minipage} & \begin{minipage}[t]{\linewidth}\centering \fixedimg{}\\\end{minipage} & \begin{minipage}[t]{\linewidth}\centering \fixedimg{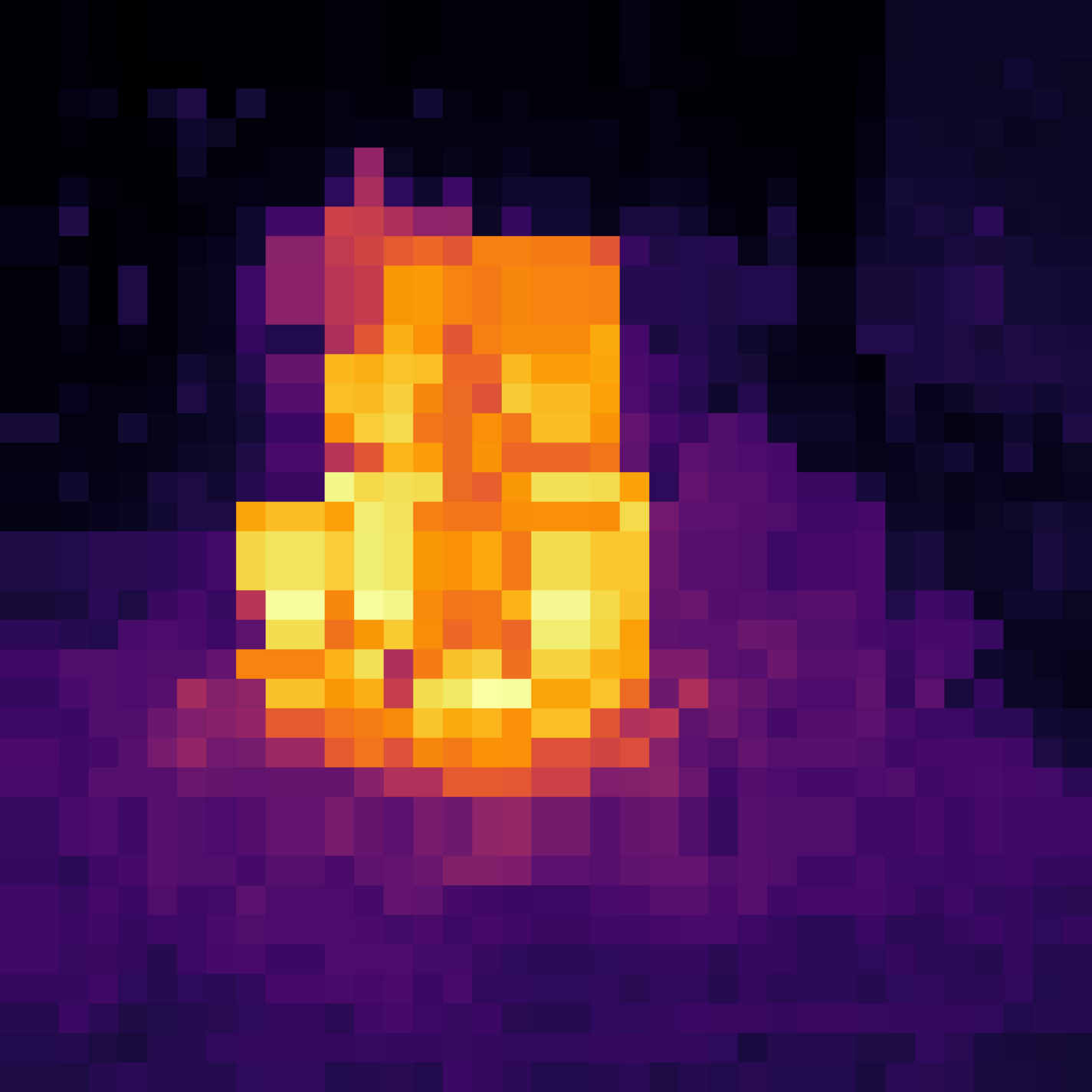}\end{minipage} & \begin{minipage}[t]{\linewidth}\centering \fixedimg{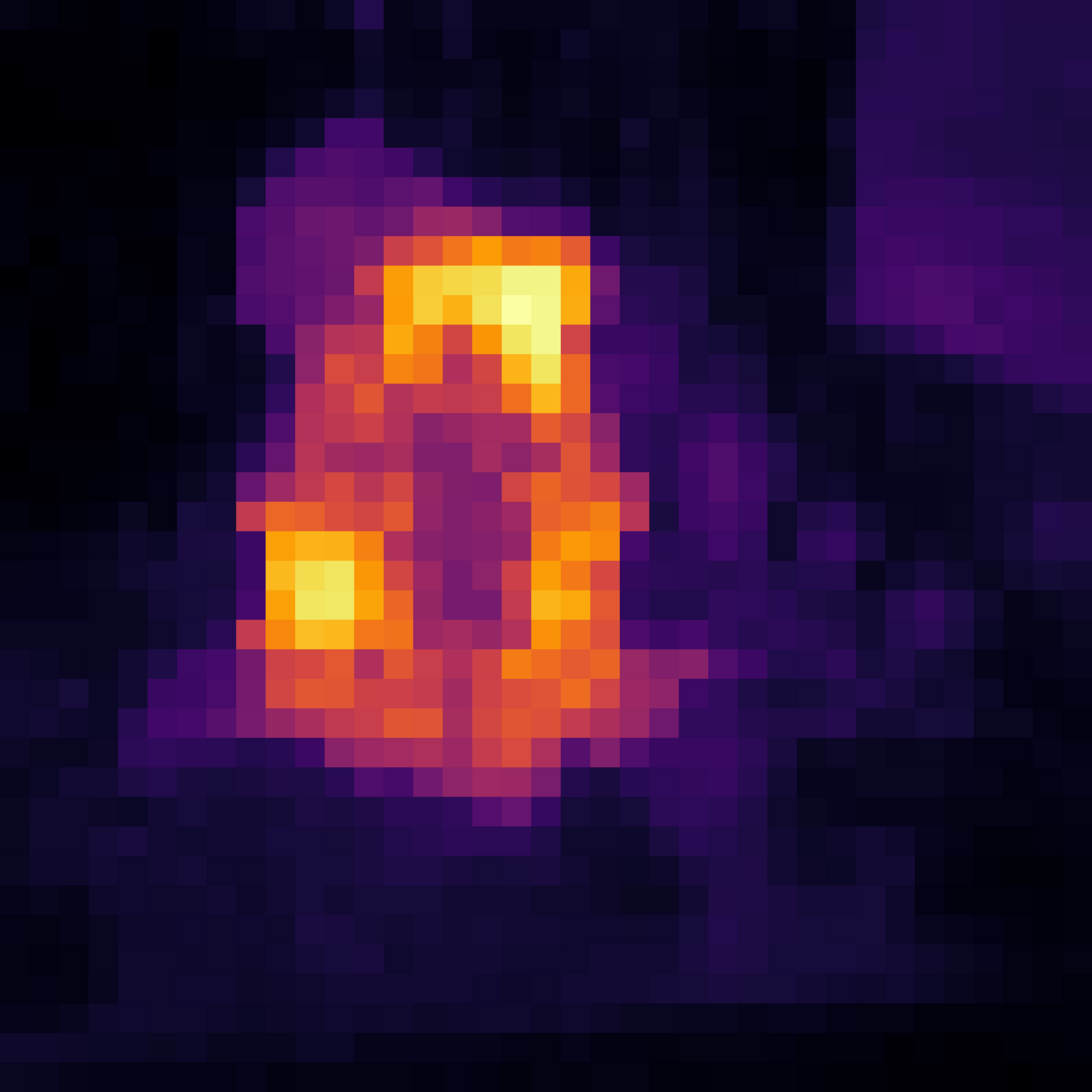}\end{minipage} & \begin{minipage}[t]{\linewidth}\centering \fixedimg{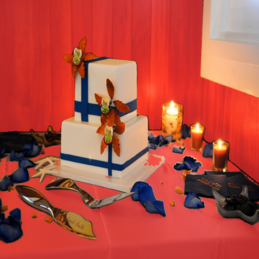}\\\end{minipage} & \begin{minipage}[t]{\linewidth}\centering \fixedimg{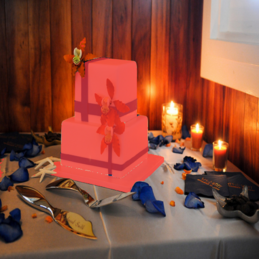}\end{minipage} & \begin{minipage}[t]{\linewidth}\centering \fixedimg{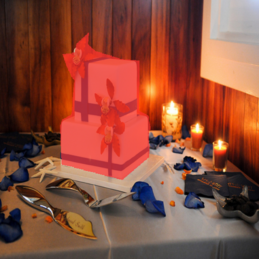}\end{minipage} \\
\centering cake & \centering Cake & \multicolumn{2}{>{\centering\arraybackslash}p{4cm}}{\textit{baked goods made from or based on a mixture of flour, sugar, eggs, and fat}} & \centering 0.87 mIoU & \centering 74.98 mIoU &  \\

\begin{minipage}[t]{\linewidth}\centering \fixedimg{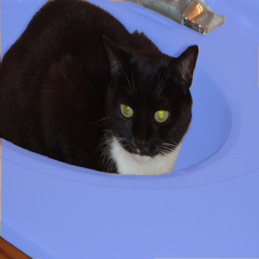}\\\end{minipage} & \begin{minipage}[t]{\linewidth}\centering \fixedimg{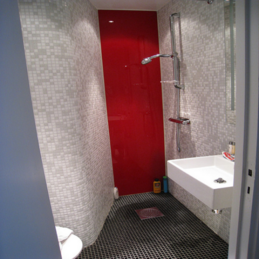}\\\end{minipage} & \begin{minipage}[t]{\linewidth}\centering \fixedimg{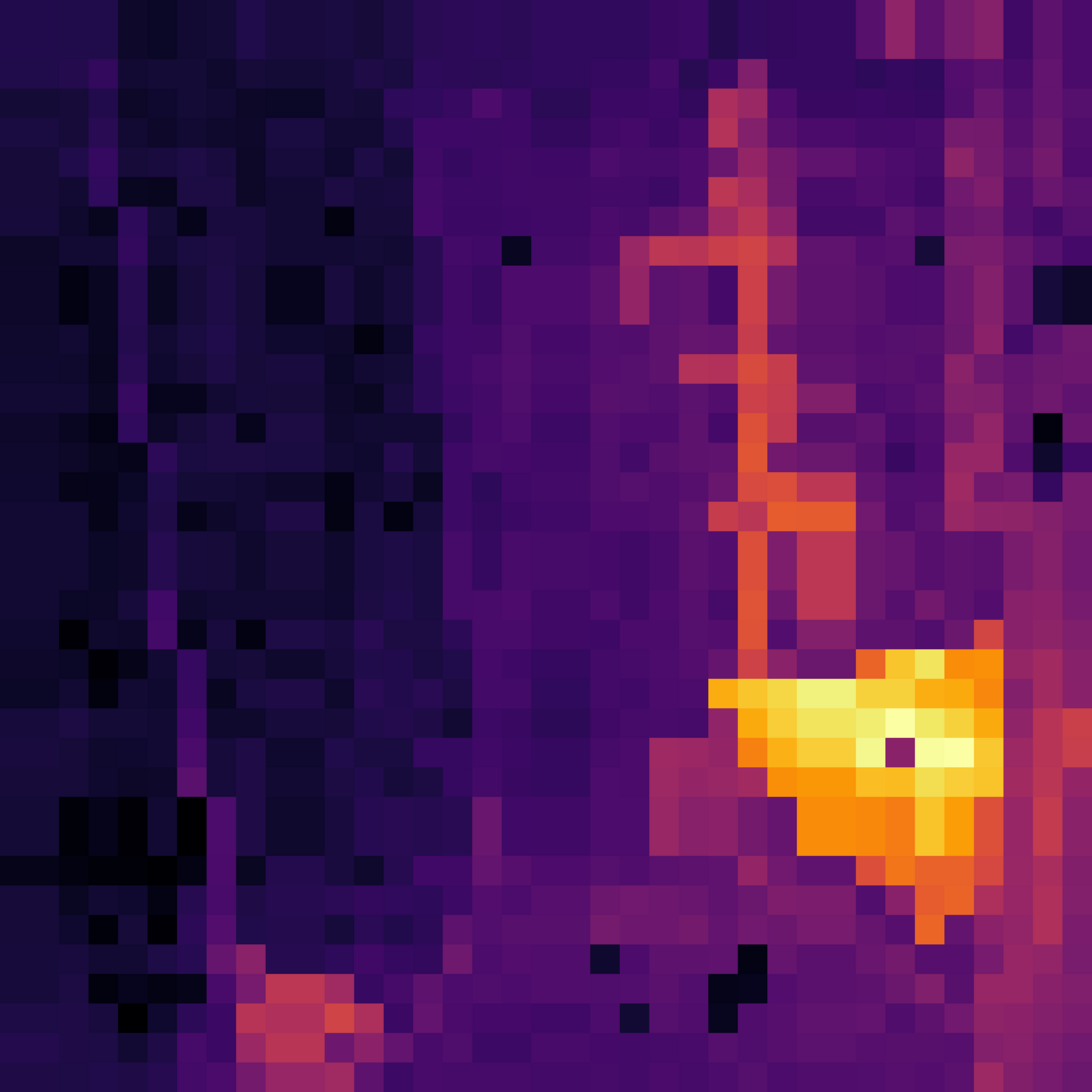}\end{minipage} & \begin{minipage}[t]{\linewidth}\centering \fixedimg{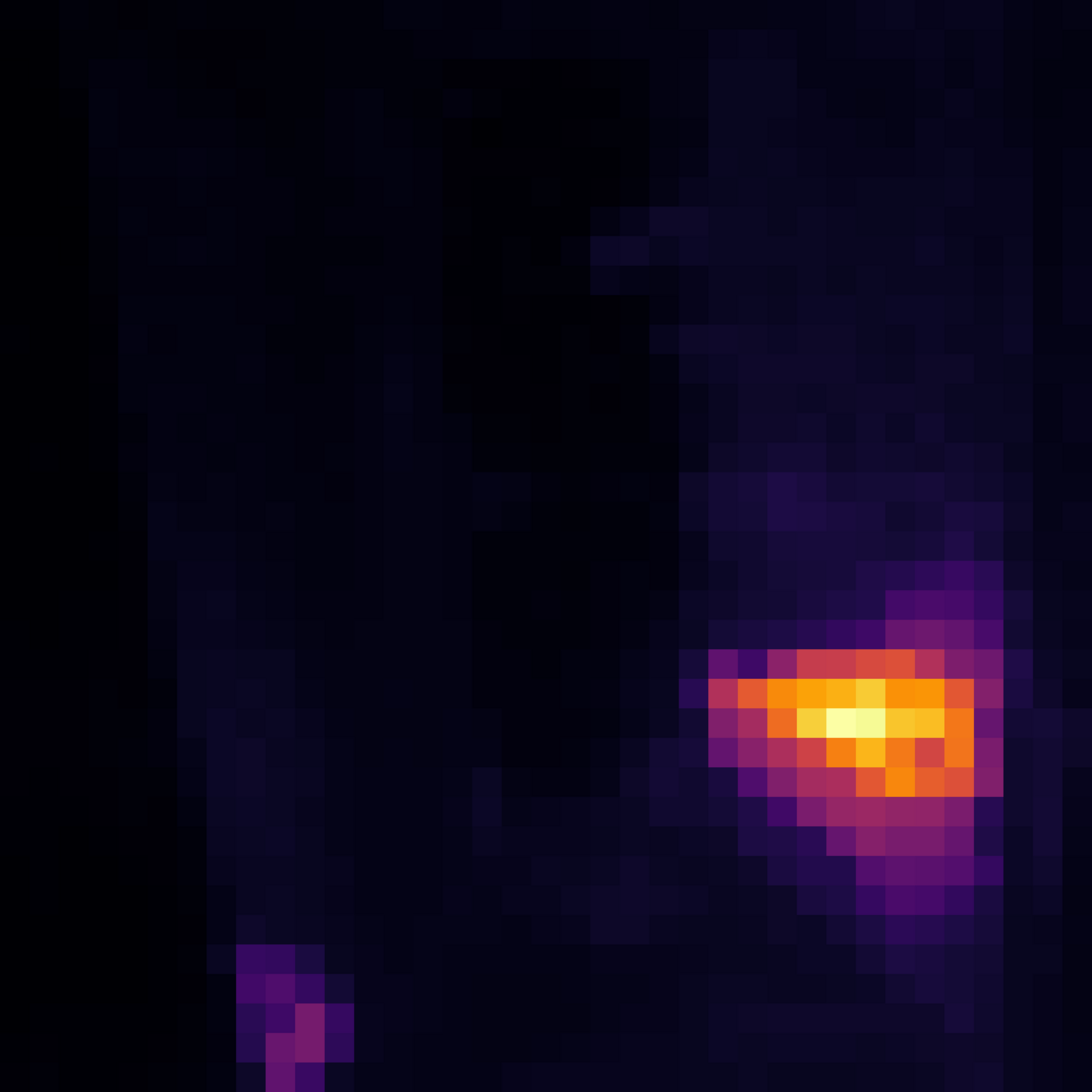}\end{minipage} & \begin{minipage}[t]{\linewidth}\centering \fixedimg{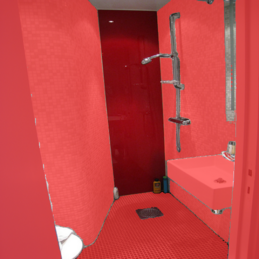}\\\end{minipage} & \begin{minipage}[t]{\linewidth}\centering \fixedimg{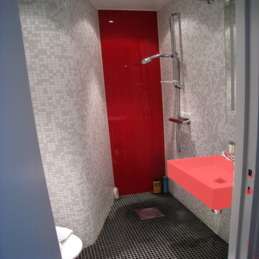}\end{minipage} & \begin{minipage}[t]{\linewidth}\centering \fixedimg{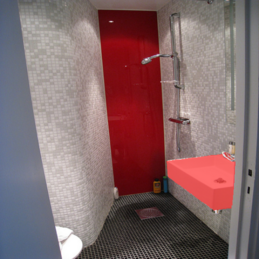}\end{minipage} \\
\centering sink & \centering Sink & \multicolumn{2}{>{\centering\arraybackslash}p{4cm}}{\textit{*Description not found*}} & \centering 5.07 mIoU & \centering 94.34 mIoU &  \\

\begin{minipage}[t]{\linewidth}\centering \fixedimg{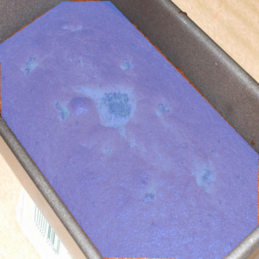}\\\end{minipage} & \begin{minipage}[t]{\linewidth}\centering \fixedimg{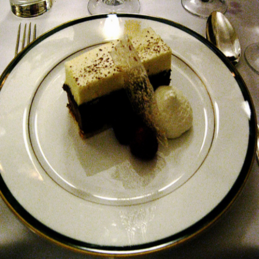}\\\end{minipage} & \begin{minipage}[t]{\linewidth}\centering \fixedimg{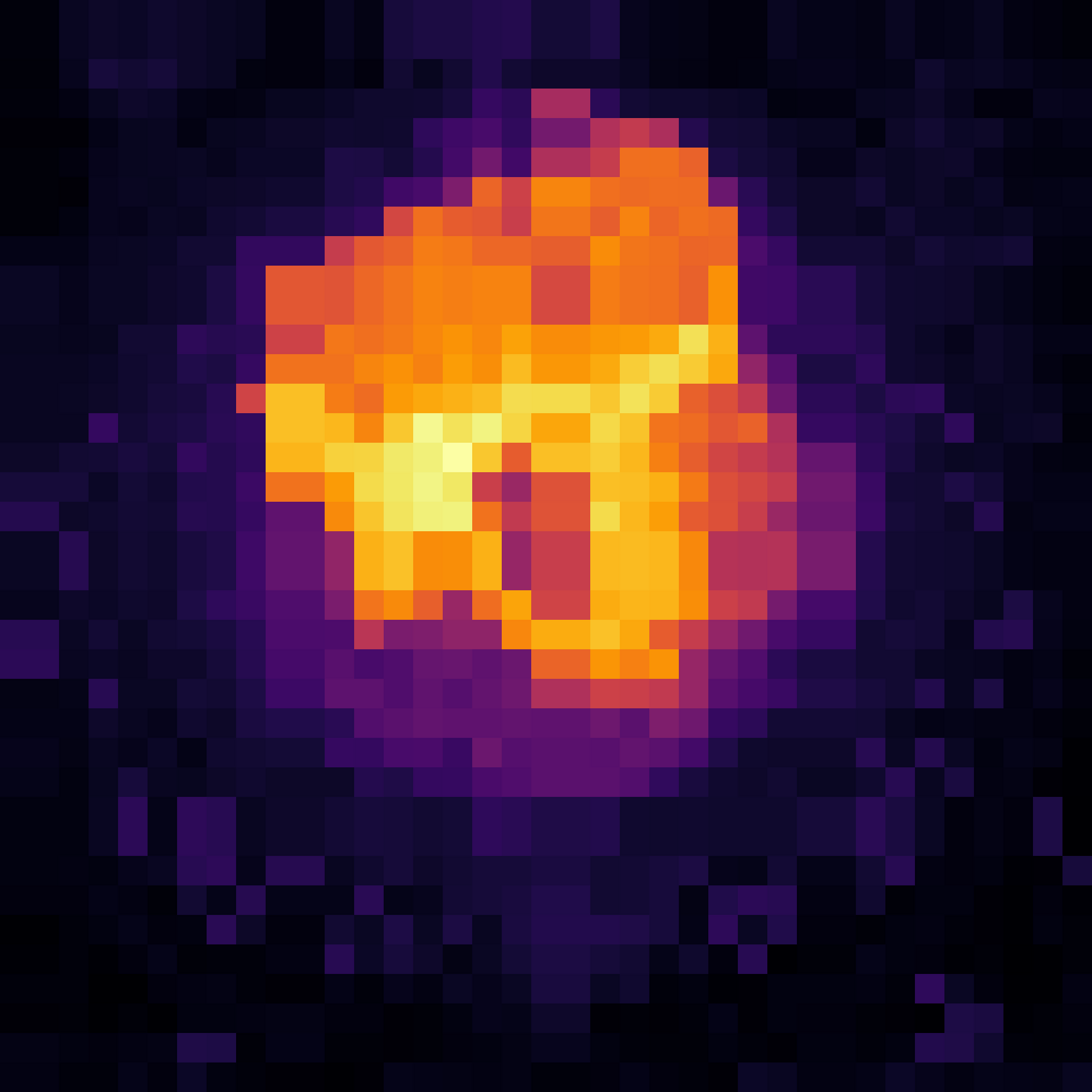}\end{minipage} & \begin{minipage}[t]{\linewidth}\centering \fixedimg{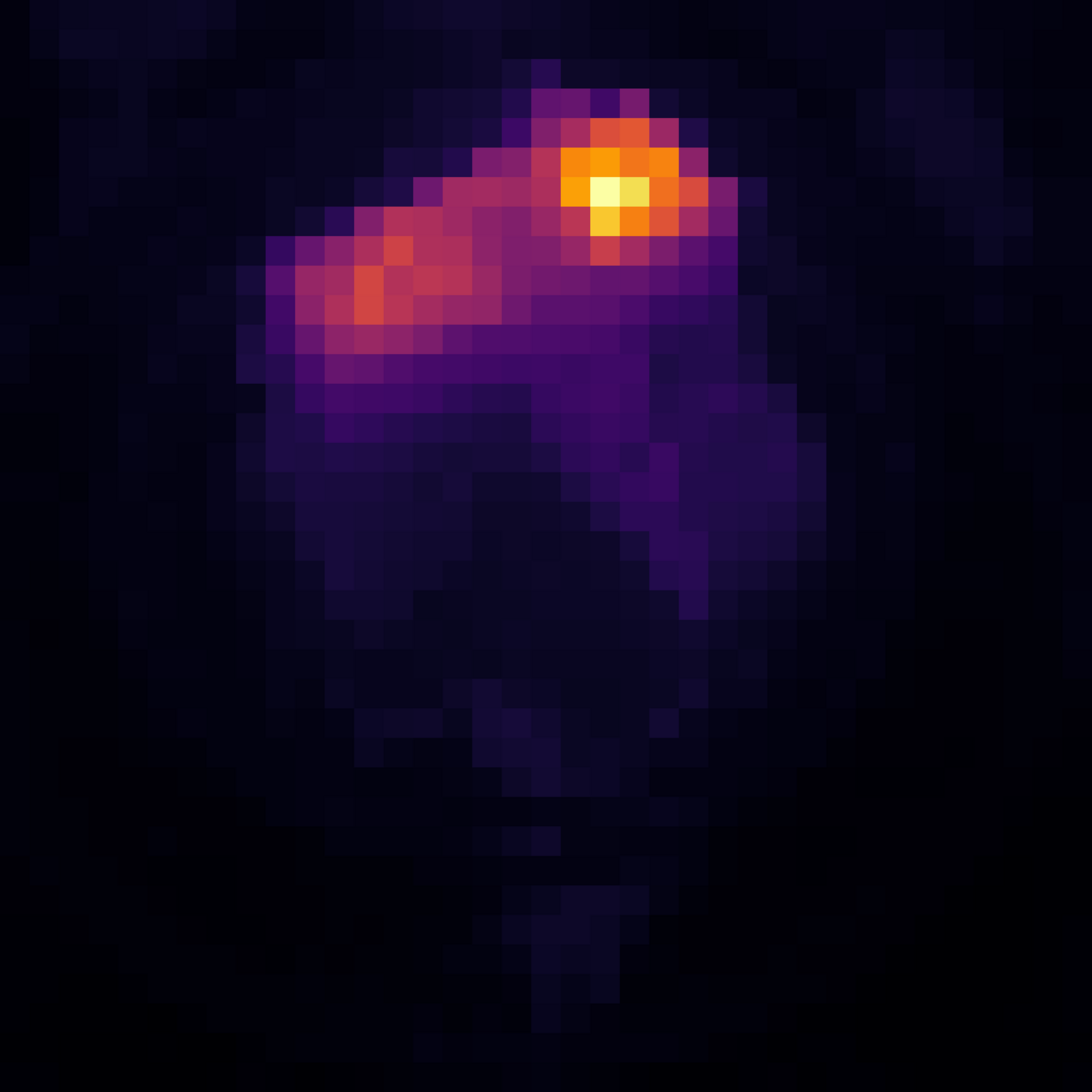}\end{minipage} & \begin{minipage}[t]{\linewidth}\centering \fixedimg{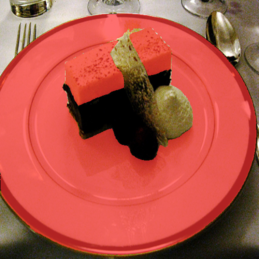}\\\end{minipage} & \begin{minipage}[t]{\linewidth}\centering \fixedimg{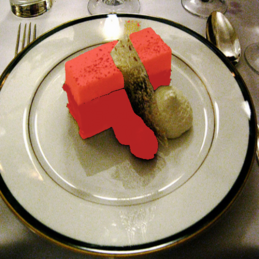}\end{minipage} & \begin{minipage}[t]{\linewidth}\centering \fixedimg{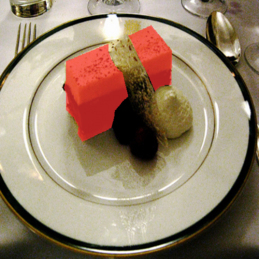}\end{minipage} \\
\centering cake & \centering Cake & \multicolumn{2}{>{\centering\arraybackslash}p{4cm}}{\textit{*Description not found*}} & \centering 8.21 mIoU & \centering 73.46 mIoU &  \\

\begin{minipage}[t]{\linewidth}\centering \fixedimg{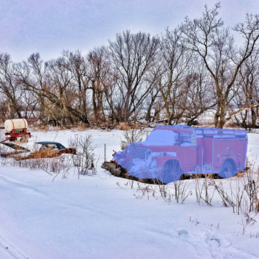}\\\end{minipage} & \begin{minipage}[t]{\linewidth}\centering \fixedimg{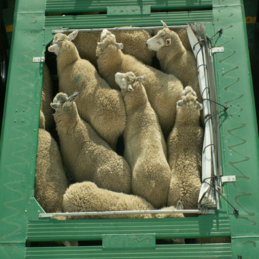}\\\end{minipage} & \begin{minipage}[t]{\linewidth}\centering \fixedimg{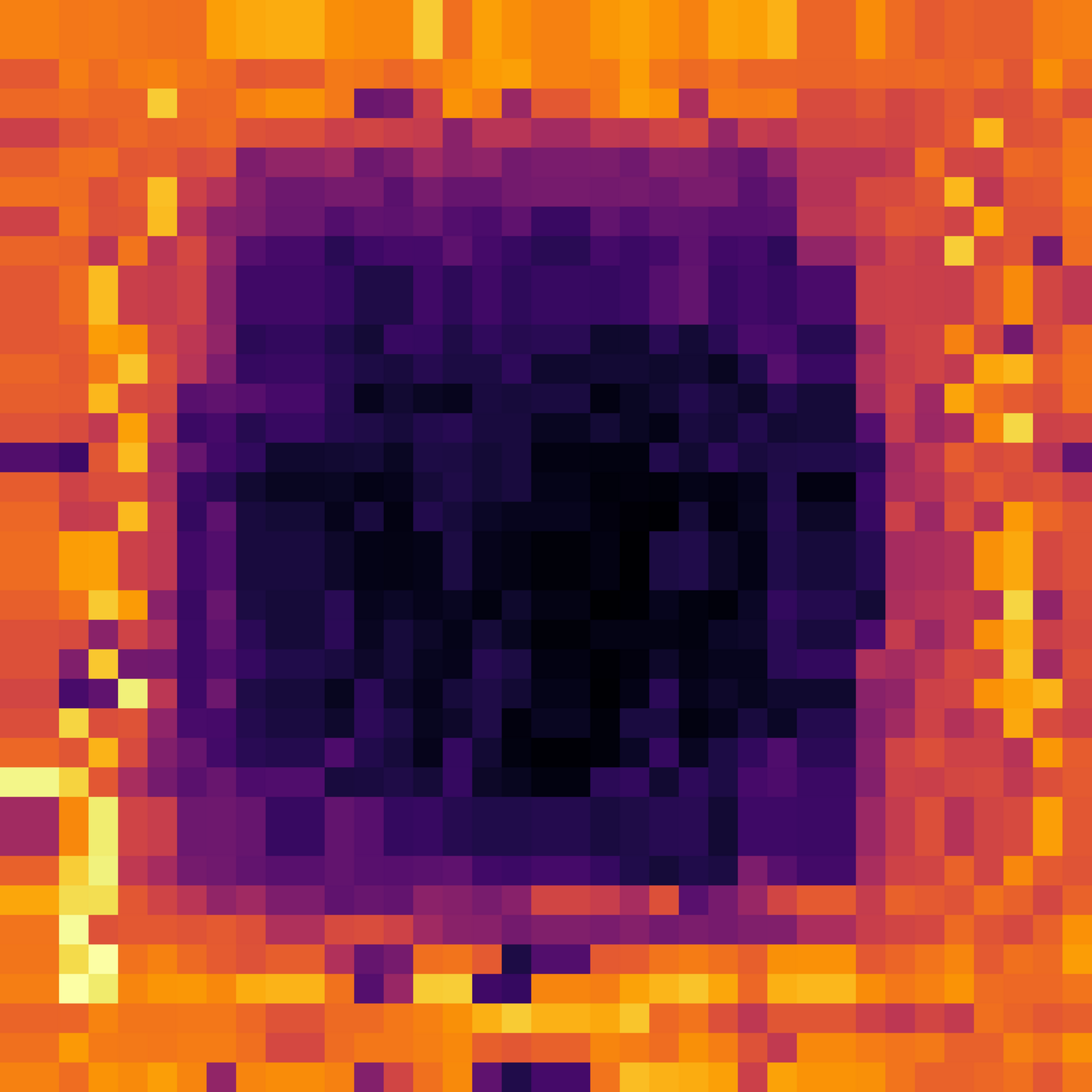}\end{minipage} & \begin{minipage}[t]{\linewidth}\centering \fixedimg{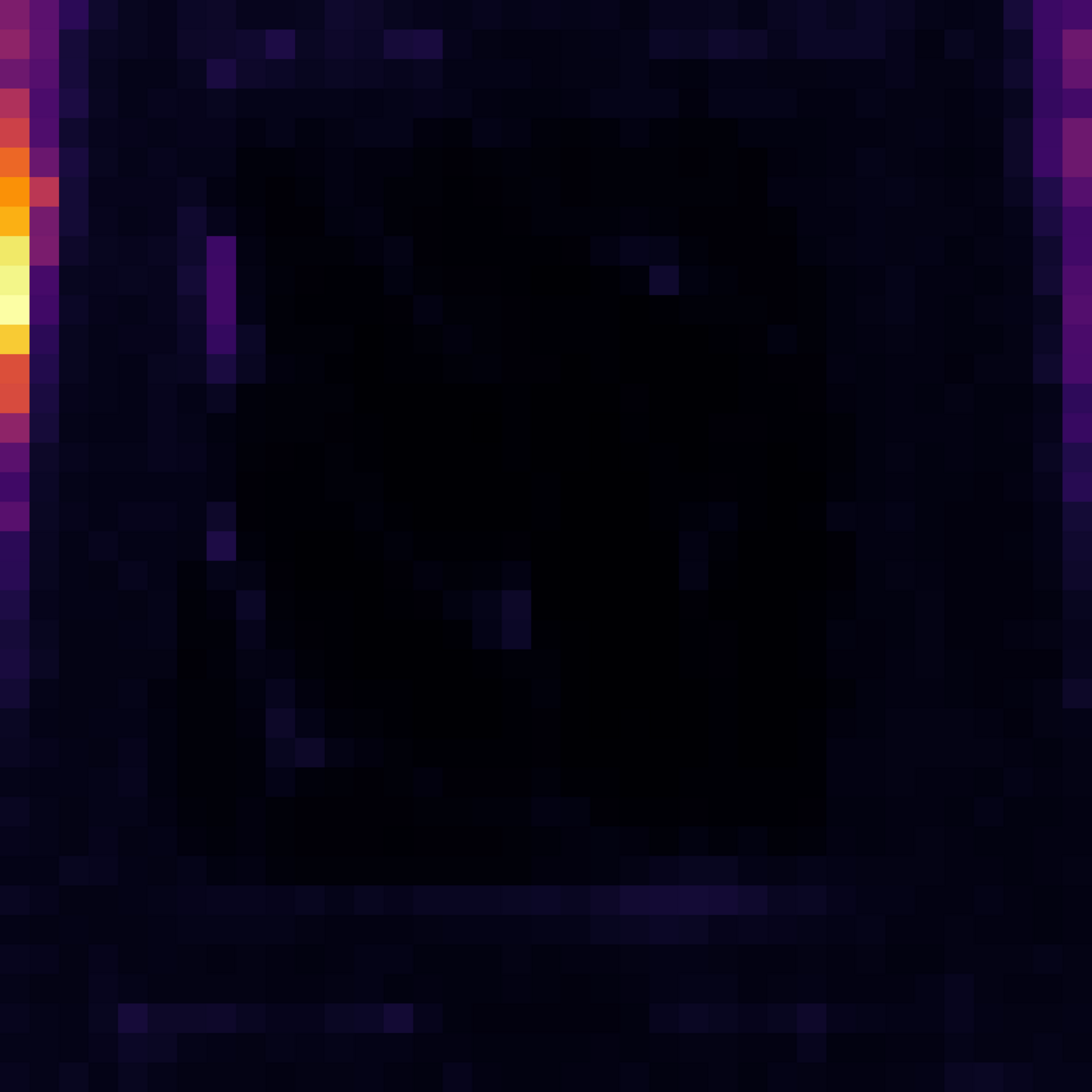}\end{minipage} & \begin{minipage}[t]{\linewidth}\centering \fixedimg{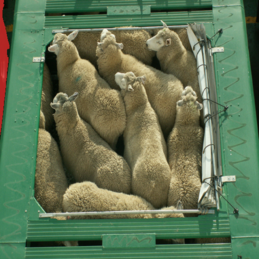}\\\end{minipage} & \begin{minipage}[t]{\linewidth}\centering \fixedimg{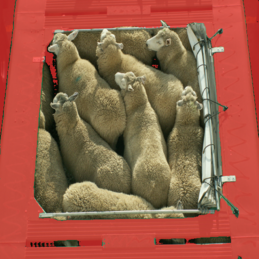}\end{minipage} & \begin{minipage}[t]{\linewidth}\centering \fixedimg{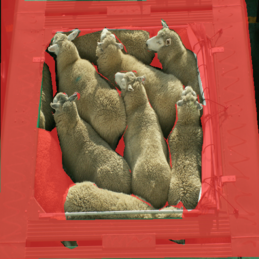}\end{minipage} \\
\centering truck & \centering Truck & \multicolumn{2}{>{\centering\arraybackslash}p{4cm}}{\textit{an automotive vehicle suitable for hauling}} & \centering 1.51 mIoU & \centering 75.79 mIoU &  \\

\bottomrule
\end{longtable}

\end{document}